\documentclass{IJCAS}

\usepackage{url}
\usepackage{array,tabularx}
\usepackage{multicol} 
\usepackage{tikz}
\usepackage{tikz-3dplot}
\usepackage{subcaption}
\usetikzlibrary{3d,decorations.markings,arrows.meta,positioning,shapes.geometric} 

\journalvolumn{VV}
\journalnumber{X}
\journalyear{YYYY}
\setarticlestartpagenumber{1}

\allowdisplaybreaks

\usepackage{amsmath,amsfonts,bm}
\usepackage{bbm}

\newcommand{\so}{\mathfrak{so}(3)}
\newcommand{\se}{\mathfrak{se}(3)}
\newcommand{\SO}{SO(3)}
\newcommand{\SE}{SE(3)}

\newcommand{\circoo}{~\raisebox{1pt}{\tikz \draw[line width=0.6pt] circle(1.1pt);}~}

\newcommand{\Real}{\mathbb{R}}

\newcommand{\X}{\mathcal{X}}

\newcommand{\tr}{\operatorname{tr}}

\newcommand{\removed}[1]{}

\newcommand{\G}{\mathbb{G}}
\newcommand{\Hgg}{\mathbb{H}}

\newcommand{\fg}{\mathfrak{g}}
\newcommand{\fgl}{\mathfrak{gl}}
\newcommand{\fse}{\mathfrak{se}(3)}
\newcommand{\fso}{\mathfrak{so}(3)}
\newcommand{\Lie}{\mathcal{L}}

\newcommand{\Id}{\mathbbm{1}}
\newcommand{\Ad}{ \operatorname{Ad}}
\newcommand{\ad}{ \operatorname{ad}}
\newcommand{\T}{\mathcal{T}}
\newcommand{\F}{F} 
\newcommand{\W}{k} 
\newcommand{\dX}{\mathcal{O}_X} 
\newcommand{\dY}{\mathcal{O}_Y} 

\newcommand{\eg}{e_{_G}}
\newcommand{\ev}{e_V}

\def\eqref#1{Equation~(\ref{#1})}

\def\1{\bm{1}}

\DeclareMathAlphabet{\mathsfit}{\encodingdefault}{\sfdefault}{m}{sl}
\SetMathAlphabet{\mathsfit}{bold}{\encodingdefault}{\sfdefault}{bx}{n}

\newcommand{\R}{\mathbb{R}}

\DeclareMathOperator*{\argmax}{arg\,max}

\begin{document}
\title{$SE(3)$-Equivariant Robot Learning and Control: A Tutorial Survey}

\author{Joohwan Seo*\orcid{0000-0003-2865-1844}, Soochul Yoo, Junwoo Chang\orcid{0009-0008-5843-6890},  Hyunseok An, Hyunwoo Ryu\orcid{0000-0002-0218-2675}, Soomi Lee\orcid{0009-0006-8514-9155}, Arvind Kruthiventy\orcid{0009-0002-8423-3735}, Jongeun Choi*\orcid{0000-0002-7532-5315}, and Roberto Horowitz\orcid{0000-0003-2807-0054}}

\begin{abstract}
Recent advances in deep learning and Transformers have driven major breakthroughs in robotics by employing techniques such as imitation learning, reinforcement learning, and LLM-based multimodal perception and decision-making.
However, conventional deep learning and Transformer models often struggle to process data with inherent symmetries and invariances, typically relying on large datasets or extensive data augmentation. Equivariant neural networks overcome these limitations by explicitly integrating symmetry and invariance into their architectures, leading to improved efficiency and generalization. This tutorial survey reviews a wide range of equivariant deep learning and control methods for robotics, from classic to state-of-the-art, with a focus on $SE(3)$-equivariant models that leverage the natural 3D rotational and translational symmetries in visual robotic manipulation and control design. Using unified mathematical notation, we begin by reviewing key concepts from group theory, along with matrix Lie groups and Lie algebras. We then introduce foundational group-equivariant neural network design and show how the group-equivariance can be obtained through their structure. Next, we discuss the applications of $\SE$-equivariant neural networks in robotics in terms of imitation learning and reinforcement learning. The $\SE$-equivariant control design is also reviewed from the perspective of geometric control. Finally, we highlight challenges and future directions of equivariant methods in developing more robust, sample-efficient, and multi-modal real-world robotic systems.
\end{abstract}


\maketitle

\makeAuthorInformation{
\textbf{Accepted to IJCAS}

Joohwan Seo, Soomi Lee, Arvind Kruthiventy, Jongeun Choi, and Roberto Horowitz are with Department of Mechanical Engineering, University of California, Berkeley, CA 94720, USA (\texttt{e-mails: \{joohwan\_seo, soomi\_lee, arvindkruthiventy, horowitz\}@berkeley.edu}).
Soochul Yoo, Junwoo Chang, Hyunseok An, and Jongeun Choi are with School of Mechanical Engineering, Yonsei University, Seoul, Republic of Korea, (\texttt{e-mails: \{usam205, junwoochang, hs991210, jongeunchoi\}@yonsei.ac.kr}).
Hyunwoo Ryu is with Computer Science and Artificial Intelligence Laboratory, MIT, Cambridge, MA 02139, USA (\texttt{e-mail: hwryu@mit.edu})\\
* Corresponding authors.
}

\runningtitle{2024}{Joohwan Seo et al.}{Manuscript Template for the International Journal of Control, Automation, and Systems: ICROS {\&} KIEE}{xxx}{xxxx}{x}

\section{INTRODUCTION}
Recent advances in deep learning have led to significant breakthroughs in the field of robotics. These advancements are frequently demonstrated through imitation learning, reinforcement learning, and recent large language model (LLM)-based language augmentation, enabling robots to better understand and adapt to complex environments. Nevertheless, one major challenge persists: Traditional deep learning and transformer models are not inherently designed to handle data with symmetries or invariance properties, often requiring large datasets or extensive data augmentation for optimal performance. Equivariant neural networks have emerged as a promising solution by explicitly integrating symmetry and invariance properties into their architecture, significantly improving efficiency and generalization.

Recently, fully end-to-end $SE(3)$-equivariant models have been proposed for visual robotic manipulation using point cloud inputs \cite{ryu2022equivariant, ryu2024diffusion, kim2023robotic}. These models leverage the inherent symmetries in $SE(3)$—the group of 3D rotations and translations—to achieve more efficient and accurate learning. Specifically, \cite{ryu2022equivariant, kim2023robotic} introduced $SE(3)$-equivariant energy-based models that utilize representation theory of Lie groups to enable highly efficient end-to-end learning. In \cite{ryu2024diffusion}, this approach was extended by incorporating diffusion methods to further enhance learning efficiency and inference speed.

In parallel, control design on the $SE(3)$ manifold structure has been investigated in \cite{seo2023geometric, seo2024comparison}. These methods were subsequently extended in \cite{seo2023contact} to learn manipulation tasks, demonstrating that incorporating equivariance principles at both the force and control levels is essential for effective learning of manipulation behaviors.

 
This tutorial survey paper aims to provide an up-to-date review of the emerging field of equivariant deep learning models for robotic learning, highlighting their potential to transform the way robots perceive and interact with the world. The contributions of this paper can be summarized as follows:
%
\begin{itemize}
    \item We review a group equivariant deep learning and control for robotics applications.
    \item We provide an in-depth review of important concepts and structures to facilitate understanding of group equivariant deep learning and their mathematics, from fundamental to applications.
    \item We also review the topics related to the group equivariant control method from the geometric control perspective.
    \item To avoid confusion from the different notations prevalent in robotics society, we describe the topics in unified mathematical notations throughout the paper.
\end{itemize}

This paper begins with an overview of relevant concepts in group theory, smooth manifolds, Lie groups and algebras, and the Special Euclidean group $\SE$ in Section~\ref{sec:Prelim}. 
Next, we introduce equivariant deep learning for the image data ($SE(2)$-equivariance) and point cloud data ($\SE$-equivariance) in Section~\ref {sec:Equivariant}.
Moreover, we present how the equivariance can be incorporated into the model architecture together with important equivariant neural network designs that could serve as backbones of equivariant models. Our discussion on these equivariant models focuses on group convolutional networks, steerability on $\SE$, and graph convolutional networks.

In Section~\ref{sec:Related}, we present two critical applications for equivariant learning methodology in robotics: imitation learning from expert demonstrations and reinforcement learning. We also present a detailed review of selected imitation learning and reinforcement learning algorithms from the literature. 

Furthermore, in Section~\ref{sec:GIC}, we briefly introduce concepts required to derive a control method on $\SE$ manifold structure, such as error function (metric), potential energy function, velocity error vector, and Riemannian metric. These essential concepts can also be exploited to derive $\SE$ equivariant deep learning models.

Finally, we discuss the current challenges and future directions of equivariant deep learning research in robotics in Section~\ref{sec:Future_work}. We highlight the need for more robust and efficient models that can handle complex and dynamic environments and discuss the potential of equivariant deep learning for multi-modal sensor fusion and lifelong learning in robotics. Concluding remarks are provided in Section~\ref{sec:Conclusion}.

\section{Preliminaries}\label{sec:Prelim}

In this section, we introduce mathematical preliminaries to understand geometric deep learning and control for robotics applications. We first introduce groups, Lie groups, and Lie algebras, with a focus on 
$SE(3)$, which is fundamental for analyzing rigid body transformations and numerous robotic tasks in vision and manipulation. 
Lie groups and, in particular $\SE$, are also fundamental concepts in group equivariant deep learning, which will be discussed in Section~\ref{sec:Equivariant}, as well as in the formulation of geometric control in Section~\ref{sec:GIC}. 
We also briefly touch upon concepts related to smooth manifolds, which in turn will be helpful in providing a fundamental understanding of infinitesimal variations on   Lie groups. As will be shown, the group action on velocities (twists) and forces (wrenches) can be best understood via differential geometry in smooth manifolds.

\subsection{Groups}

A group $(\G, \cdot)$ is defined by a set $\G$ along with a group product (a binary operator) $``\cdot"$  
that satisfies  the following axioms:
\begin{itemize}
\item $\forall h, g \in \G$ we have $h \cdot g \in \G$ (Closure);
\item  $\exists \Id \in \G$  such that $\Id \cdot g = g \cdot \Id =  g$ (Identity); 
\item 
$\forall g \in \G$, $\exists$  $g^{-1} \in \G$ such that
$g^{-1} \cdot g = g\cdot g^{-1}= \Id$ (Inverse);
\item  $\forall g, h, f\in \G$ we have $(g \cdot h) \cdot  f = g\cdot (h\cdot f)$ (Associativity).
\end{itemize}
We now provide several examples.

The {\em translation group} $(\mathbb{R}^n, +)$ consists of the set of  vectors in $\mathbb{R}^n$ equipped with the group product given by vector addition, $x_1 \cdot x_2 = x_1 + x_2$ for $x_1, \, x_2 \in \mathbb{R}^n$. The identity element is $\Id = 0$ (the origin of $\mathbb{R}^n$) and  the  inverse is  $g^{
-1}=(-g)$.

The {\em general linear group} over $\Real$, denoted by  $GL(n, \mathbb{R})$, consists of the set of $n \times n$ matrices with real element entries {\em that can be inverted},  $GL(n,\mathbb{R}):= \{ A \in \mathbb{R}^{n \times n} |\det (A) \neq 0\},$
together with the group product $``\cdot"$  given by matrix multiplication, i.e. $A \cdot B = AB$ for $A,B \in GL(n,\mathbb{R})$. The group inverse is given by the matrix inverse and the group identity element $\Id$ is given by the identity matrix $I_n$. Similarly, $GL(n, \mathbb{C})$ consists of the set of $n \times n$ matrices with complex  element entries that can be inverted $GL(n,\mathbb{C}):= \{ A \in \mathbb{C}^{n \times n} |\det (A) \neq 0\},$

A subgroup is a subset of a group $(\mathbb{H} \subset \G)$, which is itself a group that is closed under
the same group product $``\cdot"$   of $G$. Following are some important subgroups of $GL(n, \mathbb{R})$ and $GL(n, \mathbb{C})$.
By invoking the stronger restriction on elements of $GL(n, \mathbb{R})$ that their determinant must be a positive real number, we obtain  the sub group $GL^+(n, \mathbb{R}):= \{ A \in \mathbb{R}^{n \times n} |\det (A) >0\}$. 
 The {\em unitary group} is a  subgroup  of $GL(n,\mathbb{C})$ that consists of  matrices whose complex conjugate transpose are their inverses,    
$U(n):=\{A \in \mathbb{C}^{n \times n} | AA^* = {\Id}\} \subset GL (n,\mathbb C)$, where $A^*$ denotes the complex conjugate transpose of $A$.
Similarly, the {\em orthogonal group} consists of  real matrices, whose transpose are their inverses,   
$O(n):=\{A \in \mathbb{R}^{n \times n} | AA^T = {\Id}\}$.
By imposing the additional constraint  on elements of the orthogonal group $O(n)$  that their determinants must be one ($\det (A)=+1$), we obtain  the {\em special orthogonal group}, 
$SO(n) := \{ R \in \mathbb{R}^{n \times n} | R^TR=RR^T={\Id}, \det (R)=+1 \}$.
This important subgroup consists of the set of all rotation matrices.  

An important application of groups involves
their actions on other sets, groups, manifolds, or vector spaces. If $\G$ is a group and $M$ is a set, a {\em left
action of $\G$ on $M$} is a map  $\G \times M \to M$ written as $(g,p) \to g \circ p$, where $g \in \G$ and $p \in M$ and $``\circ"$ is the binary left action, which  satisfies 
\[
g_1 \circ (g_2 \circ p) = (g_1 \cdot g_2) \circ p, \hspace{1em} e \circ p = p, \hspace{1em}\forall g_1,g_2 \in \G,\, p\in M.
\]
{\em Right actions} are similarly defined.
\[
(p \circ g_1) \circ g_2  = p \circ (g_1 \cdot g_2) , \hspace{1em} p \circ e = p, \hspace{1em}\forall g_1,g_2 \in \G,\, p\in M.
\]
As an example of a left action, consider pure rigid body rotations  on three-dimensional real vectors, which can be represented by the action map $\SO \times \R^3 \to \R^3$, given by $(R,p) \to R \circ p = Rp$, where $R \in \SO$ is a rotation matrix, and $p \in \R^3$ is a three-dimensional vector and $R \circ p = R p$ is the matrix to vector multiplication operation.

Group actions give us a powerful  way to construct new groups. Suppose $\mathbb{H}$ and $\mathbb{N}$ are  groups, and $\theta_h : \mathbb{H} \times \mathbb{N} \to \mathbb{N}$ is a  left action of $\mathbb{H}$ on $\mathbb{N}$ given by $\theta_h(n) = h \circ n$, where $h \in \mathbb{H}$ and $n \in \mathbb{N}$. We  can then define a new  group $\mathbb{N} \rtimes \mathbb{H}$, called a {\em semidirect product} of  $\mathbb{N}$ and  $\mathbb{H}$, which has elements of the form $(n, h)$ with the group operation as follows.
\[
(n,h)\cdot (n',h') = (n\cdot \theta_h(n'), h \cdot h') = (n\cdot (h \circ n'), h \cdot h')
\]
Unlike a direct product, the semi-direct product incorporates the action of $\mathbb{H}$ on $\mathbb{N}$, creating an interaction between the two groups. 

We now introduce the important rotation-translation group, also known as  the 3-dimensional {\bf special Euclidean  group} denoted by 
$SE(3)$, which consists of all rigid body transformations (translations and/or rotations). This group can be generalized to $SE(n)$ and defined as the {\bf semidirect product} of $SO(n)$ and the translation group ($\R^n, +)$ (treating $\R^n$ as a group with group product given by vector addition).
Thus, the special Euclidean  group $SE(n)$ is the set  $\mathbb{R}^n \rtimes SO(n)$ of translation vectors in  $\mathbb{R}^n$ and rotations in $SO(n)$, where $\rtimes$ denotes semi-direct product defined above. The group product and the inverse are then, respectively, given by
\[ g_1 \cdot g_2 = (p_1 + R_1p_2 , R_1R_2)\hspace{1em}\mbox{and}\hspace{1em} g^{-1}=(-R^{-1}p, R^{-1}),
\]
where $g_1=(p_1,R_1)$ and $g_2=(p_2, R_2)$ are elements of $SE(n)$. This action preserves lines, distances, and angle measures, and consequently all of the relationships of Euclidean geometry.

\subsection{Matrix Lie Groups and Algebras}
\subsubsection{Lie Groups} \label{sec:lie_group}
A Lie group is a continuous group that is also a differentiable manifold, i.e., it is a group equipped with a smooth structure that allows continuous transformations and compositions, preserving both its algebraic and geometric properties.  Lie groups play a fundamental role in various areas of mathematics and physics, serving as a framework to study symmetries and transformations in a continuous manner.
Smooth manifolds are the foundation of differential geometry. 
Although an in-depth review of smooth manifold theory is beyond this paper's scope, we have included a brief review of basic manifold concepts in Appendix~\ref{sec:appendix_smooth} to aid readers.

Lie group examples include the translation group $(\mathbb{R}^n,+)$ 
the general linear group $GL(n,\Real)$, the group of invertible $n \times n$ matrices with real entries, $SO(n)$, the group of rotations in $n$-dimensional space, and $SE(n)$, the special Euclidean  group in $n$-dimensional space.

Lie groups emerge from symmetry groups governing physical systems, with their corresponding Lie algebras representing infinitesimal symmetry motions as tangent vectors in the vicinity of the identity element $\Id  = I_n$. 
In robotics, Lie groups of rigid-body motion, i.e., $\SE$, represent the symmetries in the motion of various points on the robot, such as the end-effector. On the other hand, their Lie algebras, such as twists, capture the small incremental movements of a system via the robot's joints.
This elegant mathematical framework connects Lie groups, Lie algebras, and physical motions, offering a powerful tool to model symmetries and transformations in various robotic applications, which will be further explored in the following sections.

In order to make our exposition more concrete, we will limit our discussion and examples to matrix Lie groups and their respective Lie algebras in this paper. A matrix Lie group $(\G, \cdot)$ is a group in which each element $g \in \G$ is an $n \times n$ matrix, the binary group operation $\cdot$ is matrix multiplication, and the mappings $a(g_1, g_2) = g_1 \cdot g_2$ and matrix inversion $b(g) = g^{-1}$  are both analytic.
Notice that the dimension of a Lie group is the dimension of the associated manifold $\G$, which in many instances will be different from the dimension of the matrices, i.e. $g \in \G \subset \Real^{n \times n}$. Every matrix Lie group considered in this paper will be a subgroup of $GL(n, \Real)$ or $GL(n, \mathbb{C})$. Thus, when we refer to Lie groups throughout this paper, what will be meant for the most part is matrix Lie groups.

\paragraph*{Notation}
 
In our notation, the group action on the group itself is denoted by $(\cdot)$, which represents standard matrix multiplication for matrix groups; hence, we typically omit the symbol when the context is clear. This is again because we will constrain our domain of discussion to only matrix Lie groups. The operation $(\circ)$ is frequently used to represent the action of a matrix group on other sets, groups, manifolds or vector spaces, including for example the multiplication of a group element $g$ (utilizing homogeneous representations) with a vector $q$, augmented to $\bar{q}$ by putting $1$ in its last row. As noted in Chapter \ref{sec:SE(3)_as_Lie_group} and further implied in Chapter \ref{sec:equiv_gnn}, this operation is ultimately denoted more concisely as $gq$.
Finally, we also utilize $(\circ)$  to denote the composition of two functions, for example,   $h(f(x)) = (h \circ f)(x)$, particularly in the Smooth Manifolds appendix in section \ref{sec:appendix_smooth}.  

\begin{figure}[!t]
\begin{center}
\includegraphics[width=0.8\columnwidth]{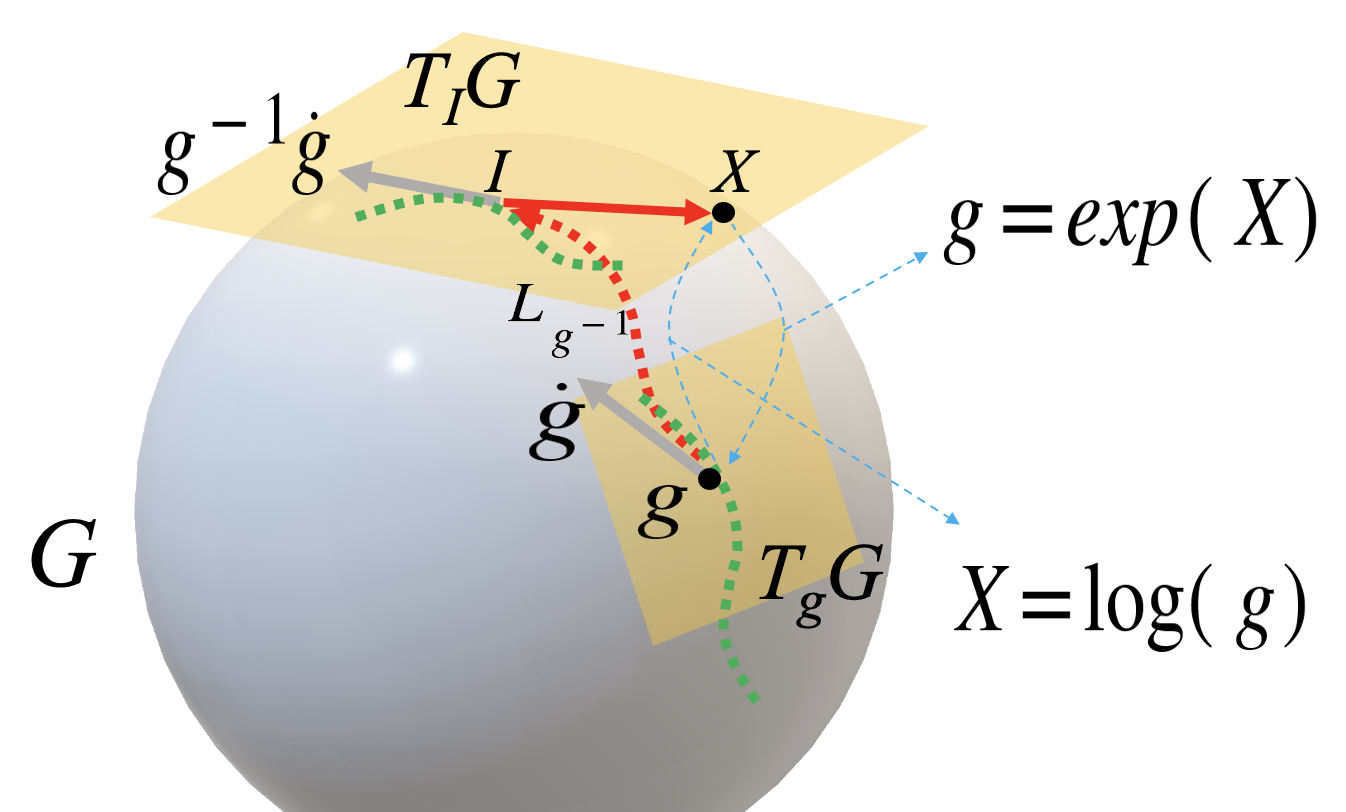} 
\vskip -0.5pc
\caption{Illustration of   a Lie group $\G$ and two of its tangent spaces.
The Lie algebra $\fg = T_{I_n}\G$ is the tangent space at the identity $I_n$.
}\label{TG}
\end{center}
\end{figure}

\subsubsection{Lie Algebras}
Consider an element of the general linear matrix Lie group $\G \subset GL(n,\mathbb{R})$, which is extremely close to the identity matrix $I_n$. Such a {\em perturbed} element can be denoted as $g(\epsilon) = I_n +\epsilon X$, where $\epsilon \in \Real$ is a small number ( $|\epsilon | \ll 1$) and the matrix $X$ is and element of $\fg = T_{I_n} \G$, the tangent space of $\G$ at the identity element $I_n$ (see Fig. \ref{TG} ).  $X \in \fg \subset \mathbb{R}^{n \times n}$ is often called a generator and $\fg = T_{I_n} \G$, the tangent space of $\G$ at its identity element $I_n$, is called the Lie alegra of $\G$.
This notation of $g(\epsilon)$ is often called a single- or one-parameter subgroup, as they are dependent on a one-parameter $\epsilon \in \mathbb{R}$.
The Lie algebra $\fg$ can be interpreted as the set of all infinitesimal transformations generated by the Lie group, starting from $I_n$. 

As an illustrative example, we will determine the Lie algebra of $SO(2)$  (see also Chapter 5 of \cite{ping2002group} and \cite{huang2020lie}). The matrix representation of $SO(2)$ is given by the planar rotation matrix $R_\theta = R(\theta) \in \mathbb{R}^{2\times2}$, with  $\theta$ being the rotation angle with respect to the $z$- axis.
\begin{equation}
\label{eq:RtheSO2}
    R_\theta = R(\theta) = \begin{bmatrix}
        \cos{\theta} & -\sin{\theta} \\
        \sin{\theta} & \cos{\theta}
    \end{bmatrix},
\end{equation}
with  $R(0)=I_2$. An infinitesimal perturbed element $R(\epsilon) \in SO(2)$ close to the identity matrix $I_2$ can be represented as
\begin{equation*}
   R(\epsilon) \approxeq I_2 + \epsilon \left.\dfrac{\partial R(\theta)}{\partial \theta}\right|_{\theta=0} = I_2 + \epsilon X , \:\: |\epsilon | \ll 1,
\end{equation*}
where the generator $X \in \mathfrak{so}(2) = T_{I_2}SO(2)$ is given by 
 \begin{equation*}
 \label{eq:Xso2}	
    X  = \left.\dfrac{\partial R(\theta)}{\partial \theta}\right|_{\theta=0} = \begin{bmatrix}
        0 & -1 \\
        1 & 0 
    \end{bmatrix}
\end{equation*}
Here, the Lie algebra of $SO(2)$ is $\mathfrak{so}(2)  = \{ A \in \mathbb{R}^{2 \times 2} | A^T = - A \}$, the set of all $2 \times 2$ skew symmetric matrices. Notice that both $SO(2)$ and $\mathfrak{so}(2)$ have dimension 1. However, while $SO(2)$ is a group but not a vector space, its Lie algebra $\mathfrak{so}(2) = T_{I_s} SO(2)$ - the tangent space of $SO(2)$ at the identity $I_2$, is not a group but is a one-dimensional vector space. 

We  now utilize the infinitesimal transformation $R(\epsilon) = I_2 + \epsilon X$, with $X \in \mathfrak{so}(2)$, 
to obtain a finite transformation $R(\theta)$ from the identity element $I_2$ by multiplying the  infinitesimal perturbation $R(\epsilon)$ to the identity element $I_2$ infinitely many times. Defining the infinitesimal angle $\epsilon = \theta / N$, where $N \to \infty$, leads to
\begin{equation} \label{eq:SO(2)_derivation}
    \begin{split}
    R(\theta) &= \lim_{N\to \infty}  \left ( R(\epsilon) \right )^N I_2 = \lim_{N\to \infty}  \left(I_2 + \frac{\theta}{N}X\right)^N \\
    &= I_2 + \theta X + \frac{\theta^2}{2!}X^2 + \frac{\theta^3}{3!}X^3 + \cdots \\
    &= \exp{(X\theta)},
    \end{split}
\end{equation}
where $\exp(\cdot)$ used in \eqref{eq:SO(2)_derivation} is the standard matrix exponential.
This result from the $SO(2)$ example can be generalized to any matrix Lie group $\G$. Consider the general matrix Lie group $GL(n,\mathbb{R})$, the set of invertible real $n\times n $ matrices, and a subgroup $\G \subset  GL(n,\mathbb{R})$. We denote $\fgl(n, \Real) = T_{I_n} GL(n, \Real)$ and $\fg = T_{I_n}\G$ as the Lie algebras of $GL(n,\mathbb{R})$ and $\G$ respectively. We also note that $\fgl(n, \Real)$ and $\fg \subset \fgl(n, \Real)$ are both vector spaces.  

An element $g \in \G $ sufficiently close to the identity matrix $I_n$ can be  written using the matrix exponential, for some $X \in \fg \subset \fgl(n, \Real)$ and some $t \in \Real$ via the matrix exponential
\begin{equation} \label{eq:exp}
    g(t) = \exp{(tX)} = \sum_{n=0}^\infty \frac{1}{n!}(tX)^n.
\end{equation}
Thus, the (matrix) Lie algebra $\fg$ of $\G$ can also  formally be defined as the set of all matrices $X \in \Real^{n \times n}$ such that the exponential of each $tX$ results in an element of $\G$, i.e., 
\[
\fg= \{ X \in \Real^{n \times n}\, | \: \exp(tX) \in \G \subset GL(n, \Real), \forall t \in \Real \}.
\]
For
any  matrix  $X \in \fg$, its matrix exponential $\exp(X)$ is   given by  the series in \eqref{eq:exp}  with $t = 1$. 
The exponential maps $X \in \fg$ into $\exp(X) \in \G$. From this definition, it is straightforward to show that 
\begin{equation*}
    \left.\dfrac{d}{dt}\exp{(tX)} \right|_{t=0} = X.
\end{equation*}

The inverse of the exponential map $\exp:\fg \to \G$ is defined as a log map $\log: \G \to \fg$, i.e., if $g = \exp{(X)}$, it follows that $X = \log{(g)}$. Associated with the log-map, given that $\fg$ is a vector space of dimension $l$, 
we will define the isomorphism of the hat-map as follows. 
\begin{equation*}
    \widehat{(\cdot )}: \Real^l \rightarrow \fg,
\end{equation*}
that maps a vector in $\Real^l$ to $ \fg = T_{I_n} \G$. Its inverse map, vee-map, is defined as
\begin{equation*}
    (\cdot )^\vee: \fg : \rightarrow \Real^l,
\end{equation*}
where, as stated above, $l$ is the dimension of $ \fg = T_{I_n} \G$.
This definition of hat-map and vee-map and their usage in robotics applications will be further elaborated  in the later part of the paper. As the notion of various morphisms appears frequently in the robotics literature following differential geometric formulation. We will provide the definitions of those and their examples in the later section, Section \ref{sec:group_representations}.


\subsubsection{Lie Bracket}
 A Lie algebra $\fg$ is a vector space equipped with a Lie bracket operation
$[ \cdot, \cdot ] : \fg \times \fg \rightarrow \fg$ which is bilinear, anti-symmetric, $[a, b] = -[b, a]$, and satisfies the
Jacobi identity
$[a, [b, c]] + [b, [c, a]] + [c, [a, b]]=0.$ 

Let $\mathfrak{X}$ represent the set of smooth vector fields on $\G$. For any smooth function $f :\G\rightarrow \Real$,   the Lie bracket $[\cdot,\cdot]:\mathfrak{X} \times \mathfrak{X}\rightarrow \mathfrak{X}$ is then defined such that $[X,Y](f) := X(Y(f))-Y(X(f))$.
The intuitive meaning of a Lie bracket is that it works as an infinitesimal measure of noncommutativity. For example, in the Lie algebra of $\SO$, the basis vectors of $\fso$ can be written as $L_x$, $L_y$, and $L_z$, i.e.,:
\begin{equation*}
    L_x \!=\! \begin{bmatrix}
        0 & 0 & 0 \\
        0 & 0 & -1 \\
        0 & 1 & 0
    \end{bmatrix},
    L_y \!=\! \begin{bmatrix}
        0 & 0 & 1\\
        0 & 0 & 0\\
        -1 & 0 & 0
    \end{bmatrix},
    L_z \!=\! \begin{bmatrix}
        0 & -1 & 0\\
        1 & 0 & 0\\
        0 & 0 & 0
    \end{bmatrix}.
\end{equation*}
Then, the following holds:
\begin{equation*}
    [L_x, L_y] = L_z
\end{equation*}
and their cyclic permutations also hold. The physical meaning is that rotation around $x$ and $y$, compared to its reverse order, results in a small rotation around $z$.

\subsubsection{Group Representations} \label{sec:group_representations}
Before introducing group representation, we first provide a nomenclature for mappings that is frequently utilized in Lie Group theory.  

 In algebra, a {\bf homomorphism} is a structure-preserving map between two algebraic structures of the same type, such as two groups. Therefore, a {  homomorphism} between Lie groups is a smooth  $\mathcal{C}^\infty$ map $f: \G \to \Hgg$ between  Lie groups $(\G, \cdot)$ and $(\Hgg, \hat \cdot)$ that respects the group structure, i.e., $\forall g_1, g_2 \in \G$, $f (g_1 \cdot g_2) = f(g_1) \hat \cdot f(g_2)$, and $f({e_G}) = e_H$, where $e_G$ and $e_H$ are the identity elements of $\G$, and $\Hgg$, respectively. 
 
 A homomorphism is an {\bf isomorphism} if and only if it is bijective (one-to-one). In other words, an isomorphism of Lie groups is a structure-preserving mapping between two groups that can be reversed by an inverse mapping. An {\bf automorphism} is an isomorphism from a group to itself.


A {\bf homeomorphism} is an isomorphism between two topological spaces. More formally, given two topological spaces $X$ and $Y$, a continuous map ${ \phi \colon X\rightarrow Y}$ is called a homeomorphism if it is a bijection (i.e., one-to-one), continuous, and its inverse ${ \phi^{-1} \colon X\rightarrow Y}$ is also continuous. For example, a function $\phi:(0,1) \to \Real$, defined as $\phi(x) = \tan{\left(\pi x - \tfrac{\pi}{2}\right)}$ is a homeomorphism. 

A {\bf diffeomorphism} is an isomorphism between spaces equipped with a differential structure, typically differentiable manifolds. More formally, given two manifolds 
$M$ and $N$, a differentiable map ${ \phi \colon M\rightarrow N}$ is called a diffeomorphism if it is a bijection and its inverse 
${\phi^{-1}\colon N\rightarrow M}$ is differentiable as well. 

\paragraph*{Group Representation}
If a linear map is a bijection, i.e. injective (one-to-one) and surjective (onto), then it is called a linear isomorphism.
Assume that $V$ is an $n$-dimensional vector space over $\Real$. We define 
\[GL(V ) = \{ A : V \rightarrow V\; |\; A \text{ is a linear isomorphism} \}  \]

Let $\G$ be a Lie group and $V \in \Real^n$  a vector space. 
A representation of a Lie group is a map $D: \G \rightarrow  GL(V )$, which satisfies 
the following property:
\begin{equation}
    D(g) D(h) = D(g \cdot h) \quad \forall g,h \in \G
\end{equation}
 Two different representations $ {D}$ and $ {D}'$ are said to be of \textit{equi\underline{valen}ce} (not   \textit{equi\underline{varian}ce})
 if there exists a non-degenerate change of basis $ {U}$ such that
 \begin{equation}
      {D}'(g)= {U} {D}(g) {U}^{-1}, \quad\forall g\in \G
 \end{equation}
Note also that a representation is said to be \textit{reducible} if there exists a change of basis such that the representation can be decomposed (block-diagonalized) into smaller subspaces. An \textit{irreducible representation} is a representation that cannot be reduced any further. 

\paragraph*{Lie Group Left and Right Translations}
Note first that what we refer to as "translation'' in this section is different from a translational motion used in rigid-body kinematics and dynamics.
Let $g, h \in \G$ be elements of a Lie group $\G$. We define the left translation of $\G$ by $g$ as the map
$L_g : \G \to \G,$ such that  $L_g(h) = gh$,  $\forall h \in \G$, and right translation as the map $R_g: \G \to \G$, such that $R_g(h) = hg$, $\forall h \in \G$. 

Notice that 
\[L_g^{-1} (h)=L_{g^{-1}}(h) \text{ and  } R_g^{-1} (h)=R_{g^{-1}}(h)\]
 since $L_g (L_g)^{-1} (h)=g (g^{-1})h=h$ and  $R_g (R_g)^{-1} (h)=h g (g^{-1})=h.$ 
Henceforth, both $L_g$ and $R_g$ are diffeomorphisms.  In addition, they commute with each other, i.e., $L_g R_h=R_h L_g$.

A left-invariant vector field $X \in T G$, where $TG$ is the tangent bundle of $G$ (see Appendix~\ref{sec:appendix_smooth}), satisfies 
\[
X(gh)=dL_g(h) X(h), 
\]
where $dL_g(h)$ is the differential (tangent map) of $L_g$  evaluated at $h$. If $h=\Id,$ the group identity element, then $X(g)=dL_g(\Id) \zeta,$ where $\zeta=X(\Id).$ Moreover, since for Lie matrix groups, $L_g(h) = g \cdot h = g h$ is the left matrix multiplication of $h$ by $g$, we can simplify notation as follows, a left invariant vector field $X \in TG$ of a Lie matrix satisfies $X(g)= g X(\Id)$.

\subsubsection{Adjoint Representations}
Due to the fact that matrix  multiplication and inverse maps are smooth, left and right translations,
 $L_g$ and $R_g$, are diffeomorphisms.
Therefore, the inner automorphism $R_{g^{-1}} \circoo L_g= R_{g^{-1}}L_g$  is also smooth and given by  
$\Psi_g(h)=R_{g^{-1}}L_g(h)
 = g h g^{-1}.$

\paragraph*{Adjoint Representation of a Lie group $\G$}
The differential of $\Psi_g=R_{g^{-1}} L_g$ at the identify $\Id$ i.e.,   $d\Psi_g(\Id): \fg \to \fg$  is an isomorphism of Lie algebras, denoted by $\Ad_g: \G \times \fg \to \fg$.

\begin{figure}[!t]
    \centering
    \includegraphics[width=0.8\columnwidth]{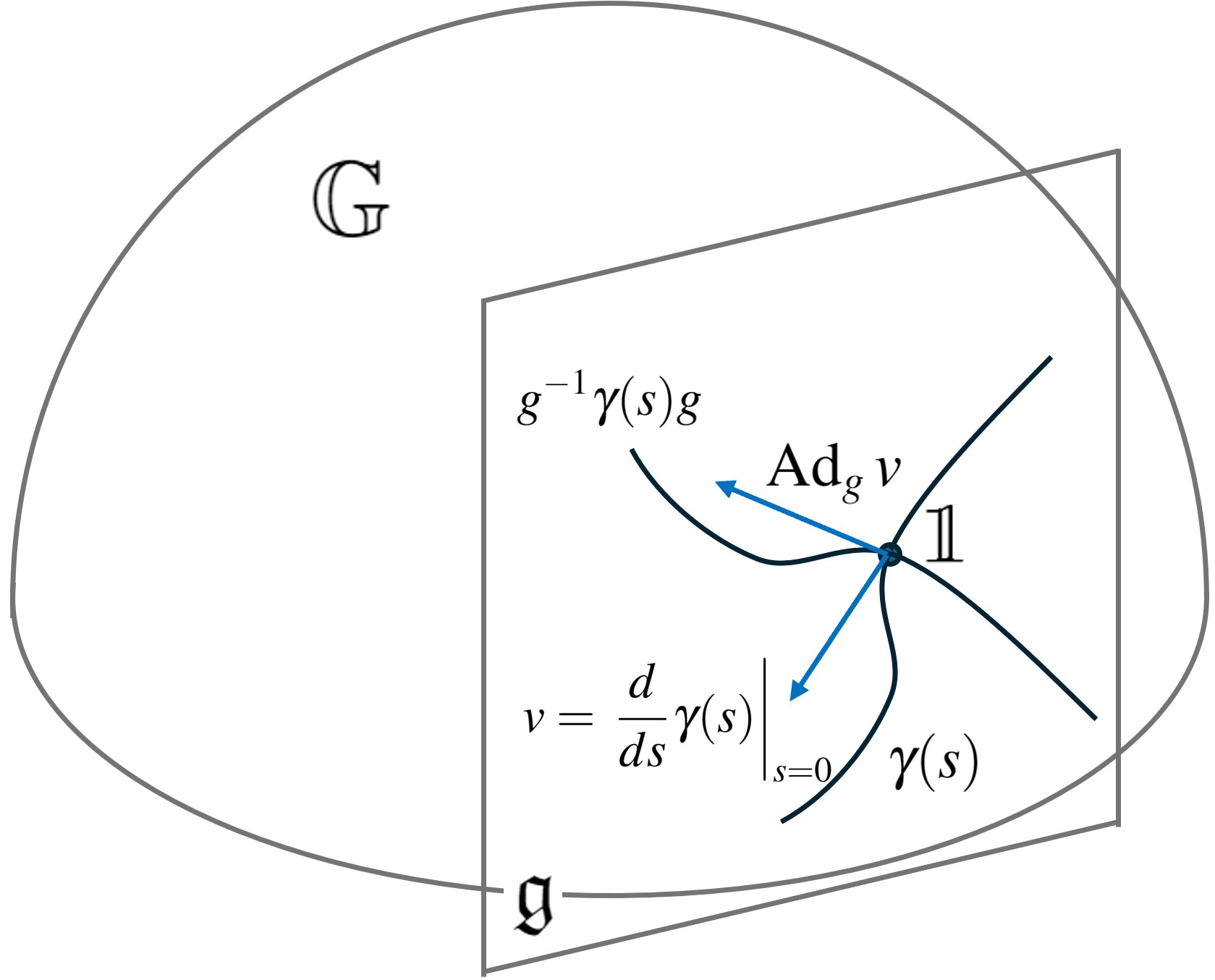}
    \caption{An illustration of   a Lie group $\G$ and its Lie algebra  $\fg$ of the tangent space at $\Id$.
    The  Adjoint  of $g$ applied on an element of $\fg$, i.e.,  $v=\frac{d}{ds} \gamma(s)|_{s=0}$ is illustrated.    Note that $Ad_g v \in \fg.$
     $\Psi_g(\Id)=g\Id g^{-1}=\Id$
    implies  that any curve $\gamma(s)$ through $\Id$
     on $\G$
     is mapped by this homomorphism $\Psi_g$ to another  curve  $g \gamma(s) g^{-1}$ on  $\G$ through $\Id$.
     }
     \label{fig:Adjoint}
\end{figure}

Given an element $X \in \fg$ of the Lie algebra and an element $g \in \G$ of the Lie group, the large adjoint (Adjoint) operator is then obtained as
\[
\begin{split}
\Ad_g X &:= \frac{d}{dt} (g \cdot \exp(tX) \cdot g^{-1})|_{t=0}\\
&= \frac{d}{dt} \exp (t g X g^{-1}) |_{t=0} \\
&=gXg^{-1}.
\end{split}
\]
The large adjoint representation of $\G$ gives a homomorphism from the group $\G$ into the set of all invertible linear transformations of $\fg$ onto itself.  
This  Lie
algebra isomorphism enables change of frames (see Fig~\ref{fig:Adjoint}).

Note that in the robotics community, a large adjoint representation is often interpreted as a linear operator in $\SE$, i.e., $\Ad_g: \SE \times \Real^6 \to \Real^6$, where $\Real^6$ is an isomorphic vector space of $\fg \in \se$ and $g \in \SE$. Although it might be confusing, as both adjoint representations are one-to-one, we will not distinguish their notations.

 \paragraph*{Adjoint Representation of a Lie algebra  $\fg$}
 The differential of $\Ad$ at the identity $\Id$, i.e.,  $d(\Ad_{\Id}): \fg \rightarrow \fgl(\fg)$  is a map of Lie algebras, denoted by
  $ad: \fg \rightarrow \fgl(\fg)$, and called the adjoint representation of $\fg$. It is  define by 
\[\ad_X Y=[X,Y].\]

The relationship between $\Ad$ and $\ad$ is given by
\[\Ad_{(\exp tX)}=\exp(t \cdot \ad_{(X)}).\]
(as we have $g=\exp (X)$.)

\begin{figure}[!t]
    \centering
    \centering
    \resizebox{0.9\columnwidth}{!}{
    \begin{tikzpicture}[node distance=2cm]

\coordinate (N1up) at (-3,3);
\coordinate (N2up) at (3,3);

\coordinate (N1p) at (-1,2.1);
\coordinate (N2p) at (1,2.1);

\coordinate (N3p) at (-1,-0.9);
\coordinate (N4p) at (1,-0.9);

\coordinate (N1) at (-3,2);
\coordinate (N2) at (3,2);
\coordinate (N3) at (-3,-1);
\coordinate (N4) at (3,-1);

\node (s1up)  at (N1up){Lie Algebra};
\node  (s2up) at (N2up) {Lie Group};

\node (s1)  at (N1){$X=\log(g) \in \fg $};
\node  (s2) at (N2) {$g=\exp(X) \in \G$};
\node (s3) at (N3) {$\ad_X(\cdot) \in \fgl(\fg)$};
\node (s4) at (N4) {$\Ad_g(\cdot) \in GL(\fg)$};

\draw[-{Latex}] (s1) -- (s2) node[midway, below]{\(\exp\)}; 
\draw[-{Latex}] (s2) -- (s4) node[midway, right ]{\(\Ad\)}; 
\draw[-{Latex}] (s3) -- (s4) node[midway, below]{\(\exp\)}; 
\draw[-{Latex}] (s1) -- (s3) node[midway, left]{\(\ad\)};

\draw[-{Latex}] (N2p) -- (N1p) node[midway, above]{\(\log\)}; 
\draw[-{Latex}] (N4p) -- (N3p) node[midway, above]{\(\log\)};

\end{tikzpicture}}
    \caption{Summary of Lie algebra and Lie group and their adjoints that take an element of Lie algebra for $(\cdot)$.
    }\label{fig:Summary}    
\end{figure}
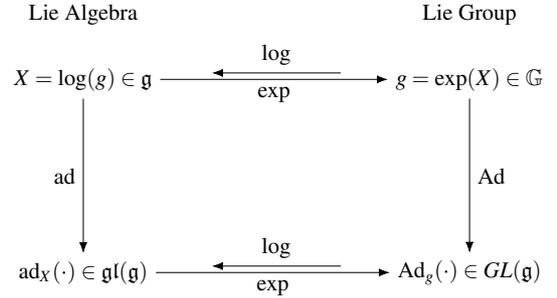

Fig.~\ref{fig:Summary} provides an illustrative summary of the relationships between Lie algebras, Lie groups and their adjoints.

The differential of $L_{g^{-1}}$ at $g$ is a linear
map  from   the tangent space at $g$, $T_g \G$, to  the Lie algebra, $T_\Id \G=\fg$,  because $L_{g^{-1}}g=g^{-1}g=\Id$. More precisely,
given a differentiable curve $g(t)$ on $\G$ with its velocity $\dot g(t):=\frac{d g(t)}{dt}$, the  differential of $L_{g^{-1}}$, i.e., $dL_{g^{-1}} : T_g \G \rightarrow \fg$ maps $\dot g$ to an element of $\fg$ such as 
\[dL_{g^{-1}}(\dot g) =g^{-1} \dot g \in \fg.\]
Similarly, the right differential can be defined as 
\[dR_{g^{-1}}(\dot g)=\dot g g^{-1} \in \fg.\]

In the next section, we  show that for $g\in SE(3)$,  $g^{-1} \dot g$ and $\dot g g^{-1}$  respectively represent the body and spatial velocities.

\paragraph*{Inverse and Composition Rules}
Recall that $\Ad_g$ can be defined as the differential of the inner automorphism $\Psi_g(h) = R_{g^{-1}} L_g(h) = ghg^{-1}$, evaluated at the identity $h = \Id$. The adjoint representation is  thus linear and acts on the Lie algebra $\mathfrak{g}$ of $\G$ as follows:
\[
\Ad_g: \mathfrak{g} \rightarrow \mathfrak{g}, \quad \Ad_g(X) = d(\Psi_g)_\mathrm{\Id}(X) = g X g^{-1},
\]
where $X$ is an element of the Lie algebra $\mathfrak{g}$. 

For $g, h \in \G$, we see that  $(\Psi_g)^{-1}$ must satisfy  
\[
   \begin{split}
(\Psi_g) (\Psi_g)^{-1}h&=g [ (\Psi_g)^{-1}h] g^{-1} = g [ g^{-1} h g ] g^{-1} = h,\\
  (\Psi_g)^{-1} (\Psi_g) h&= (\Psi_g)^{-1}  g h g^{-1} = g^{-1}  [ g h g^{-1}  ] g = h.
  \end{split}
\]
Therefore, we have
\[
(\Psi_g)^{-1}= \Psi_{g^{-1}}.
\]

For $g, h, k \in \G$,  the composition rule  of inner automorphisms is given by 
\[\Psi_{gh (k)}=\Psi_{g}  \Psi_{h} (k),\]
which shows that it is a homomorphism.{\color{red} } This can be shown as follows:
\[
    \begin{split}
    \Psi_{g}  \Psi_{h} (k) &=   
    \Psi_g( \Psi_h (k))\\
    &=R_{g^{-1}}L_g( \Psi_h (k))
     = g  \Psi_h (k) g^{-1}= g h k h^{-1}g^{-1}\\
     &=(gh)k (gh)^{-1}=
    \Psi_{gh}(k).
    \end{split}
\] 

For $X \in \fg,$ we see that $(\Ad_g)^{-1}$ needs to satisfy 
\[
    \begin{split}
(\Ad_g) (\Ad_g)^{-1}X&=g [ (\Ad_g)^{-1}X] g^{-1}=X,\\
  (\Ad_g)^{-1} (\Ad_g) X&=(\Ad_g)^{-1} g X g^{-1}=X.
    \end{split}
\]

As a result, similar to the previous case, we have
\[
(\Ad_g)^{-1}= \Ad_{g^{-1}}
\]

The Adjoint composition rule  is therefore given by
\begin{equation}\label{eq:Adchain}
 \Ad_{g} \Ad_{h}(X) = \Ad_{gh}(X)   
\end{equation}
 for any arbitrary $X$ in the Lie algebra $\mathfrak{g}$.
This can be shown by
\[
\Ad_g \Ad_h(X) = g(hXh^{-1} )g^{-1} = (g h)X(gh)^{-1}= \Ad_{gh}X,
\]
which shows that $Ad_{(\cdot)}$ is a homomorphism.

\paragraph*{Adjoint Representations and Translations}
$\Psi_g(\Id)=g\Id g^{-1}=\Id$
implies that any curve $\gamma(s)$ through $\Id$
on $\G$
is mapped by this homomorphism $\Psi_g$ to another curve $g \gamma(s) g^{-1}$ on $\G$ through $\Id$ (see Fig.~\ref{fig:Adjoint}). 

The adjoint representation ($\Ad_g =d \Psi_g (\Id) $) maps any tangent vector $v$ in $\fg=T_{\Id}\G$
 to another tangent vector $Ad_g v=gvg^{-1}$ in $T_{\Id} \G$ (see Fig.~\ref{fig:Adjoint}), while  $L_g$ and $R_g$
 map tangent vectors in $T_{\Id} \G$  to tangent vectors in $T_{g} \G$. 

\subsubsection{Lie Derivative of a Function on a Lie Group}
Let $X \in \fg$, the Lie algebra of the Lie group $\G$, and let $f: \G \rightarrow \Real$ be analytic. 
We define the Lie derivative of a function $f(g)$ 
with respect to $X$ as follows:
\begin{equation} \label{eq:lie_deriv_group}
    \begin{split}
        \Lie_X f(g) &:= \lim_{t \rightarrow 0} \frac{f(g \exp(tX)) -f(g)}{t}\\
        &=  \frac{d}{dt} f  \left (   R_{\exp(tX)} (g ) \right ) \bigg |_{t=0} = \frac{d}{dt}    f(g  \exp(tX)) \bigg |_{t=0} .
    \end{split}
\end{equation}
Note that the definition of Lie derivative in \eqref{eq:lie_deriv_group} is referred to as the {\em right-Lie derivative} in \cite{chirikjian2011stochastic}. This nomenclature comes from the fact that the right translation, $R_{\exp(tX)} $, is utilized in the definition instead of the left-translation.  It is also mentioned in \cite{chirikjian2011stochastic} that the definition in \eqref{eq:lie_deriv_group} is often referred to as the left-Lie derivative, as it is left-invariant, i.e. invariant with respect to a left-translation (more on this latter). As we are interested in utilizing left-invariant Lie derivatives, we will only consider the form in \eqref {eq:lie_deriv_group} as our definition of the Lie derivative. 

\subsection{Special Euclidean Group $SE(3)$}
In this subsection, we focus on the Special Euclidean Group $SE(3)$ as it is widely utilized in robotics applications. Although there are many other well-written robotics references, such as \cite{lynch2017modern}, we will use  \cite{murray1994mathematical} as a reference for this subsection and will attempt to follow its notations closely.
Therefore for example, whereas we have been using $X$ to denote an element of the Lie algebra $\fg$. As we move our attention to the specific $\SE$ group, which is closely connected to  robotics applications, we will henceforth follow the robotics convention of using the hat-map notation to denote elements of the Lie algebra for the rest of the paper, e.g., given $\omega \in \mathbb{R}^3$, then $\hat{\omega} \in \fso$.

\subsubsection{Lie Group Properties of $SE(3)$} \label{sec:SE(3)_as_Lie_group}
\label{sec:LGPSE3}
\begin{figure}[!t]
\centering
\resizebox{\columnwidth}{!}{
\begin{tikzpicture}[scale=1,>=latex,decoration={
    markings,
    mark=at position 0.6 with {\arrow{latex}}}]

\draw[thin,->] (0,0,0) -- (2,0,0) node[anchor=north east]{\(y\)};
\draw[thin,->] (0,0,0) -- (0,2,0) node[anchor=north west]{\(z\)};
\draw[thin,->] (0,0,0) -- (0,0,2) node[anchor=south]{\(x \quad \)};
\node at (0,0,0) [below right] {A};

\coordinate (BOrigin) at (5,2.5,0);
\begin{scope}[shift={(BOrigin)}, rotate around x=30, rotate around y=20]

      
   \draw[fill=gray!30, opacity=0.4, smooth cycle,tension=.7] plot coordinates{(-1,-1) (0,1) (1.5,1) (2.5,1.1) (3.5,-1.1)};
   \draw     [smooth cycle,tension=.7] plot coordinates{(-1,-1) (0,1) (1.5,1) (2.5,1.1) (3.5,-1.1)};

    \draw[thin, ->] (0,0,0) -- (2,0,0) node[anchor=north east]{\(x \)};
    \draw[thin,->] (0,0,0) -- (0,2,0) node[anchor=north west]{\(y\)};
    \draw[thin,->] (0,0,0) -- (0,0,2) node[anchor=south]{\(z \quad \)};
    \node at (0,0,0) [below right] {B};


\filldraw[black] (2.6,-0.5) circle (2pt) node[below right]{$q$};

\end{scope}

\draw[dotted,->] (0,0,0) -- (BOrigin) node[midway,above] {\( p_{ab} \)};

\draw[thin,-{latex}] (4.5,2,-0.1) to [bend left=30] node[midway, below] {\(\quad g_{ab} \)} (1,0.2,0.1);

\end{tikzpicture} }
\caption{Coordinate frames $\{A\}$ and $\{B\}$ for specifying rigid motions.}
\label{fig:SE(3)}
\end{figure}
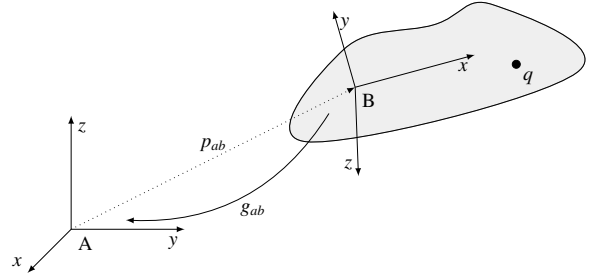

\paragraph*{Rigid Body Motions in $SE(3)$}
Let us consider the position and orientation of a coordinate frame $\{ B\}$, attached to the rigid body relative to an
inertial frame $\{A\}$ (see Fig.~\ref{fig:SE(3)}).

The configuration space of the system is denoted by the special Euclidean group    $SE(3)$, which consists of the pair $(p \in \Real^3, R \in SO(3))$ and is the semi-direct product  
\[
\begin{split}
SE(3) &= \{( p, R) : p\in \Real^3, R\in SO(3) \}\\
 &= \Real^3 \rtimes SO(3).  
\end{split}
\]
An element $g = (p, R)\in SE(3)$ informs  
both a specification of the configuration of a rigid body and a coordinate transformation that converts the coordinates of a point from one frame to another.
{\color{black}
For example, $g_{ab} = (p_{ab}, R_{ab})\in SE(3)$ is the specification of the configuration of
coordinate frame  $\{B\}$ relative to coordinate frame  $\{A\}$ (see Fig.~\ref{fig:SE(3)}).  $R_{ab} \in SO(3)$ are the orthonormal coordinates of frame $\{ B\}$ written relative to frame $\{A\}$ while $p_{ab}$ is the position vector of the origin of frame $\{B\}$ relative to frame $\{A\}$.

We may also write
$g\circ q$ to denote the action of a rigid transformation $g = (p, R) \in SE(3)$ on a point, $g\circ q = p + Rq,$  Thus, denoting the coordinates of a point $q$ relative to frame $\{B\}$ by $q_b$, we can find the coordinates of the same point relative to frame $\{A\}$, $q_a$, by  
\[q_a = g_{ab} \circ q_b = p_{ab} + R_{ab}q_b,\]

\paragraph*{Homogeneous Matrix Representations in 
$SE(3)$}
The $4 \times 4$ matrix 
\[
\bar g = \left[
\begin{array}{cc}
  R    &  p  \\
 0 & 1   
\end{array}
\right], \hspace{2em}R \in SO(3),\hspace{1em} p \in \Real^3
\]
is referred to as  the homogeneous representation of $g = (p, R)\in SE(3)$. By defining the homogeneous coordinates of a point by adding an additional element equal to $1$, we can perform  the previously defined coordinate transformation from frame $\{B\}$  to frame $\{A\}$ by using simple matrix-vector multiplication as follows
\[
\bar q_a:= \left[
\begin{array}{c}
 q_{a}   \\
     1 
\end{array}
\right]
 =
\left[
\begin{array}{cc}
  R_{ab}    &  p_{ab}  \\
 0 & 1   
\end{array}
\right]
 \left[
\begin{array}{ccc}
 q_{b}   \\
     1 
\end{array}
\right]
=:\bar g_{ab} \bar q_b,
\]
where $\bar q_a,\,  \bar{q}_b \in \Real^4$ are the homogeneous coordinates of point $q$ relative to frame $\{A\}$ and $\{B\}$ respectively.

For the remainder of the paper, we will 
 denote the homogeneous matrix representation of $\SE$, $\bar g$ simply as $g$. Similarly, we will also denote the coordinate transformation of a point as $q_a = g_{ab} q_b$.

Using  homogeneous
representations,  it is straightforward to show that the set $SE(3)$ of rigid body transformations constitutes a group:
\begin{itemize}
\item If $g_1, g_2\in SE(3)$, then $g_1g_2\in SE(3)$.
\item The  identity  $ I_4 \in  SE(3)$.
\item If $g\in SE(3)$, then the inverse of $g$ is determined by the 
matrix inversion:
\[
g^{-1}= 
 \left[
\begin{array}{ccc}
 R^T  & -R^T p  \\
    0 & 1 
\end{array}
\right] \in SE(3).
\]
\item The composition rule for rigid body transformations is associative.
\end{itemize}

\subsubsection{Lie Algebra Properties of $\SE$}
\paragraph*{Hat- and Vee-maps in $SE(3)$.}
 The Lie algebra of $SO(3)$, denoted by  $\fso$, defines skew-symmetric matrices.
 The hat-map for $\fso$ is defined by 
$\widehat{(\cdot)}\, (\text{or} \, (\cdot )^{\wedge} ): \Real^3 \rightarrow \fso,$ which maps an angular velocity vector $\omega \in \Real^3$ to its skew-symmetric cross product matrix, i.e.,  
\[\hat \omega=- \hat \omega^T
=
  \left[
\begin{array}{ccc}
0  & -\omega_3 & \omega_2   \\
 \omega_3 & 0 & -\omega_1 \\
  -\omega_2  & \omega_1 & 0
\end{array}
\right]  \in \Real^{3 \times 3} \,.
\]

Hence, for $v \in \Real^3$, $\hat \omega v = \omega \times v$ where $\times$ represents the cross product of vectors. Because of this, the hat map of the vector is often represented with the notation as $\hat{\omega} = [\omega \times]$ in \cite{lynch2017modern}.
The inverse of the hat map is the   vee-map    
 $
 (\cdot )^\vee: \fso : \rightarrow \Real^3.
 $
Similarly, the hat-map for $\fse$ is defined as  $\widehat{(\cdot)}: \Real^6 \rightarrow \fse.$
The inverse of the hat map is the   vee-map    
 $
 (\cdot )^\vee: \fse : \rightarrow \Real^6.
 $
Thus, we have
\[
 \hat \xi = 
\begin{bmatrix}
     \hat \omega  & v  \\
    0 & 0 
\end{bmatrix} \in \fse \subset \Real^{4 \times 4}, \;\;
\forall  \xi = 
\begin{bmatrix}
  v  \\
    \omega
\end{bmatrix} \in \Real^6,
\]
where $v, \omega \in \Real^3, \hat \omega \in \fso.$

\paragraph*{Twists}
An element of $\fse$ is referred to as a twist or a (infinitesimal) generator
of the Euclidean group.  Recall that the adjoint action  of $g \in \SE$ on $\hat{\xi} \in \fse$, $\Ad_g: SE(3) \times \fse \rightarrow \fse$, is then defined by
\[
\Ad_g \hat{\xi} = g \hat{\xi} g^{-1}.
\]
Let $\hat{\xi}' = g \hat{\xi} g^{-1}$. $\hat{\xi}'$ is $\xi$ applied by a constant rigid motion $g$ with $\dot{g}=0$.
With the previous notion of $\Ad_g:\SE \times \mathbb{R}^6 \to \mathbb{R}^6$ as an linear operator, we can write:
\[
    \hat{\xi}' = g\hat{\xi}g^{-1}, \quad \xi' = \Ad_g \xi.
\]
We can interpret $\hat{\xi}$ as a twist with the coordinate $\xi \in \mathbb{R}^6$, and $\hat{\xi}'$ a twist with the coordinate $\Ad_g \xi \in \mathbb{R}^6$, $\forall g \in \SE$.




Consider a curve parameterized by time,  representing a trajectory of
a rigid body such that
\[
g_{ab}(t)= \left[
\begin{array}{cc}
     R_{ab}(t) & p_{ab}(t)  \\
      0 &   1
\end{array}
\right]
\]
Now, we define the spatial velocity $\widehat{V}_{ab}^s$ of a rigid motion $g_{ab} \in SE(3)$
\[
\widehat{V}_{ab}^s:= \dot g_{ab} g_{ab}^{-1}=
\left[
\begin{array}{cc}
\dot R_{ab}   R_{ab}^T &  -\dot R_{ab}   R_{ab}^T p_{ab} +  \dot  p_{ab} \\
  0 & 0
  \end{array}
\right],
\] 
\[
{V}_{ab}^s=\left[
\begin{array}{c}
 v_{ab}^s \\
   \omega_{ab}^s  
\end{array}
\right]
=
\left[
\begin{array}{c}
  -\dot R_{ab}   R_{ab}^T p_{ab} +  \dot  p_{ab} \\
      ( \dot R_{ab} R_{ab}^T)^\vee
\end{array}
\right]
\] 
We also define the body (or generalized) velocity of a rigid motion $g_{ab} \in SE(3)$
\[
\begin{split}
\widehat{V}_{ab}^b:= g_{ab}^{-1} \dot g_{ab} &=
\left[
\begin{array}{cc}
 R_{ab}^T  \dot R_{ab} &   R_{ab}^T    \dot p_{ab}  \\
  0 & 0
  \end{array}
\right]\\
&=
\left[
\begin{array}{cc}
\widehat \omega_{ab} &   v_{ab}  \\
  0 & 0
  \end{array}
\right],
\end{split}
\] 
where $\widehat \omega_{ab}^b =R_{ab}^T  \dot R_{ab} $ and $v_{ab}^b=R_{ab}^T    \dot p_{ab}$.
\[
{V}_{ab}^b=\left[
\begin{array}{c}
 v_{ab}^b \\
   \omega_{ab}^b  
\end{array}
\right]
=
\left[
\begin{array}{c}
  R_{ab}^T    \dot  p_{ab} \\
      (  R_{ab}^T \dot R_{ab})^\vee
\end{array}
\right]
\] 

Body velocity can be transformed into the spatial velocity by the adjoint transformation $Ad_g$, i.e.,
\[
{V}_{ab}^s=
\left[
\begin{array}{c}
 v_{ab}^s\\
   \omega_{ab}^s  
\end{array}
\right]
=
\underbrace{
\left[
\begin{array}{cc}
 R_{ab} & \widehat{p}_{ab} R_{ab}\\
    0 & R_{ab}
    \end{array}
\right]}_{=\Ad_{g_{ab}}}
\left [ V_{ab}^b=
\left[
\begin{array}{c}
 v_{ab}^b \\
   \omega_{ab}^b  
\end{array}
\right]
\right ] \] 
\[
i.e., \,
 {V}_{ab}^s=\Ad_{g_{ab}}{V}_{ab}^b. 
\]

Dropping the subscript $ab$, we can write the (large) adjoint transformation $\Ad_g : \Real^6 \rightarrow \Real^6$ which maps one coordinate to another as:
\[
\Ad_g = \begin{bmatrix}
R & \widehat{p} R\\
    0 & R
\end{bmatrix}, \quad g \in \SE
\]
$\Ad_g$ is invertible, and its inverse is given by
\[
\Ad_g^{-1}=\left[
\begin{array}{cc}
R^T & -R^T\widehat{p} \\
    0 & R^T
    \end{array}
\right]
=\Ad_{g^{-1}}.
\]

The body and spatial velocities are related by the (large) adjoint transformation.
\[
V^s=\Ad_g V^b, \quad  V^b=\Ad_{g}^{-1} V^s=\Ad_{g^{-1}} V^s.
\]

As mentioned previously, the large adjoint map used herein is a linear operator version. 



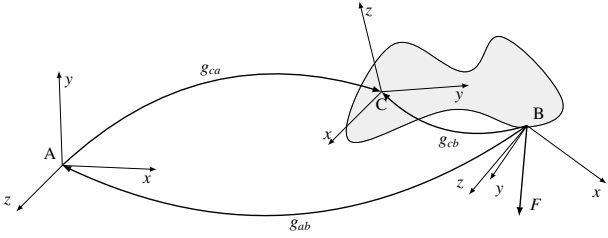
\begin{figure}[!t]
\centering
\resizebox{\columnwidth}{!}{
\begin{tikzpicture}[scale=1,>=latex,decoration={
    markings,
    mark=at position 0.6 with {\arrow{latex}}}]

\coordinate (COrigin) at (4,2.5,0);
\begin{scope}[shift={(COrigin)}, rotate around x=10, rotate around y=-10, rotate around z=10]
    \draw[fill=gray!30, opacity=0.4, smooth cycle,tension=.7] plot coordinates{(-1,-1) (0.5,1) (2,0.3) (3,1) (4,-0.5) (3,-1) (1.5, -0.5)};
     \draw[ smooth cycle,tension=.7] plot coordinates{(-1,-1)  (0.5,1) (2,0.3) (3,1) (4,-0.5) (3,-1) (1.5, -0.5)};
      \draw[thin,->] (0,0,0) -- (2,0,0) node[anchor=north east]{\(y\)};
    \draw[thin,->] (0,0,0) -- (0,2,0) node[anchor=north west]{\(z\)};
    \draw[thin,->] (0,0,0) -- (0,0,2) node[anchor=south]{\(x\)};
    \node at (0,0,0) [below ] {C};

\end{scope}
    
\coordinate (BOrigin) at (7,1.85,0.1);
\begin{scope}[shift={(BOrigin)}, rotate around x=100, rotate around y=-40, rotate around z=0]
    \draw[thin,->] (0,0,0) -- (2,0,0) node[anchor=north east]{\(x\)};
    \draw[thin,->] (0,0,0) -- (0,2,0) node[anchor=north west]{\(y\)};
    \draw[thin,->] (0,0,0) -- (0,0,2) node[anchor=south]{\(z \quad \)};
    \draw[thick,->] (0,0,0) -- (1,1,1) node[anchor=south]{\(\quad \quad F\)};
\node at (0,0,0) [above right ] {B};
\end{scope}

\coordinate (AOrigin) at (-2.5,1,0);
\begin{scope}[shift={(AOrigin)}, rotate around x=5, rotate around y=-5, rotate around z=0]
    \draw[thin,->] (0,0,0) -- (2,0,0) node[anchor=north east]{\(x\)};
    \draw[thin,->] (0,0,0) -- (0,2,0) node[anchor=north west]{\(y\)};
    \draw[thin,->] (0,0,0) -- (0,0,2) node[anchor=south]{\(z \quad \)};
\node at (0,0,0) [above left ] {A};
\end{scope}


\draw[thick,-{latex}] (BOrigin)  to [bend left=30] node[midway, below] {\( g_{cb} \)}  (COrigin);
\draw[thick,-{latex}] (BOrigin)  to [bend left=30] node[midway, below] {\( g_{ab} \)}  (AOrigin);
\draw[thick,-{latex}] (AOrigin)  to [bend left=30] node[midway, above] {\( g_{ca} \)}  (COrigin);

\end{tikzpicture}}
\caption{Transformation of wrench $F$ between coordinate frames.}
\label{fig:Wrench}
\end{figure}


\paragraph*{Dual of the Lie algebra} 
The co-adjoint    action of $g$ (or sometimes used with $g^{-1}$) on  $\xi^* \in \fse^*$, which is the dual of $\xi$, is denoted as $\Ad_{g^{-1}}^* : \SE \times  \fse^* \rightarrow \fse^*$. Similar to the large adjoint operator of twist, we will not distinguish the linear operator version of $\Ad_{g}$, from its representation defined by a matrix in $\Real^{6 \times 6}$ 
\[
\Ad_{g^{-1}}^* =\Ad_{g^{-1}}^T.
\]

The dual of the Lie algebra $\fg$ of $g$, can be defined.
Let $\Ad_g^*: G \times \fg^* \rightarrow \fg^*$ be the dual of $\Ad_g$ defined by
\[
\langle \Ad_{g^{-1}}^* (\alpha), \xi \rangle= \langle \alpha, \Ad_{g} (\xi) \rangle 
\] 
for $\alpha \in \fg^*$ and $\xi \in \fg$. The co-adjoint action of $g$ on $\fg^*$ is then given by the map $\Phi^*: G \times \fg^* \rightarrow \fg^* $
\[
\Phi^* (g, \alpha) =\Ad_{g^{-1}}^* \alpha.
\]

\paragraph*{Wrenches}
A pair of a force and a moment acting on a rigid body 
is referred to as a wrench, i.e., 
a generalized force.
\[
F= 
\left[
\begin{array}{c}
  f  \\
  \tau    
\end{array}
\right] \in \mathbb{R}^6,
\]
 where $f \in \Real^3$ and $\tau \in \Real^3$ are linear and rotational force components, respectively.

Consider the motion of a rigid body by $g_{ab}(t)$ with respect to an inertial frame A and a body frame B.
Let $V_{ab}^b \in \Real^6$ and $F_b\in \Real^6$ be the instantaneous body velocity and an applied wrench.
The infinitesimal work is obtained by the inner product of $V_{ab}^b \in \Real^6$ and $F_b$.
\[
\delta W=   \langle V_{ab}^b,   F_b \rangle=({v_{ab}^b}^T f+ {\omega_{ab}^b} ^T \tau).
\]
The net work generated by applying $F_b$ through  a twist $V_{ab}^b$ during the  interval $[t_1, \, t_2]$ is given by
\[
W=\int_{t_1}^{t_2}  \langle V_{ab}^b, F_b  \rangle dt.
\]

Let $g_{bc} =(p_{bc},R_{bc})$ be the configuration of frame C relative to B.
To find an equivalent wrench $F_c$ applied at C, we equate two instantaneous work quantities by $F_c$ and $F_b$.
\[
\begin{split}
  \langle V_{ac}, F_c \rangle&=\langle  V_{ab}, F_b \rangle\\
  &= (\Ad_{g_{bc}}V_{ac})^T F_b = \langle V_{ac} , \Ad_{g_{bc}}^T F_b \rangle,
\end{split}
\]
where we use the relationship between body velocities, i.e., $V_{ab}=\Ad_{g_{bc}} V_{ac}.$

The coordinate transformation between wrenches $F_c$ and $F_b$  (see Fig.~\ref{fig:Wrench}) can be represented by different ways.
\[
\begin{split}
F_c &= \Ad_{g_{bc}}^T F_b=\Ad_{g_{ba}g_{ac}}^T F_b\\
&=(\Ad_{g_{ba}}\Ad_{g_{ac}} )^T F_b=\Ad_{g_{ac}}^T\Ad_{g_{ba}}^T F_b\\
    &=\Ad_{g_{cb}^{-1}}^T F_b=\Ad_{(g_{ca}g_{ab})^{-1}}^T F_b =\Ad_{ g_{ab}^{-1} g_{ca}^{-1} }^T F_b\\
    &= (\Ad_{ g_{ab}^{-1}} \Ad_{g_{ca}^{-1} } )^T F_b=  \Ad_{g_{ca}^{-1} }^T  \Ad_{ g_{ab}^{-1}}^T F_b,
\end{split}
\]
where the adjoint composition  rule in \eqref{eq:Adchain} was used.

\paragraph*{Exponential Map from $\fse$ to $SE(3)$.}
Given $\hat \xi \in \fse$ and $\theta \in \Real$, the exponential of $\hat \xi \theta$ is in $SE(3)$, i.e., 
\[
\begin{split}
    \exp{(\hat \xi \theta)} &= \exp{   \left (\left[
\begin{array}{ccc}
 \hat \omega  & v  \\
    0 & 0 
\end{array}
\right]   \theta  \right ) }\in SE(3)\\
&=I+ {    \left[
\begin{array}{ccc}
 \hat \omega  & v  \\
    0 & 0 
\end{array}
\right]   \theta    } + { \frac{1}{2!}   \left[
\begin{array}{ccc}
 \hat \omega^2  & \hat \omega v  \\
    0 & 0 
\end{array}
\right]   \theta^2    } \\ &+ { \frac{1}{3!}   \left[
\begin{array}{ccc}
 \hat \omega^3  & \hat \omega^2 v  \\
    0 & 0 
\end{array}
\right]   \theta^3    }+ \cdots
\end{split}
\] 
This shows that the exponential map from $\fse$ to $SE(3)$ is the matrix exponential on a linear combination of the
generators.
We interpret the transformation $g=\exp(\hat \xi \theta)$ 
as a mapping  from an initial coordinate, $p(0)\in \Real^3$, to its
coordinate $p(\theta)$ after the rigid motion is applied such that
\[
p(\theta)= \exp{(\hat \xi \theta)} p(0).
\]
Let $g_{ab}(0)$ represent the initial configuration of a rigid body relative to a frame A, then the final configuration,
 with respect to A, is given by
\[
g_{ab}(\theta)= \exp{(\hat \xi \theta)} g_{ab}(0).
\]
The exponential map for a twist gives the relative motion of a rigid
body.

\subsubsection{Forward Kinematics}
The forward kinematics of the manipulator is given by the product of the exponentials formula, i.e., a map $g: Q \rightarrow SE(3):$
\[
g(\F)=e^{(\hat \xi_1 \theta_1)}e^{(\hat \xi_2 \theta_2)} \cdots e^{(\hat \xi_n \theta_n)} g(0),
\]
where $e^{(\cdot)}=\exp(\cdot)$ and $g(\theta)$ is the homogeneous representation of the end-effector in the spatial frame, $\xi_i \in \Real^6$ is a twist represented in the spatial frame and $\theta_i \in \Real$ a joint angle of the $i$-th joint, for $i=1, 2, \cdots n$, starting from the base. See more details in \cite{murray1994mathematical}.


\section{Equivariant Deep Learning}\label{sec:Equivariant}
Equivariant deep learning has been primarily employed to process visual inputs such as 2D images or 3D point clouds in robotic applications.
In this chapter, we will introduce and review the concept of equivariance and equivariant neural network architectures to handle such vision inputs. We start by introducing the equivariant structure for the convolutional neural networks (CNN) to deal with image domain data, followed by the equivariant group convolutional neural networks (G-CNNs) for point cloud data. For a more detailed review of these concepts, the readers are referred to \cite{veefkind2024probabilistic, UvA_website}.
\subsection{Regular Group CNNs}
\subsubsection{Standard CNNs: Cross-correlation Kernels and Weight-sharing}
Before introducing the equivariant group CNNs (G-CNNs),  let us first consider the standard CNNs, which use cross-correlation (or convolution) kernels to extract features from images. 
\paragraph*{Cross-correlation Kernels} The cross-correlation (or convolution) operation between an input image $f(x)$ and a correlation kernel $\W(x)$ (or a convolution kernel $\hat{k}(x)$, respectively) at   $x$ can be written as:
\begin{equation}\label{eq:conv}
\begin{split}
y(x) &= (\W \star f )(x) = \int_{\mathbb{R}^n} f( \tilde  x)k(\tilde x-x)d \tilde  x\\
&= (\hat{k} \ast f) (x) =  \int_{\mathbb{R}^n} f( \tilde  x) \hat{k}(x- \tilde x)d \tilde  x,
\end{split} 
\end{equation}
where $\W(x)= \hat{k}(-x)$, $\star$ denotes cross-correlation operator, $\ast$ denotes convolution operator and $\tilde x$ is the dummy variable of integration.
An example of a function $f (\cdot)$ in vision processing applications is a $256 \times 256$ RGB image, where $f(x) \in \Real^3$ is sampled on a discrete pixel grid $\X = \{ 0, 1, \cdots , 255 \} \times \{ 0, 1, \cdots , 255 \} \subset \Real^2$ and $f(x)$ represents  RGB values at each discrete pixel location $x = (x, y) \in \X$.

As can be seen in Fig.~\ref{fig:Alex}, many redundant rotated edge filters in different angles are generated in the standard CNNs (by training under image augmentation via rotation). 
Interpreting standard CNNs as cross-correlation kernels allows us a straightforward extension to the group equivariant CNNs. 
Henceforth, we will develop our discussion using the cross-correlation kernel $k(x)$, but it could be straightforwardly generalized to convolution.

\begin{figure}
    \centering
    \includegraphics[width=1.0\columnwidth]{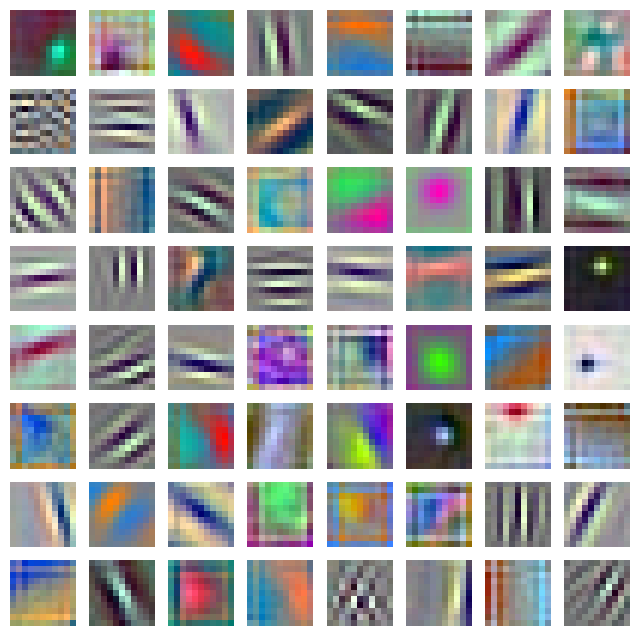}
    \caption{Convolutional filters in the first layer of AlexNet\cite{krizhevsky2012imagenet} trained on ImageNet \cite{deng2009imagenet}. Many rotated edge filters with similar shapes can be spotted. This standard CNN is not efficient without weight sharing in the rotational angle. This figure is generated using pre-trained AlexNet provided by PyTorch \cite{PytorchAlexNet}.
    }
    \label{fig:Alex}
\end{figure}

\paragraph*{Left-regular Representation} 
Before we further investigate the cross-correlation kernels 
undergoing group action, one notable concept must be reviewed: left-regular representation. A left-regular representation is a useful way to let a group $\G$ act on functions. 
In this case, a function is transformed by transforming the function's domain $\mathcal{X}$ by the action of the group $\G$.
The left-regular representation of $g \in \G$ acting on $f(x)$ with $x \in \mathcal{X}$ is defined by
\begin{equation}\label{eq:left}
L_g(f)(x) = g \circ  f(x)= f(g^{-1} \circ x)
\end{equation}
where $\circ$ denotes the action of the group $\G$ on the appropriate domain or codomain. In the case of $\mathcal{X} = \mathbb{R}^2$, i.e., an image plane, $f \in \mathbb{L}^2(\mathbb{R}^2)$ 
is a square-integrable function.
We also notice that
\begin{equation}\label{eq:g1g2}
\begin{split}
L_{g_1g_2}(f)(x) &= g_1g_2 \circ  f(x)=L_{g_1}L_{g_2}(f)(x)\\
&= f((g_1g_2)^{-1} \circ x)=f(g_2^{-1}g_1^{-1} \circ x).
\end{split}
\end{equation}

\paragraph*{Weight-sharing}\label{weightsharing}

We now define the left-regular representation of the translation group $(\mathbb{R}^n,+)$  acting on a correlation kernel as in \eqref{eq:left}  
\[ \T_x(\W) (\tilde x):= L_x(\W)(\tilde x) = \W(\tilde x-x),\] 
\eqref{eq:conv} can then be represented as an inner product. 
\[
\begin{split}
y(x) = (\W \star f )(x) &= \int_{\mathbb{R}^n} f(\tilde x)\W(\tilde x-x)d \tilde x\\
&= \int_{\mathbb{R}^n} f(\tilde x)\T_x(\W)(\tilde x) d \tilde x\\
&=\langle \T_x (\W), f\rangle_{\mathbb{L}_2 (\mathbb{R}^n)}.
\end{split}
\]
From the inner product form above, it is clear that the convolution  utilizes the kernel $k(\cdot)$ in template matching for feature extraction over different locations in $x$. This leads to efficient learning via sharing the same kernel weights across various locations in the data. 
Standard  CNNs are effective for data that exhibits translation symmetry. However, they may not efficiently learn features such as  edge shapes with different angles without weight sharing in rotational angles. In fact, for the standard CNNs, data augmentation  such as rotating images with different angles are used. As a result, CNNs generate a larger number of kernels (or filters) with various orientations, as in  Fig.~\ref{fig:Alex}. 
We will show how to 
substitute the original CNN kernels with ones that transform according to left-regular group representation in order to produce group-equivariant CNNs, in the following sections.

\subsubsection{$SE(2)$ Regular Group CNNs}


To incorporate weight sharing under 2D rotations, G-CNNs \cite{cohen2016group} introduced the concept of convolution over the 2D \textit{roto-translation group}, i.e., $SE(2)$. This approach involves two key steps: 1) Lifting convolution and 2) Group convolution. Lifting convolution first maps the input image from the $\mathbb{R}^2$ domain to the group domain $\mathbb{G}$, followed by a group convolution performing the convolution directly on the group domain.
The following paragraphs provide a more detailed explanation of these processes.

\paragraph*{$SE(2)$ Lifting Correlations} 
We now consider a transformation operator in $SE(2)$, $L_g$ with the roto-translation parameter $g=(x,\theta)$, where $x$ and $\theta$ are the translation vector and the rotation angle, respectively. Due to \eqref{eq:g1g2} with $g_1=(x)$ and $g_2=(\theta)$, $L_g=L_{(x,\theta)}$ can be obtained as follows. 
\[
\begin{split}
[ L_{g_1g_2} \W ](\tilde x)&=[ L_{(x,\theta)} \W ](\tilde x) =[L_x L_\theta \W ](\tilde x)\\ &=\W (g_2^{-1} g_1^{-1} \tilde x)) = \W (R^{-1}_\theta (\tilde x - x)).
\end{split}
\]
The roto-translation lifting correlation from $\Real^2$ to the space of $SE(2)$ can be used to produce a lifted  feature in $SE(2)$, i.e., $f_{SE(2)}$. 
\begin{equation}\label{eq:LiftingCov}
\begin{split}
f_{SE(2)}(g)&=y_{SE(2)}(x, \theta)=(\W \star_{SE(2)}  f )  (x, \theta)\\
&=  \langle L_{(x, \theta)}\W, f \rangle_{\mathbb{L}_2 (\Real^2)} =\langle L_{x} L_\theta \W, f \rangle_{\mathbb{L}_2 (\Real^2)} \\
&=\int_{\Real^2} \W (R^{-1}_{\theta} (\tilde x-x)) f(\tilde x) d \tilde x.
\end{split}
\end{equation}
where $\star_{SE(2)}$ denotes a lifting cross-correlation to $SE(2)$ domain and $R_\theta \in SO(2)$ is defined in \eqref{eq:RtheSO2}.

\paragraph*{$SE(2)$ Regular Group Correlations} $SE(2)$ group convolutions are roto-translation equivariant. 
Following the initial roto-translation lifting correlation on a 2D image, a feature map is generated on   $SE(2)$. The feature layers resulting from the lifting operation are then parameterized by patterns of position-orientation localized features, which can be transformed by the action of the group element $L_{(x, \theta)}$. This transformation enables the use of $SE(2)$-equivariant G-CNNs that operate on $SE(2)$ kernels, thereby allowing for the efficient extraction of features that are invariant to both translation and rotation.

After the $SE(2)$ lifting correlation, the $SE(2)$ group correlation can be performed on the higher-dimensional lift feature map of $SE(2)$.
\begin{equation}\label{eq:SEfilter}
    \begin{split}
    y(g) &= 
     \int_{SE(2)} \W (g^{-1} \cdot \tilde g) f_{SE(2)}(\tilde g) d \tilde g\\
    &= \int_{\Real^2} \int_{S^1} \W(R^{-1}_\theta (\tilde x- x),  \tilde \theta -\theta) f_{SE(2)}(\tilde x, \tilde \theta) d \tilde x d \tilde \theta.
    \end{split}
\end{equation}

In $SE(2)$ regular G-CNNs, full equivariance to the continuous $SE(2)$ group is not directly achieved. Instead, the $SE(2)$ is approximated by the finite subgroup $\mathbb{R}^2 \rtimes C_N$, where $C_N$ represents a cyclic group of order $N$, which discretizes planar rotations into $N$ evenly spaced angles. While this discretization provides a practical approximation of planar rotation and easy implementation from off-the-shelf packages, the computational complexity grows exponentially with increasing dimensions. Hence, discretization poses significant challenges for higher-dimensional rotations, such as $SO(3)$.

\subsubsection{Equivariant Maps} Suppose $\G$ is a Lie group, and $M$ and
$N$ are both smooth manifolds with smooth left or right $\G$-actions. A map (e.g., a neural network)
$\Phi : M \rightarrow N$ is said to be equivariant with respect to   $\G$-actions   if for each
$g \in \G$, 
\[
\Phi(g \circ p)=g \circ \Phi(p),\quad \text{for left actions},\]
and 
\[
\Phi(p \circ g)= \Phi(p) \circ g, \quad  \text{for right actions \cite{leesmooth}.} \] 
Equivariance implies that no information is lost by input transformations, i.e., 
the output will be shifted to different locations in the output domain that correspond to the input transformation. It also allows for weight sharing over the transformations in the
group (e.g., $SE(2)$ or $\SE$ including rotation), which reduces a significant number of learnable parameters as compared to the standard CNNs (e.g., see Fig.~\ref{fig:Alex}). 
For example, $SE(2)$ equivariant CNNs (i.e., G-CNNs) for 2D images can be built using lifting and group convolutions from \eqref{eq:LiftingCov} and \eqref{eq:SEfilter}, respectively. Although the regular G-CNNs are more efficient due to weight sharing compared to the standard CNNs,  the  filtered  outputs in discretized $\theta$
need to be computed and stored for the higher dimensional feature space in $SE(2)$. Steerable filters can be used to mitigate these computational and memory burdens.

\subsection{Steerable Group CNNs} 
More efficient equivariant convolutional neural networks (i.e., steerable G-CNNs) can be constructed using 
steerable feature vectors, which will remove the discretization in rotation as in G-CNNs with the $SE(2)$ regular group convolutions. 
Steerability of a vector (say $h$) implies that for a transformation group $g$, the vector $h$ transforms via matrix-vector multiplication $D(g)h$. 
Steerable methods avoid discretization of $SO(n)$ in regular G-CNNs by exploiting irreducible representations, which allows the use of Fourier theory on Lie groups. Steerable methods build on the following concepts: steerable feature vectors, spherical harmonics, Wigner-D matrices, and the Clebsch-Gordan tensor product. We first introduce spherical harmonics, as they play a crucial role in the formulation of steerable G-CNN. Next, Wigner-D matrices will be introduced, and we will show that the Wigner-D matrices are a representation of spherical harmonics and, thereby, steerable vectors/features can be formulated together with spherical harmonics and Wigner-D matrices. Equipped with these, the steerable group convolution will be defined. Finally, the Clebsch-Gordan Tensor product will be introduced, which is used to build the steerable linear layers for deep learning.

\subsubsection{Spherical Harmonics (SH)}
Spherical harmonics (SH) provide a theoretical foundation for the steerable functions as they serve as a basis function for $\SO$ transformation. SH is a class of functions on the sphere $S^2$ and can be used as a Fourier basis on the sphere. As will be mentioned later, they are steered by Wigner-D matrices and allow the interpretation of steerable vectors as functions on $S^2$. Moreover, spherical harmonics allow the embedding of 3D displacement vectors into arbitrarily large steerable vectors, i.e., allows lifting. 
Although the theory of SH is well established among physics/chemistry fields, those with engineering backgrounds might be unfamiliar with this concept. Therefore, we present the motivations and formulation of SH in Appendix~\ref{Appendix:SH} for readers who want to investigate further. 

In a nutshell, a spherical harmonic function $Y_m^l : S^2 \to \mathbb{C}$ is given as follows:
\begin{equation} \label{eq:sol_Y}
    Y_m^l(\theta,\phi) = N e^{im\phi} P_m^l(\cos{\theta}),
\end{equation}
where $\theta$ and $\phi$ are the azimuthal and longitudinal angles respectively (see (Fig. \ref{fig:change_of_coordinates}), and $P_m^l$ are associated Legendre polynomials. 
$l\in\{0,1,2,\cdots\}$ is called the \textit{degree} and $m\in\{-l,\cdots,l\}$ is called the \textit{order} of the spherical harmonic function. 
Although we have built the SH in the complex domain, general applications of SH are built in the real domain, and the interchange between the two domains is possible, e.g., \cite{lee2018real}.
Here, the \textit{(real) spherical harmonics} $Y_m^{l}(\theta, \phi)$ are orthonormal functions that form the complete basis of the Hilbert space on the sphere $S^2$. 

Spherical harmonic functions span the sub-representation of $V_l$ vector spaces:
\begin{equation}
    \begin{split}
        V_0 &= \text{Span}\{Y_0^0(n)\}\\
        V_1 &= \text{Span}\{Y_{-1}^{1}(n), Y_0^1(n), Y_1^{1}(n)\} \\
        &\vdots \\
        V_k &= \text{Span}\{Y_{-k}^{k}(n), Y^k_{-k+1}(n), \cdots , Y^k_0(n), \cdots, Y^k_k(n)\},
    \end{split}
\end{equation}
where $n$ is a unit vector on a unit sphere and is a one-to-one correspondence to the $(\theta,\phi)$ representation. We call the $V_l$'s vector space as a \textit{type-$l$} (or \textit{spin-$l$}) vector field and an arbitrary vector $v \in V_l$ as type-$l$ vector. Type-$l$ vectors are identical to themselves when they are rotated by $\theta = 2\pi/l$. Type-$0$ vectors, or \textit{scalars} are invariant to rotations ($\theta=\infty$). Type-$1$ vectors are the familiar 3-dimensional space vectors ($\theta=2\pi$). 

A critical property of  SH is that they are equivariantly transformed through Wigner-D matrix representations of the  rotation matrix $R \in \SO$ as detailed below. 

Consider the $2l+1$-dimensional vector field, ( i.e., a type-$l$ vector field):
\begin{equation}
    \mathbf{Y}^l = \left(Y_{-l}^{l}, Y_{-l+1}^l, \cdots, Y^l_{l-1}, Y^l_l\right)^T.
\end{equation}
Then, $\mathbf{Y}^l$ is equivariant in the unit sphere $S^2$ relative to $R \in SO(3)$ as folllows
\begin{equation}
    \mathbf{Y}^l\left(R \frac{x}{\|x\|}\right) = D^l(R) \mathbf{Y}^l\left(\frac{x}{\|x\|}\right),
\end{equation}
where $D^l(R)$ is a Wigner-D matrix representation of $R\in SO(3)$ of order $l$.

\subsubsection{Wigner-D Matrices as Orthogonal Basis of $SO(3)$}
 $D^{l}_{mn} : SO(3) \rightarrow \Real$, the $mn$-element of $D^l$, form complete orthogonal basis for functions on $SO(3)$.
Therefore, any function $h$ on $SO(3)$  can be represented in terms of a Fourier series of Wigner-D Matrices.
\begin{equation} \label{eq:fourier_transform_on_function}
\begin{split}
h(R) &=\sum_{l}^\infty  \sum_{m=-l}^{l} \sum_{n=-l}^{l} \hat h_{mn}^{(l)}D_{mn}^{l} (R) 
= \sum_l^\infty \tr \left( \hat h^{(l)} (D^{l} (R))^T\right)\\
&=\sum_l^\infty \tr \left ( \hat h^{(l)} D^{l} (R^{-1})\right )
\end{split}
\end{equation}
where $R \in \SO$, and the equation above can be considered as the inverse Fourier transform. Note that we have utilized the fact that $D^l(R^{-1}) = D^l(R^{T})= (D^l(R))^T$.
Because of the orthogonality of the Wigner-D matrices, the Fourier transform coefficients of a function $h(R)$ on $SO(3)$ can be then obtained by the inner product using the basis functions as follows:
\[
\hat h^{(l)}_{mn}=\langle D_{mn}^l(R), h(R) \rangle = \int_{SO(3)} h(R) D_{mn}^l(R) dR.
\]

\subsubsection{Wigner-D matrices as an $SO(3)$ Group Representation}
For the $SO(3)$ group, any representation $ {D}( {R})$ for $ {R}\in SO(3)$ can be reduced into a direct sum of $(2l+1)\times(2l+1)$ dimensional irreducible representations $ {D}^{l}( {R})$ of \textit{degree} $l\in\{0,1,2,\cdots\}$ such that
 \begin{equation}
      {D}( {R})= {U}\left[\bigoplus_{l=1}^{N}{ {D}^{l}( {R})}\right] {U}^{-1}\qquad \forall  {R}\in SO(3)
 \end{equation}
 where $\bigoplus$ denotes a direct sum\footnote{The direct sum can be intuitively understood as a concatenation/block-concatenation for vectors/matrices.}.
 Although there are infinitely many equivalent representations for such $ {D}^l$, a particularly preferred choice is one with a \textit{real basis}\footnote{Another commonly used choice of basis is the \textit{spherical basis} whose representations are unitary.}. On this basis, all the representations $ {D}^l$ are orthogonal matrices. These matrices are called the \textit{(real) Wigner D-matrices} \cite{aubert2013alternative, se3t, thomas2018tensor}.

\subsubsection{Steerable Function and Vectors}
\paragraph*{Steerable Function}
A function $\W$, e.g., a group convolution, is steerable under a transformation group $\G$ if any transformation $g \in \G$ on $\W$ can be represented as a linear combination of a fixed, finite set of basis functions $\{ \phi_i\}$.
\[
{L}_g \W= \sum_{i=1}^n \alpha_i (g) \phi_i = \alpha^T(g) \Phi,
\]
where ${L}_g f$ is the left-regular representation of the group $G$ that transforms    $f$.
The concept of steerability was first coined in the computer vision community \cite{freeman1991design}.
Steerable neural networks work with feature vectors consisting of steerable feature vectors.

\paragraph*{Steerable Feature Vectors}
Steerable feature vectors are vectors that transform via Wigner-D matrices \cite{aubert2013alternative, se3t, thomas2018tensor}, which are representations of $SO(3)$ and $SU(2)$. Wigner-D matrices are used to define any representation and are irreducible. A type-$l$ steerable vector space $V_l$ is a $(2l+1)$-dimensional vector space that is acted upon by a Wigner-D matrix of order $l$. For example, if $u$ is a type-$3$ vector in $V_3$, then it is transformed by $g \in SO(3)$ via  $u \rightarrow D^3(g)u$.
As mentioned above, SH is used to formulate a type-$l$ vector and is steerable via the Wigner-D matrix. 

Any vector $x \in \Real^3$ can be converted into a type-$l$ vector by the evaluation of spherical harmonics $Y_m^{l}: S^2 \rightarrow \Real$ at $n = \frac{x}{ \| x\|}$. In particular, for any $x \in \Real^3$, $\mathbf{Y}^l$ is a type-$l$ steerable vector. 
\begin{equation*}
    \mathbf{Y}^l = \left(Y_m^l\left(\frac{x}{\|x\|}\right) \right)^T_{m = -l, -l+1, \dots, l}
\end{equation*}

Consider a function in terms of spherical bases, with {\it coordinates} $a^l = [a_0^l, \cdots, a_l^l]^T$:
\begin{equation*}
    \W_a = \sum_{l=0}^\infty \sum_{m=-l}^l a_m^l Y_m^l = \sum_{l=0}^\infty (\mathbf{a}^l)^T \mathbf{Y}^l,
\end{equation*}
where $\mathbf{a}^l = (a_{-l}^l, a_{-l+1}^l, \cdots, a_{l-1}^l, a_l^l)^T$.
It is steerable since we can show that  \cite{brandstetter2021geometric}
\begin{equation}\label{eq:Lgf}
\begin{split}
    L_g \W_a(n) &=  \W_{D(g)a} (n) \\&=\sum_{i=1}^n \alpha_i (g) \phi_i = \alpha^T(g) \Phi=(D(g) a)^T \Phi,
\end{split}
\end{equation}
 where $\alpha(g)=D(g) a$ and $\Phi= (Y_0^{0}, Y_{-1}^{1}, Y_0^{1}, \cdots)^T$. Alternatively,
\begin{equation}
    \begin{split}
        L_g \W_a(n) &= \W_a(g^{-1}  n) \\
        &= \sum_{l=0}^\infty (\mathbf{a}^l )^T D^l(g^{-1})\mathbf{Y}^l = \sum_{l=0}^\infty (\mathbf{a}^l )^T D^l(g)^T\mathbf{Y}^l\\
        &= \sum_{l=0}^\infty (D^l(g) \mathbf{a}^l)^T \mathbf{Y}^l
    \end{split}
\end{equation}
where the definition of left-regular representation is used, and the property of the Wigner-D matrix, i.e., $D^l(g^{-1}) = D^l(g)^T$, for $g\in \SO$ is utilized \cite{lee2018real}.  $D(g)$ in \eqref{eq:Lgf} can be understood as block diagonal stack of $D^l(g)$, i.e., $D(g) = \text{blkdiag}(D^0(g), D^1(g), D^2(g), \cdots)$.

\subsubsection{Steerable Group Convolution}
 Consider an input feature map $f: \Real^3 \rightarrow \Real$.
In many applications, we want to perform template matching for arbitrary translations and rotations. We can then   use the roto-translation lifting correlation from $\Real^3$ (i.e., the original  feature map input space) to the space of $SE(3)$ with $g=(x,R)$ to produce the lifted feature map $f_{\SE}$ such as 
 \[
 \begin{split}
&f_{\SE}(g)=(\W \star_{SE(3)}  f )  (g) = \langle L_{g}\W, f \rangle_{\mathbb{L}_2 (\Real^3)}\\ &=  \langle L_{(x, R)}\W, f \rangle_{\mathbb{L}_2 (\Real^3)}= \langle L_{x} L_R \W, f \rangle_{\mathbb{L}_2 (\Real^3)} \\
&=\!\int_{\Real^3} \W(g^{-1} \circ  \tilde x)f(\tilde x) d \tilde x \!=\!\int_{\Real^3} \W(R^{-1} (\tilde x-x)) f(\tilde x) d \tilde x,
 \end{split}
 \]
where $\star_{\SE}$ denotes a lifting cross-correlation operator to $\SE$.

The subsequent group correlation can then be performed in the lifted higher dimensional feature map of $SE(3)$, i.e., $f_{\SE}(g)$ .
\[
\begin{split}
y(g) 
 &= \langle L_{g}\W, f_{\SE} \rangle_{\mathbb{L}_2 (SE(3))}\\
&= 
 \int_{SE(3)} \W (g^{-1} \cdot \tilde g) f_{\SE}(\tilde g) d \tilde g.
 \end{split}
  \]

Consider now a 3D convolution kernel that is expanded in a spherical
harmonic basis (up to degree $L$):  
\begin{equation} \label{eq:convolution_kernel_SH_expanded}
\W(x)=\W_{c (\| x\|)} (x):= \sum_{l=0}^L \sum_{m=-l}^{l} c_m^{l} (\| x\|) Y_m^{l} \left ( \frac{x}{\|x\|}\right ),
\end{equation}
where $c= (c_0^{0}, c_{-1}^{1}, c_0^{1}, \cdots)^T$ is the collection of basis coefficients that depends on $\|x\|.$
Due to \eqref{eq:Lgf}, we conclude that the kernel is steerable via the transformation of $c$ by its $SO(3)$ representation $D(R)$
\[
L_g  \W(x)=\W(g^{-1} x)=\W_c(R^{-1}x)=\W_{D(R)c(\| x\|)} (x).
\]
Using this steerable kernel, we can perform $SE(3)$ lifting convolution, i.e.,  $g = (x,R) \in \SE$, such as
\begin{equation} \label{eq:SE3_lifting_convolution}
    \begin{split}
    &y(g) =y(x,R)
    = \langle L_{g}\W, f \rangle_{\mathbb{L}_2 (\Real^3)}\\
    &=\int_{\Real^3} L_g \W(\tilde x) f(\tilde x) d \tilde x =\int_{\Real^3} \W_{c(\|\tilde{x}-x\|)}( R^{-1}  (\tilde x-x)) f(\tilde x) d \tilde x\\
    &=\int_{\Real^3} \W_{D(R)c(\|\tilde x - x\|)} (\tilde x-x)  f(\tilde x) d \tilde x \\
    &=\int_{\Real^3} ({D(R)c(\|\tilde{x} -x\|)})^T \Phi (\tfrac{\tilde x- x}{\|\tilde{x} - x\|})  f(\tilde x) d \tilde x\\
    &=D(R)^T \int_{\Real^3}  c(\|\tilde{x} -x\|))^T \Phi (\tfrac{\tilde x- x}{\|\tilde{x} - x\|})  f(\tilde x) d \tilde x\\
    &=D(R)^T \hat f^{\Phi}_c (x),
    \end{split}
\end{equation}
where
\begin{eqnarray*}
    \hat{f}^{\Phi}_c(x) &\triangleq& \int_{\Real^3}  c(\|\tilde{x} -x\|))^T \Phi (\tfrac{\tilde x- x}{\|\tilde{x} - x\|})  f(\tilde x) d \tilde x \\
      &=& (\hat{f}^{0}_{0}(x), \hat{f}^{1}_{-1}(x), \hat{f}^{1}_{0}(x), \hat{f}^{1}_{1}(x), \cdots )^T
\end{eqnarray*}
is a steerable vector of responses. 

Here, any transformation $g \in \mathbb{G}$ on $\W$ can be represented as a linear combination of a set of basis $\Phi$. Hence, ${D(R)}^T $ can be pulled out of the convolution integral. Therefore, $y(g)$ can be expressed as a function of $g$ without further convolution since the feature vector is already convoluted with a collection of basis, which does not require discretizations in $R$ as in  the non-steerable group convolution in \eqref{eq:LiftingCov}.

\subsubsection{Clebsch-Gordan Tensor Product}
Clebsch-Gordan (CG) tensor product is used in steerable linear layers \cite{veefkind2024probabilistic}, which map between steerable input and output vector spaces. Similar to regular neural networks, where matrix-vector multiplication is used to map between input and output vector spaces, the CG tensor product maps between steerable input and output vector spaces. The learnable CG tensor product is the key component of steerable neural networks, similar to the learnable weight matrix-feature multiplication in regular neural networks.

For any two steerable vectors $u \in  V_{\ell_1}$ (type-$l_1$) and $v \in  V_{\ell_2}$ (type-$l_2$), the CG tensor product’s output is
again steerable with a $SO(3)$ representation $D(R)$, $R\in \SO$ as follows. 
\[
D(R) (u \otimes v)= (D^{l_1} (R) u ) \otimes (D^{l_2} (R) v ),
\]
where $u \otimes v = u v^T$ for vectors $u$ and $v$. 

Let's define $\text{vec} (u \otimes v) =  \text{vec} (u v^T)$ as the vector concatenation of the tensor product.
Then, $\text{vec} (u \otimes v)$ is steered by the Kronecker product matrix representation $D(R) = D^{l_2}(R) \otimes D^{l_1}(R)$ as follows
\[
\begin{split}
\text{vec} \left ( (D^{l_1} (R) u) \otimes (D^{l_2} (R) v)   \right ) &=
\text{vec} \left ( (D^{l_1} (R) u) (D^{l_2} (R) v)^T   \right ) \\
&=\text{vec} \left ( ( D^{l_1} (R) u v^T D^{l_2} (R)^T   \right )\\
&=(D^{l_2} (R) ) \otimes (D^{l_1} (R)   )\text{vec} (u v^T)\\
&=D(R)  \text{vec} (u v^T)\\
& = D(R)  \text{vec} (u \otimes v) ,
\end{split}
\]
where $\text{vec}(AXB) = (B^T \otimes A) \text{vec}(X)$ was used in the third line and   $A \otimes B$ denotes the Kronecker product of matrices $A$ and $B$.



Tensor products are also important because they can be used to construct different types of new steerable vectors.
By a change of basis, the tensor product $ {u}\otimes  {v}$ can be decomposed into the direct sum of type-$l$ vectors using the \textit{Clebsch-Gordan coefficients} \cite{thomas2018tensor, zee, griffiths2018introduction}. Let $u \in  V_{\ell_1}$ (type-$l_1$) and $v \in  V_{\ell_2}$ (type-$l_2$) be steerable vectors. Then, we can utilize the Clebsch-Gordan (CG) tensor product to define the $m$-th component of the type $l$ sub-vector from the output of the tensor product between these two steerable vectors, $( {u}\otimes_{cg}  {v})^{l}_m \in  V_{\ell}$, as follows

\begin{equation}
    ( {u}\otimes_{cg}  {v})^{l}_m=\sum_{m_1=-l_1}^{l_1}\sum_{m_2=-l_2}^{l_2} C^{(l,m)}_{(l_1,m_1)(l_2,m_2)}u_{m_1}v_{m_2} \,,
\end{equation}
where  the $C^{(l,m)}_{(l_1,m_1)(l_2,m_2)}$'s are the Clebsch-Gordan coefficients in the real basis. The CG tensor product is generally sparse since $C^{(l,m)}_{(l_1,m_1)(l_2,m_2)} = 0$ 
 for ${|l_1-l_2|\leq l \leq l_1+l_2}$.

\subsection{$SE(3)$-Equivariant Graph Neural Networks} \label{sec:equiv_gnn}
Generated from RGBD input, 3D point clouds have become another major source of vision inputs. Unlike 2D images, where only the projected information is provided, 3D point clouds provide more spatial information. Importantly, the points in the point cloud can be interpreted as $\mathbb{R}^3$ vectors, i.e., $x,y,z$ coordinates, equipped with feature information vectors such as, for example, the RGB value of each point. Point clouds can be handled very efficiently by $SE(3)$ equivariant graph neural networks.
 Let $\mathcal{V}$ be a (steerable) vector space and $\mathcal{X}$ and $\mathcal{Y}$ be (point-cloud data) sets. Let $G$ be a Lie matrix group with  group action $\circ$ such that, $\forall g,h\in G$, $g\circ (h\circ \dX)=(g \cdot h)\circ \dX$ and $g\circ (h\circ \dY)=(g\cdot h)\circ \dY$ where $\dX\in\mathcal{X}$, $\dY\in\mathcal{Y}$, and $\mathcal{O}$ denotes point-cloud data.
 A map $ {f}(\dX|\dY):\mathcal{X}\times\mathcal{Y}\rightarrow\mathcal{V}$ is said to be $G$-\textit{equivariant} if
 \begin{equation}
      {D}_{\mathcal{V}}(g) {f}(\dX|\dY)= {f}(g\circ \dX|g\circ \dY),
 \end{equation}
 where $ {D}_{\mathcal{V}}(g)$ is the representation of $g \in G$ acting on $\mathcal{V}$. In the special case where $ {D}_{\mathcal{V}}(g) = {I}$, the map $ {f}(\dX|\dY)$ is said to be \textit{$G$-invariant}.

 A translation invariant and $SO(3)$-equivariant type-$l$ vector field, or simply an \textit{$SE(3)$-equivariant type-$l$ vector field}, $ {f}( {x}|\dX):\mathbb{R}^3\times\mathcal{X}\rightarrow\mathbb{R}^{2l+1}$ is a special case of $SE(3)$-equivariant map such that
 \begin{equation}
      {D}^l( {R}) {f}( {x}|\dX)= {f}( {R} {x}+ {p}|g\circ \dX)\quad \forall g=(  {p}, R)\in SE(3), 
 \end{equation}
 where  $ {p}\in\mathbb{R}^3$, $ {R}\in SO(3)$ and $g \circ {x}= gx = {R} {x}+ {p}$ (following our pre-established notation).
Graph neural networks are often used to model point cloud data \cite{wang2019dynamic, te2018rgcnn, shi2020point}. 
$\SE$-equivariant graph neural networks \cite{thomas2018tensor, se3t, liao2022equiformer,brandstetter2021geometric} exploit the roto-translation symmetry of graphs with spatial structures.
We briefly introduce Tensor Field Networks (TFNs) \cite{thomas2018tensor} and one of its variants, the $\SE$-transformers \cite{se3t}, as the backbone graph networks for equivariant deep learning models. Furthermore, we briefly introduce some works to reduce the computational complexity of the tensor product - Equivariant Spherical Channel Network (eSCN) and Gaunt Tensor Product.

\subsubsection{Tensor Field Networks} Tensor field networks (TFNs) \cite{thomas2018tensor} are $SE(3)$-equivariant models for generating representation-theoretic vector fields from a point cloud input. TFNs construct equivariant output feature vectors from equivariant input feature vectors and spherical harmonics. The objective of the TFN is to achieve a $\SE$ equivariant map, which is called a $\SE$ representation induced from the representation of $\SO$, known as ``induced representation'' \cite{weiler20183d}. 
The feature vectors herein can be considered as some high-dimensional vectors attached to points, i.e. a  vector field, which is the reason for its name. Spatial convolutions and tensor products are used to attain  equivariance.

Consider a featured point cloud input with $M$ points given by 
\[\dX=\{( {x}_1, {f}_1),\cdots,( {x}_M, {f}_M)\},\] where $ {x}_i\in\mathbb{R}^3$ is the position and $ {f}_i$ is the equivariant feature vector associated with the $i$-th point.
Let $ {f}_i$ be decomposed into $N$ vectors such that $ {f}_i=\bigoplus_{n=1}^{N} {f}_i^{(n)}$, where $ {f}_i^{(n)}$ is a type-$l_n$ steerable vector, which is $(2l_n+1)$ dimensional. 
Therefore, we define the action of $g=( p, {R})\in SE(3)$ on $\dX$ as
\begin{equation*}
   g\circ \dX=\{(g{x}_1, {D}( {R}) {f}_1),\cdots,(g {x}_M, {D}( {R}) {f}_M)\},
\end{equation*}
where ${p}\in\mathbb{R}^3$, ${R}\in SO(3)$,  and $ {D}( {R})=\bigoplus_{n=1}^{N} {D}^{l_n}( {R})$. 
Note that the feature vector $f_i$ can be understood as the result of a lifting correlation (convolution) such as  \eqref{eq:SE3_lifting_convolution}; the feature vector $\hat{f}^{\Phi}_c$ is equivariant. However, in this equivariant graph neural network formulation, it is often assumed that  points are already associated with equivariant feature vectors.


Consider the following input feature field $ {f}_{(in)}( {x}|\dX)$ generated by the point cloud input $X$ as
\begin{equation} \label{eq:high_order_feature}
     {f}_{(in)}( {x}|X)=\sum_{j=1}^{M} {f}_{j}\delta^{(3)}( {x}- {x}_j),
\end{equation}
where $\delta^{(3)}( {x}- {y})=\prod_{\mu=1}^3\delta(x^\mu-y^{\mu})$ is the three-dimensional Dirac delta function centered at $ y \in \mathbb{R}^3$, with $x^\mu,y^\mu \in \mathbb{R}$,
where $x = [x^1, x^2, x^3]^T$ and $y = [y^1,y^2,y^3]^T$. The meaning of the Dirac delta function is to read the feature value $f_j$ associated with the point $x_j$.
Note that this input feature field is an $SE(3)$-equivariant field, that is:
\begin{equation*}
      {f}_{(in)}(g {x}|g\circ \dX) =  {D}( {R}) {f}_{(in)}( {x}|\dX)\quad\forall\ g=(p,  {R})\in SE(3).
\end{equation*}

Now consider the following output feature field generated by a cross-correlation (convolution)
\begin{equation}
\label{eqn:tfn}
    \begin{split}
     {f}_{(out)}&( {x}|\dX)=\bigoplus_{n'=1}^{N'} {f}_{(out)}^{(n')}( {x}| \dX)\\
    &=\int_{\Real^3} \W( {x}- {y}) {f}_{(in)}( {y}|\dX) dy^3\; =\sum_{j} \W( {x}- {x}_j) {f}_{j},
    \end{split}
\end{equation}
where the convolution kernel $ \W( {x}- {y})\in \mathbb{R}^{dim( {f}_{(out)})\times dim( {f}_{(in)})}$
 whose $(n',n)$-th block $ \W^{(n',n)}( {x}- {y})\in\mathbb{R}^{(2l_{n'}+1)\times(2l_n+1)}$ is defined as follows:
\begin{equation}
    \label{eqn:tfn_kernel_const}
    \begin{split}
        \left[ \W^{(n',n)}( {x})\right]^{m'}_{m}\!=\!\sum_{J=|l_{n'}-l_n|}^{l_{n'}+l_n}\phi^{(n',n)}_{J}(\| {x}\|)\sum_{k=-J}^{J}C^{(l_{n'},m')}_{(l_n,m)(J,k)}Y^{J}_{k}\left( {x}/\| {x}\|\right)
    \end{split}
\end{equation}
and $[\cdot]_m^{m'}$ denotes $m,m'$-th element of the block.
Here, $\phi^{(n',n)}_{J}(\| {x}\|):\mathbb{R}\rightarrow\mathbb{R}$ is some learnable radial function. 
The notable difference between the \eqref{eq:convolution_kernel_SH_expanded} is that in \eqref{eqn:tfn}, there are Clebsh-Gordan coefficients. Clebsh-Gordan coefficients enable important properties to make the $\SE$-equivariant field model.
The formulation of steerable convolution kernel can be referred to \cite{weiler20183d}. In a nutshell, because of the property of the Clebsch-Gordan coefficients, known as intertwining, the steerable kernel has the following properties:
\[
    \W^{(n',n)}(Rx) = D^{l_{n'}}(R) \; \W^{(n',n)}(x) \; D^{l_{n}}(R)^{-1}
\]
Using this property, the output feature field $ {f}_{(out)}( {x}|\dX)$ in \eqref{eqn:tfn} is proven to be $SE(3)$-equivariant \cite{thomas2018tensor, se3t}. Consider type $l_{n'}$ vector $f_{(out)}^{(n')}(x|\dX)$ such that consists $f_{(out)}(x|\dX) = \oplus_{n'=1}^N f_{(out)}^{(n')}(x|\dX)$. From \eqref{eqn:tfn} and considering the property of the concatenation, it reads that 
\[
    f_{(out)}^{(n')}(x|\dX) = \int d^3y \; \sum_{p=1}^n k^{(n',p)}(x-y) \; f_{(in)}^{(p)}(y|\dX).
\]
Applying the group transformation to the equation above,
\[
\begin{split}
    &f_{(out)}^{(n')}(gx|g\circ \dX) \\
    &= \int d^3 y \sum_{p=1}^n k^{(n',p)}(g\circ(x-y)) f_{(in)}^{(p)}(gy|g\circ \dX) \\
    &= \int d^3y \sum_{p=1}^n k^{(n',p)}(R(x-y)) D^{l_p}(R) f_{(in)}^{(p)}(y|\dX)\\
    &= \int d^3y \sum_{p=1}^n D^{l_{n'}}(R) k^{(n',p)}(x-y)  (D^{l_p}(R))^{\text{-}1} D^{l_p}(R) f_{(in)}^{(p)}(y|\dX)\\
    &= D^{l_{n'}}(R)\int d^3y \sum_{p=1}^n k^{(n',p)}(x-y) f_{(in)}^{(p)}(y|\dX)\\
    &= D^{l_{n'}}(R) \; f_{(out)}^{(n')}(x|\dX)
\end{split}
\]
which shows the equivariance of the output feature field $f_{(out)}(x|\dX)$.

\paragraph*{$SE(3)$-Transformers} The $SE(3)$-Transformers \cite{se3t} are variants of TFNs with self-attention. Consider the case in which the output field is also a featured sum of Dirac deltas
\begin{equation}
     {f}_{(out)}( {x}|\dX)\!=\!\sum_{j=1}^{M} {f}_{(out),j}\delta^{(3)}( {x}- {x}_j),
\end{equation}
where $ {x}_i$ is the same point as that of the point cloud input $\dX$. The $SE(3)$-Transformers apply type-0 (scalar) self-attention $\alpha_{ij}$ to \eqref{eqn:tfn}:
\begin{equation}
\begin{split}
     {f}_{(out),i}=\sum_{j\neq i}\alpha_{ij} \W( {x}- {x}_j) {f}_{j}+\bigoplus_{n'}^{N'}\sum_{n=1}^{N} \W_{(S)}^{(n',n)} {f}_{j}^{(n)},
\end{split}
\end{equation}
where the $ \W_{(S)}^{(n',n)}$\ \ term is called the \textit{self-interaction} \cite{thomas2018tensor}. $ \W_{(S)}^{(n',n)}$ is nonzero only when $l_n'=l_n$. The self-interaction occurs where $i=j$ such that $ {\W}(x_i-x_j)=k(0)$. The self-interaction term is needed because $ {\W}$ is a linear combination of the spherical harmonics,  which are not well defined in $ {x}=0$. Details about the calculation of the self-attention $\alpha_{ij}$ can be found in \cite{se3t}.

\subsubsection{Handling Computational Complexity of Tensor Product}
Equivariant Spherical Channel Network (eSCN)~\cite{passaro2023reducing} introduces a mathematically equivalent alternative to the original tensor product operation in TFNs with improved efficiency. eSCN reduces the computational complexity of tensor product from $O(L^6)$ to $O(L^3)$ where $L$ is the highest feature type. This unlocks the utilization of much higher frequency features, which have been reported to improve the accuracy in robotic manipulation tasks \cite{gao2024riemann, hu2024orbitgrasp}. The details of the eSCN can be found in Appendix~\ref{sec:appendix_eSCN}. 

In \cite{luo2024enabling}, Clebsch-Gordan coefficients are connected to the Guant coefficients, which are integrals of products of three spherical harmonics. Through this transformation, the complexity of full CG tensor products is reduced from $O(L^6)$ to $O(L^3)$. 


\subsection{PointNet-based Equivariant Neural Network} PointNet \cite{qi2017pointnet} is a neural network architecture designed to process point clouds directly, ensuring permutation invariance so that the output remains consistent regardless of the order of the input points. PointNet introduces T-net, an affine transformation network, to learn automatic transformation. The motivation of the T-net is to somewhat mimic equivariance property, but there is no theoretical motivation or guarantee.

The Vector Neurons (VN)\cite{deng2021vector} extends scalar neurons to 3D vector neurons, enabling the construction of $SO(3)$-equivariant neural networks. VN introduces essential building blocks such as linear layers, non-linearities, pooling, and normalization, which ensure $SO(3)$-equivariance. Based on these building blocks, the VN framework constructs $SO(3)$-equivariant PointNet, known as VN-PointNet. 




\section{Equivariant Deep Learning in Robotics}\label{sec:Related}



Equivariant neural networks can significantly improve data efficiency and generalizability in robot learning, where data collection is often expensive and large datasets are needed. Early efforts to leverage symmetry in this area include Transporter Network \cite{zeng2020transporter}, which utilized the $SO(2)$ symmetry of 2D top-down settings for pick-and-place tasks. This approach was later enhanced by Equivariant Transporter \cite{huang2022equivariant}, which improved sample efficiency by leveraging equivariant networks over $SO(2)\times SO(2)$. Fourier Transporter \cite{huang2024fourier} further advanced the field by enabling convolutions directly in $SE(3)$ non-Euclidean space, leveraging a parameterization in the Fourier space.

In grasp detection, Edge Grasp Network \cite{huang2023edge} introduced a novel approach utilizing the $SE(3)$-invariance for grasp quality evaluation. Orbit Grasp \cite{hu2024orbitgrasp} took this further by representing grasp directions with spherical harmonics, allowing the model to infer efficiently from a continuous range of directions.

Building on the Diffusion Policy (DP) \cite{chi2023diffusion}, which combines diffusion models with action chunking, EquiBot \cite{yang2024equibot} incorporates $SIM(3)$-equivariance using the Vector Neuron framework, as proposed by EquivAct \cite{yang2024equivact} into the DP framework. Similarly, the Equivariant Diffusion Policy \cite{wang2024equivariant} extends the DP framework by integrating $SO(2)$-equivariant noise prediction networks. This approach leverages irreducible representations to model how $SO(2)$ acts on an $SE(3)$ gripper pose. ET-SEED \cite{tie2024seed} also leverages $SE(3)$-equivariance but restricts equivariant operations to address the challenges in training caused by the strict constraints imposed by full equivariance.

Instead of diffusion models, ActionFlow \cite{funk2024actionflow} and EquiGraspFlow \cite{lim2024equigraspflow} utilize flow matching to quickly generate $SE(3)$-equivariant action sequences and grasp poses, respectively.

The equivariant neural networks can be utilized for symmetry in robot manipulation tasks. Numerous robot learning researchers have leveraged symmetry in $SE(3)$ \cite{zeng, huang2022equivariant, fourtran, ndf, ryu2022equivariant, ryu2024diffusion}. Especially, \cite{rndf, ryu2022equivariant, ryu2024diffusion, fourtran} introduced bi-equivariance of the pick and place task. 

Incorporating group equivariance into Reinforcement learning (RL) offers a promising approach to one of RL's key challenges: sample inefficiency due to extensive trial-and-error from scratch. Recent works \cite{rl3, finzi2021residual, theile2024equivariant, wang2022so, kohler2023symmetric, wang2022robot, wang2022equivariant, nguyen2023equivariant, zhao2024equivariant, zhao2024E2, zhao2022integrating} have shown the potential of these methods to enhance both sample efficiency and performance. In particular, \cite{wang2022equivariant, wang2022so} introduced a novel theoretical framework, the group-invariant MDP, which proves the existence of group-invariant $Q$-functions in $Q$-learning methods.

In this section, we will provide a detailed review of a few notable works for two major categories of equivariant deep learning in robotics: imitation learning and reinforcement learning. 

\subsection{Imitation Learning}\label{il}
Imitation learning provides a way to teach a robot by showing it the desired behavior, simplifying the teaching process, and mitigating detailed programming or laborious reward function engineering. However, the cost of collecting data, such as expensive hardware, a well-oriented laboratory environment, expertise with the robot, and a significant amount of time, decreases the scalability and efficiency of the method. 
\paragraph*{Equivariant Descriptor Fields (EDFs)}
Neural Descriptor Fields (NDF)\cite{ndf} implement object representation with a category-level descriptor using Vector Neurons (VN) \cite{deng2021vector}, enabling the learning of manipulation tasks from 5–10 demonstrations in both simulation and real-world settings. However, NDFs require pre-trained object segmentation and do not operate in an end-to-end manner. On the other hand, Equivariant Descriptor Fields (EDF)\cite{ryu2022equivariant} achieve local equivariance by utilizing the locality of Equiformer\cite{liao2022equiformer} and $SE(3)$-Transformer\cite{se3t}. This enables EDFs to perform training in an end-to-end manner without pre-training and object segmentation. Furthermore, EDF introduces a bi-equivariant energy-based model on $SE(3)$ for the pick and place problem. The bi-equivariant energy-based model on $SE(3)$ is written as :
\[
P(g|\mathcal{O}^{scene}, \mathcal{O}^{grasp}) = \frac{\exp{\left(-E(g|\mathcal{O}^{scene}, \mathcal{O}^{grasp})\right)}}{Z}
\]
\[
Z = \int_{SE(3)} dg \; \exp{\left(-E(g|\mathcal{O}^{scene}, \mathcal{O}^{grasp})\right)}
\]
where $P$ is an energy-based model on the $SE(3)$ manifold, $E$ is an energy function, $g = (p, R) \in SE(3), p \in \Real^3, R \in SO(3)$ and $\mathcal{O}$ denotes point clouds data.
By addressing equivariance in both the scene and the grasp (Fig.~\ref{fig:biequiv}), EDF achieves high sample efficiency and robustness when handling out-of-distribution data. The bi-equivariance condition of the model can be written as follows: 
\[
\begin{split}
P(g|\mathcal{O}^{scene}, \mathcal{O}^{grasp}) &= P(\Delta g_w g | \Delta g_w \circ \mathcal{O}^{scene},\mathcal{O}^{grasp}) \\
&= P(g \Delta g_e^{-1}|\mathcal{O}^{scene}, \Delta g_e \circ \mathcal{O}^{grasp}),
\end{split}
\]
where $\Delta g_w$ denotes a transformation on the world-frame (spatial frame) and $\Delta g_e$ denotes the transformation on the end-effector frame.
\begin{figure}[t!]
    \centering
    \includegraphics[width=\linewidth]{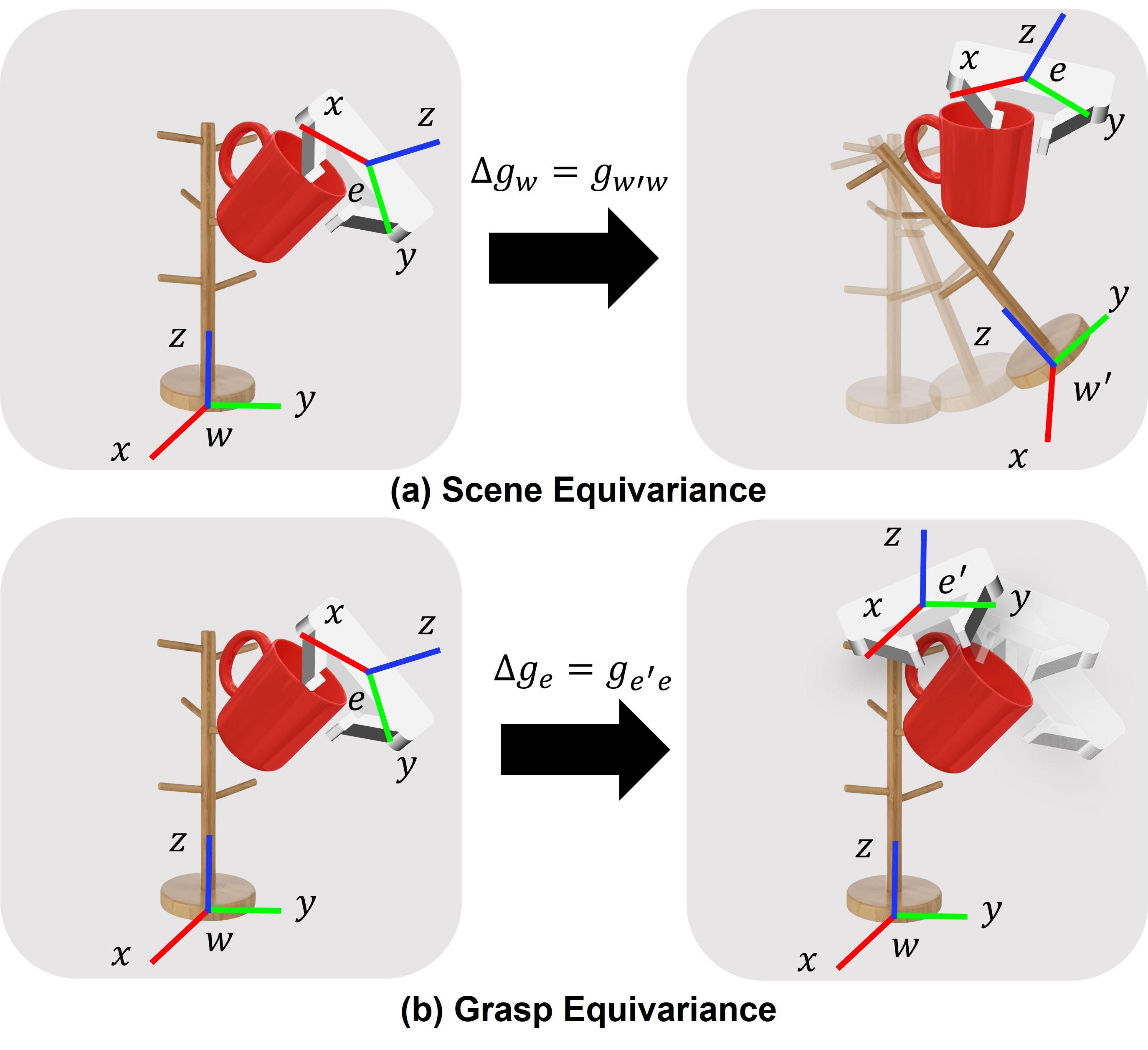}
    \caption{Bi-equivariance. (a) The scene equivariance and (b) grasp equivariance consider the transformation of world and grasped object. The end-effector pose should be transformed equivariantly with the transformation.}
    \label{fig:biequiv}
\end{figure}
The energy-based model consists of $SE(3)$-equivariant type-$l_n$ vector field generated by observation. EDF proposed the energy-based model $P(g|\mathcal{O}^{scene}, \mathcal{O}^{grasp})$ is bi-equivariant if the energy function $E(g|\mathcal{O}^{scene}, \mathcal{O}^{grasp})$ is bi-equivariant. The energy function is written as:
\[
E(g|\mathcal{O}^{scene}, \mathcal{O}^{grasp}) =
\]
\[
\int_{\mathbb{R}^3} d^3 x \; \rho(x|\mathcal{O}^{grasp})||\varphi(gx|\mathcal{O}^{scene})-D(R)\psi(x|\mathcal{O}^{grasp}) ||^2
\]
where $\rho(x|\mathcal{O}^{grasp})$ is the query density, an $SE(3)$-equivariant non-negative scalar field, $\varphi(gx|\mathcal{O}^{scene})$ is key EDF and $\psi(x|\mathcal{O}^{grasp})$ is the query EDF. 
Diffusion-EDFs\cite{ryu2024diffusion} improved the training time of EDFs, maintaining end-to-end training manner and data efficiency utilizing the bi-equivariance and locality of robotic manipulation. Diffusion-EDFs proposed a bi-equivariant score function given as follows:
\[
s(\Delta g g | \Delta g \circ \mathcal{O}^{scene}, \mathcal{O}^{grasp}) = s(g | \mathcal{O}^{scene}, \mathcal{O}^{grasp})
\]
\[
s(g\Delta g^{-1}|\mathcal{O}^{scene}, \Delta g \circ \mathcal{O}^{grasp}) = [\Ad_{\Delta g}]^{-T}s(g | \mathcal{O}^{scene}, \mathcal{O}^{grasp})
\]
with 
\[
\quad \Ad_g = \begin{bmatrix}
            R & \hat{p} R \\
            0 & R
            \end{bmatrix}\\
\]
where $s(g | \mathcal{O}^{scene}, \mathcal{O}^{grasp}) = \nabla \log{P(g | \mathcal{O}^{scene}, \mathcal{O}^{grasp})}$, $R \in \text{SO}(3)$, $p \in \Real^3$ and $\hat{p}$ is the skew-symmetric matrix of $p$.
Diffusion-EDFs split the score model into translation and rotational parts and propose the models of each part satisfying equivariance conditions as follows:
\[
s_t(g|\mathcal{O}^{scene}, \mathcal{O}^{grasp}) = [s_{\nu;t}\oplus s_{\omega ; t}](g|\mathcal{O}^{scene}, \mathcal{O}^{grasp})
\]
\[
s_{\nu ; t}(g|\mathcal{O}^{scene}, \mathcal{O}^{grasp}) =
\]
\[
 \int_{\mathbb{R}^3} d^3 x \; \rho_{\nu ; t}(x|\mathcal{O}^{grasp})\Tilde{s}_{\nu ; t}(g, x|\mathcal{O}^{scene}, \mathcal{O}^{grasp})
\]
\[
 s_{\omega ; t}(g|\mathcal{O}^{scene}, \mathcal{O}^{grasp}) =
\]
\[
\int_{\mathbb{R}^3} d^3 x \; \rho_{\omega ; t}(x|\mathcal{O}^{grasp})\Tilde{s}_{\omega;t}(g, x|\mathcal{O}^{scene}, \mathcal{O}^{grasp})
\]
\[
+ \int_{\mathbb{R}^3} d^3 x \; \rho_{\nu ; t}(x|\mathcal{O}^{grasp})x \wedge \Tilde{s}_{\nu;t}(g, x|\mathcal{O}^{scene}, \mathcal{O}^{grasp})
\]
where $s_{\nu ; t}(\cdot|\mathcal{O}^{scene}, \mathcal{O}^{grasp})$ denotes the translational score, $s_{\omega ; t}(\cdot|\mathcal{O}^{scene}, \mathcal{O}^{grasp})$ denotes the rotational score, $\rho_{\square ; t}(\cdot | \mathcal{O}^{grasp})$ is the equivariant density field, \\
$\tilde{s}_{\square ; t}(\cdot|\mathcal{O}^{scene}, \mathcal{O}^{grasp})$ is the time-conditioned score fields where $\square$ is either $\nu$ or $\omega$, and $\wedge$ denotes the wedge product.
The details of the bi-equivariant score model and diffusion-based learning on the $SE(3)$ manifold for visual robotic manipulation are provided in Diffusion-EDFs.

\paragraph*{RiEMann}
RiEMann\cite{gao2024riemann} is a $\SE$-equivariant robot manipulation framework with near real-time. RiEMann learns an affordance map  $\phi(x|\dX)$ with one type-0 vector field to get a point cloud ball $B_{ROI}$ with $\SE$-Transformer \cite{se3t}.
\[
\phi(x|\dX) = f_{(in)}(x|\dX),
\]
\[
f_{(in)}(x|\dX) = \sum_{j=1}^{M}  \bigoplus_{c=1}^{3} {f}_{j}^{(0), c} \delta^{(3)}( {x}- {x}_j)
\]
where $\dX=\{( {x}_1, {f}_1),\cdots,( {x}_M, {f}_M)\} \in \mathbb{R}^{M \times 6}$ is a colored point cloud input with $M$ points, $X$ is a point cloud input, $x \in \mathbb{R}^3$ is the position, $f_j \in \mathbb{R}^3$ is the RGB color of the $j$-th point and $c$ is the channel of each RGB.

The region $B_{ROI}$ is extracted from the affordance map with a predefined radius. This leads to a reduction in computation complexity and memory usage. A translational action network $\psi_1(x|\dX)$ and an orientation network $\psi_2(x|\dX)$ use the points in the region. Translational action network $\psi_1(x|\dX)$ outputs one type-0 vector field followed by softmax and weighted sum. Orientation network $\psi_2(x|\dX)$ outputs three type-1 vector fields followed by mean pooling and Iterative Modified Gram-Schmidt orthogonalization (IMGS) \cite{hoffmann1989iterative}. 
\[
 \psi_1(x|\dX) + \psi_2(x|\dX) = {f}_{(out)}( {x}|\dX),
\]
\[
{f}_{(out)}( {x}|\dX)\!=\!\sum_{j=1}^{M} {f}_{j}^{(0)}\delta^{(3)}({x} \!-\! {x}_j) \!+\!  \sum_{j=1}^{M} \bigoplus_{i=1}^{3} {f}_{j}^{(1), i}\delta^{(3)}( {x} \!-\! {x}_j)
\]
With the design of the translation and rotation matrix with one type-0 vector field and three type-1 vector fields each, RiEMann can predict the target object pose directly.

\paragraph*{Fourier Transporter}
Fourier Transporter \cite{fourtran} proposed $SE(3)$ bi-equivariant model using 3D convolutions and a Fourier representation of rotation for pick and place problems. 
Similar to EDFs \cite{ryu2022equivariant} and Diffusion-EDFs \cite{ryu2024diffusion}, the proposed method utilize coupled 
$SE(d) \times SE(d)$-symmetries for the problem. 
Let's assume $d = 3$, the 3D pick and place problem. The policy for pick and place is formulated as $p(a_t|o_t) = p(a_{place}|o_t, a_{pick})p(a_{pick}|o_t)$, where $a_{place}, a_{pick} \in \SE$ are the placing and picking poses, respectively. 
The $SE(3)$-equivariant pick network is a function given by 
\begin{equation*}
    f_{pick} : o_t \mapsto p(a_{pick}|o_t),
\end{equation*}
where $p(a_{pick}|o_t)$ is a probability distribution.
The base space of the steerable field determines the pick location $\Real^3$, and the fiber space encodes the distribution of pick orientations $\SE$. We can represent the distribution $p(a|o_t)$ as $\Real^3 \rightarrow \{\SO \rightarrow \Real \}$. Especially, $L_g(p)(h) = p(g^{-1} \circ h)$ is the transformation of the distribution with left-action. $L_g$ notation is a left-regular representation of group element $g$ to distribution, and $\circ$ is a group action of $g$ with respect to $h$. $a_{pick}$ can be recovered via $a_{pick} = \argmax_{a_{pick}} p(a_{pick}|o_t)$.

The pick pose distribution should follow a transformation on observation.
\[
f_{pick}(g\circ o_t) = \text{Ind}_{\rho}(g) f_{pick}(o_t), \quad \forall g \in \SE,
\]
The action of $g$ via representation $\rho$ on steerable feature vector fields $f$ is defined as $\text{Ind}_{\rho}(g)(f)(x) = \rho(R) f(g^{-1} x)$, often referred to as induced representation, where $g = (p,R)$ and $x \in \mathbb{R}^3$ is the position.
$\rho(g)$ is the fiber action which transforms the pick orientation, i.e., $\rho(g)$ is a representation of group element $g$, and the base space action $(\beta (g)f)(x) = f(g^{-1}x)$ rotates the pick location where $\beta = \text{Ind}_{\rho_{0}}$, with $\rho_0(R) = 1$ is a trivial representation.
We use $\rho$ notation instead of the previous $D$ because the representation $\rho$ can handle the representation for the fiber and more general groups, such as Euclidean group $E(N)$. The standard representation $D^1$ of $\SO$ for one group element is a standard $3 \times 3$ rotation matrix. The irreducible representation $D$ of $\SO$ is the Wigner D-matrix. The grasped object can be represented as a voxel patch centered on the pick location with the assumption that the object does not move or deform. 

The place network is 
\begin{equation*}
    f_{place}: (c, o_t) \to p(a_{place}|o_t, a_{pick}),
\end{equation*}
where $c$ is the cropped image centered at the pick location. The network should satisfy the bi-equivariance constraint in the pick and place problem. 
\begin{equation} \label{eq:fourier_transporter_place}
\begin{split}
    &f_{place}(g_1 \circ c, g_2 \circ o_t) = \text{Ind}_{\rho}(g_2)\rho_{R}(g_1^{-1})f_{place}(c, o_t), \\
    &\forall g_1, g_2 \in \SE
\end{split}
\end{equation}
Although \eqref{eq:fourier_transporter_place} was originally presented in the \cite{huang2024fourier}, but its definition does not cope well with the definitions, e.g., $\rho_R$. The definition of $\rho_R$ provided in the paper is for $f:\SO \to \R$, $(\rho_R(g)f)(h) = f(hg^{-1})$, for $g,h\in \SO$, but $f_{place}$ from \eqref{eq:fourier_transporter_place} is a mapping from $c$ and $o_t$ to $\R$. In fact, the meaning of \eqref{eq:fourier_transporter_place} can be better understood by its objective. Consider $a_{place} = (p_{place}, R_{place})\in \SE$, such that
\begin{equation*}
    a_{place} = \argmax_{a_{place}} p(a_{place}|a_{pick}, o_t).
\end{equation*}
Then, by the action of $\text{Ind}_{\rho}(g_2)$ and $\rho_R(g_1^{-1})$, $R_{place}$ will be transformed via $\rho_1(g_2) R_{place} \rho_1(g^{-1})$ and $p_{place}$ will be transformed via $\rho_(g_2) p_{place}$.
The place action should transform equivariantly with actions on the picked object and actions on the observation. 
The template matching method is used to realize place symmetry. The cross-correlation between a dynamic kernel $\kappa(c)$ and a feature map $\phi(o_t)$ is used for pick action probability as $f_{place}(c, o_t) = \kappa(c) \star \phi(o_t)$, where $\star$ denotes the cross-correlation.

The dynamic kernel $\kappa(c)$ is introduced and utilized to perform convolutions on the Fourier space. The dynamic kernel is composed of dense feature maps, which are lifted with a finite number of rotations and Fourier transforms to the channel space. The dense feature map of crop $c$ is  $\psi(c)$. We rotate the dense feature map with a finite number of rotations $\Tilde{G} = \{R_i | R_i \in \SO\}_{i=1}^m$ and stack the rotated feature maps. The stack of fully rotated signals can be defined as follows:
\[
\begin{split}
    &\mathcal{L}^{\uparrow}[f](x) = \{ f(R_1^{-1}x), f(R_2^{-1}x) \dots , f(R_m^{-1}x)\},\\
    &\forall x \in \mathbb{R}^3 , R_i \in \Tilde{G},
\end{split}
\]
The stack of fully rotated dense feature maps of the cropped image is $\mathcal{L}^{\uparrow}(\psi(c))$.

The Fourier transform over a function on $\SO$ $h : \SO \to \mathbb{R}$ can be defined as in \eqref{eq:fourier_transform_on_function}.
Let $\mathcal{F}^{+}$ denote Fourier transform to the channel-space. Then, $\mathcal{F}^{+}$ is applied to $\mathcal{L}^{\uparrow}(\psi(c))$ . Finally, the dynamic kernel becomes $\kappa (c) = \mathcal{F}^{+} [\mathcal{L}^{\uparrow} (\psi(c))] $.

The cross-correlation between the feature map $\phi(o_t)$ and the dynamic kernel $\kappa(c)$ is performed in the Fourier space. The place network is bi-equivariant if the $\psi(c)$ and the $\phi(o_t)$ satisfy the equivariant property and $\kappa(c)$ is a steerable convolutional kernel with type $0$ input. The details of $\SO$-Fourier Transform and proof of propositions are provided in Fourier Transporter \cite{fourtran}.

\subsection{Equivariant Reinforcement Learning}\label{rl}
\begin{figure}
  \centering
\includegraphics[width=\linewidth]{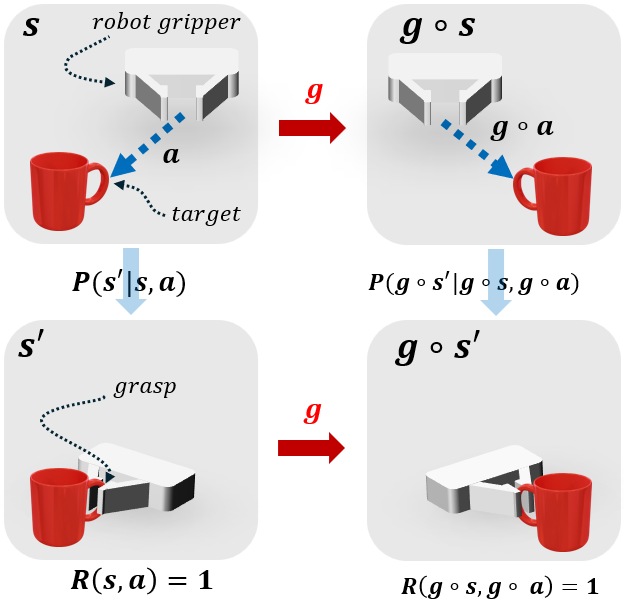}
  \caption{\textbf{Group-invariant MDP}. In a robot manipulation task, if the reward is 1 when the gripper achieves the goal, the group invariant MDP preserves the reward function and transition probability invariance under group actions.}
  \label{fig:gimdp}
\end{figure}

Many equivariant reinforcement learning methods are based on the fundamental concept of group-invariant Markov decision process (MDP) \cite{wang2022so} (see Fig.~\ref{fig:gimdp}), denoted as $\mathcal{M}_G=(S,A,P,R,G)$. 
It is based on the MDP homomorphisms \cite{ravindran2001symmetries, ravindran2004approximate}, a mapping that relates the original MDP, $M=(S,A,P,R,\gamma)$, to another abstract MDP, $M'=(S',A',P',R',\gamma)$. This homomorphism consists of two subjective functions $\alpha: S\rightarrow S'$ and $\beta: A\rightarrow A'$, which relate the state and action space of $M$ to the corresponding abstractions in $M'$. The policy in $M'$ can be utilized to derive a corresponding policy in $M$, which is given by:
\[
\begin{split}
    \pi(a|s)=\frac{\pi'(a'|\alpha(s))}{|\beta^{-1}(a')|}
\end{split}
\]
where $s\in S, a\in A, a'\in A'$ and $\pi'$ is the policy from the abstract MDP $M'$. Previous work has shown that if the abstract policy $\pi'$ is optimal, it can be used to derive the optimal policy in the original MDP, $M$ \cite{ravindran2001symmetries}.
Moreover, when the homomorphism follows a group framework, the optimal policy derived from the group-transformed MDP can be mapped back to induce the optimal policy in the original MDP. This approach leverages the symmetry within the problem to improve learning efficiency. The group-invariant MDP maintains its properties under the group-invariance of both the transition probability and the reward function, which can be expressed as:
\begin{itemize}
    \item Reward function: $R(s,a) = R(g\circ s,g\circ a)$
    \item Transition probability: $P(s'|s,a)=P(g\circ s'|g\circ s,g\circ a)$ 
\end{itemize}
where $g$ represents the group action in any group $\mathbb{G}$. Due to the group-invariant properties, the optimal $Q$ function $Q^*(s,a)$ also becomes group-invariant and the optimal policy $\pi^*(s)$ is equivariant, which can be expressed as:
\begin{itemize}
    \item Q function: $Q^*(s,a)=Q^*(g\circ s,g\circ a)$
    \item Policy: $\pi^*(g\circ s)=g\circ \pi^*(s)$
\end{itemize}

Consider the Bellman optimality equation: 
\[
Q^*(s,a)=R(s,a)+\gamma\; \sup_{a'}\int_{s'}ds'P(s'|s,a)Q^*(s',a').
\]
Then, $Q^*$ of the transformed state-action pair, i.e., \\$Q^*(g\circ s', g\circ a')$ is given as:\\
\[
\begin{split}
    &Q^*(g\circ s,g\circ a)=R(g\circ s, g\circ a)\\ &+\gamma\; \sup_{a'}\int_{s'}ds'P(g\circ s'|g\circ s,g\circ a)Q^*(g\circ s', g\circ a').
\end{split}
\]
Following the Lemma in the prior work \cite{wang2022equivariant}, let us consider the case where the state space $S$ corresponds to a visual state space that is equivariant under the group action, such that $S=g \circ S$ for all $g\in G$. By reparameterizing the integral using the substitution $s'=g \circ s''$, and under the assumption that both the transition dynamics and the reward function are group-invariant, the optimal $Q^*$ for a transformed state-action pair can be rewritten as
\[
\begin{split}
    &Q^*(g\circ s,g\circ a)=R(g\circ s, g\circ a)\\
    &+\gamma\;\sup_{a'}\int_{s'}ds'P(g\circ s'|g\circ s, g\circ a)Q^*(g\circ s', g\circ a')\\
    &=R(s, a)+\gamma\;\sup_{a'}\int_{s'}ds'P(s'|s, a)Q^*(g\circ s', g\circ a').
\end{split}
\]
Introducing a new function $\bar{Q}(s,a)=Q(g\circ s, g\circ a)$, it follows that 
\[
\begin{split}
    \bar{Q}^*(s,a)=R(s, a) +\gamma\;\sup_{a'}\int_{s'}ds'P(s'|s, a)\bar{Q}^*(s', a').
\end{split}
\] Note that $Q^*(s,a)=Q^*(g\circ s, g\circ a)$ by the uniqueness of the Bellman optimal equation. Now, consider the optimal policy $\pi$ that maximizes $Q^*(s,a)$: $\pi^*(s)=\arg\max_{a}Q^*(s,a)$. We can relate this to the transformed state $g\circ s$. The policy in the transformed state is given by: 
\[
    \pi^*(g\circ s)=\arg\max_{a}Q^*(g\circ s,a).
\]
Using the fact that $Q^*(g\circ s,a)=Q^*(s,g^{-1}\circ a)$, as implied from the uniqueness of the Bellman equation, we can rewrite: 
\[
    \pi^*(g\circ s)=\arg\max_{a}Q^*(s,g^{-1}\circ a).
\] 
Letting $a=g\circ\bar{a}$, the equation can be expressed as: 
\[
\begin{split}
    \pi^*(g\circ s)&=g\circ \arg\max_{\bar{a}}Q^*(s,g^{-1} g\circ \bar{a})\\
    &=g\circ \pi^*(s)
\end{split}    
\]

Based on the group-invariant MDP, the authors in the prior work \cite{wang2022so} propose $SO(2)$-equivariant $Q$ learning methods. To construct group-invariant and equivariant layers for group-invariant vision-based $Q$ networks, they employ the networks from \cite{weiler2019general}, which encode the equivariant constraint on the kernel to achieve the equivariance. Using these equivariant networks, they construct equivariant networks for Deep $Q$-Networks (DQN), and both group-invariant critics and equivariant policies for Soft Actor-Critic (SAC). Experimental results demonstrate that the equivariant RL achieves superior sample efficiency compared to vanilla Q-learning methods, even when data augmentation or the auxiliary methods \cite{laskin2020reinforcement, kostrikov2020image, laskin2020curl} are employed. Additionally, the prior work shows better generalization over the group actions, outperforming other methods when given a fixed orientation of the input image rather than a random orientation.

In another prior work \cite{kohler2023symmetric} build upon the work by \cite{wang2022so} to propose an equivariant policy for the visual-force problems. Their approach utilizes the force input, in addition to the visual input, to facilitate equivariant policy learning. The experimental results indicate that the proposed method outperforms the baseline approaches. Furthermore, an ablation study reveals that incorporating additional sensory inputs, such as force or proprioception, enhances sample efficiency and performance, although this improvement is task-dependent. Finally, they also show that even when visual input quality is poor, the force input can compensate for it.

In \cite{nguyen2023equivariant}, the authors extend the equivariance to partially observable Markov decision process (POMDP), denoted as $\mathcal{P}_G=(S,A,P,R,\Omega,O,b_0)$. Here, $S,A,P,$ and $R$ represent the state space, action space, transition probability, and reward function, analogous to those in $\mathcal{M}_G$. The additional components $\Omega$ (observation space), $O$ (stochastic observation function), and $b_0$ (initial belief) incorporate group-invariant conditions. These include $O(g\circ a, g\circ s',g\circ o)$ for the observation function and $b_0(g\circ s)=b_0(s)$ for the initial belief. The history $h_t=(o_0,a_0,...,a_{t-1},o_t)$ is shown to remain invariant under group action $g\in G$, and the optimal value function $V^*(h)$ and policy $\pi^*$ also exhibit symmetric properties. Using an equivariant actor-critic framework, the authors demonstrate superior performance over non-equivariant methods in simulations and validate their approach by successfully transferring policies to real hardware in a zero-shot manner.

Equivariant networks have also been applied to offline reinforcement learning \cite{tangri2024equivariant}. This work leverages equivariant networks within specific methods like Conservative Q-Learning \cite{kumar2020conservative} and Implicit Q-Learning \cite{kostrikov2021offline} and tackles the challenge of outperforming the behavior policy that generated the offline datasets. They show their approach's enhanced sample efficiency and performance through simulations involving robot manipulation tasks.

\section{Geometric Impedance Control: SE(3)-Equivariant Control Law}\label{sec:GIC}
In this Section, we introduce geometric impedance control (GIC) \cite{seo2023geometric, seo2024comparison} here and show that it serves as a $\SE$ group equivariant control law. 
Moreover, we also show that the GIC can be considered an $\SE$ group equivariant force-based policy. The main contents of this section are already published in \cite{seo2024comparison}. Readers can consider this section as the more detailed explanation version of \cite{seo2024comparison}.

\paragraph*{Necessity of geometric formulation}
As can be seen in \eqref{eq:manipulator_dynamics}, the underlying coordinate of the manipulator is described in the joint space. However, as suggested in \cite{khatib1987unified}, the manipulator will execute a task on the workspace, i.e., end-effector space. Therefore, the task can be described more intuitively in the end-effector space. The pose of the end-effector can be calculated via forward kinematics (Sec~\ref{sec:Prelim}), and thereby, it can be well described by utilizing a group structure of $\SE$. 

Conventional impedance control designs do not consider the full geometric structure of $\SE$. For instance, the papers from the early stage of robot control \cite{hogan1985impedance, khatib1987unified} could be considered. In \cite{hogan1985impedance}, the end-effector coordinate $x \in \mathbb{R}^6$ is just utilized without explicitly mentioning its definition. To correctly define the rotational errors, many representations are studied in \cite{caccavale1999six}, including Euler-angle representation, axis-angle representation of $\SO$, and quaternion representation. Similarly, in \cite{khatib1987unified}, the translational and rotational parts of the $x$ vector were separately considered. 

In this regard, by incorporating a full geometric structure of $\SE$ into the control design, one can expect a further increase in performance. 
Moreover, the intuition for the equivariant force or velocity-based policy will be clarified later in this section. 

\subsection{Manipulator Dynamics}
\subsubsection{Manipulator Dynamics in joint space}
The manipulator dynamics in the joint space is written as follows.
\begin{equation} \label{eq:manipulator_dynamics}
    M(q)\ddot{q} + C(q,\dot{q})\dot{q} + G(q) = \tau + \tau_e,
\end{equation}
where $q\in\mathbb{R}^n$ is a joint coordinate in position, $\dot{q} \in \mathbb{R}^n$ is joint velocity, $\ddot{q}\in\mathbb{R}^n$ is joint acceleration, $M(q)\in\mathbb{R}^{n\times n}$ is inertia matrix, $C(q,\dot{q})\in\mathbb{R}^{n\times n}$ is Coriolis matrix, $G(q) \in \mathbb{R}^n$ is gravity force vector in joint space, $\tau \in \mathbb{R}^n$ is joint torque as a control input, and $\tau_e \in \mathbb{R}^n$ is external disturbance acting in the form of joint torque. 

\subsubsection{Manipulator Dynamics in Workspace on $\SE$}
As the task that the manipulator performs is done by the end-effector, it is often preferable to define the task in the workspace, not in the joint space. Thus, we are now interested in the framework that regards the operational space as $\SE$. 
Employing the operational space formula from \cite{khatib1987unified}, the manipulator dynamics using the body-frame velocity $V^b$ is given by:
\begin{align} \label{eq:robot_dynamics_eef}
    &\tilde{M}(q)\dot{V}^b + \tilde{C}(q,\dot{q})V^b + \tilde{G}(q) = \tilde{T} + \tilde{T}_e, \text{ where}\\
   & \tilde{M}(q)= J_b(q)^{-T} M(q) J_b(q)^{-1}, \nonumber\\
    &\tilde{C}(q,\dot{q})=J_b(q)^{-T}(C(q,\dot{q}) \!-\! M(q) J_b(q)^{-1}\dot{J})J_b(q)^{-1}, \nonumber\\
   & \tilde{G}(q)= J_b(q)^{-T} G(q),\; \tilde{T}= J_b(q)^{-T} T,\; \tilde{T}_e= J_b(q)^{-T} T_e\nonumber,
\end{align}
where $J_b$ is a body-frame Jacobian matrix which relates body velocity and joint velocity by $V^b = J_b \dot{q}$. See \cite{seo2023geometric, seo2024comparison} for more details. 

\subsection{Error Functions: Distance-like Metrics on $\SE$}
Lyapunov stability theory provides a foundation for describing the stability property of the given autonomous or control system. 
The key idea behind the Lyapunov stability theory is to represent the dynamics of a high-dimensional nonlinear ordinary differential equation (ODE) without solving it. 
Instead, a scalar Lyapunov function is defined using the state of the ODE as input, and its convergence properties are analyzed to describe the behavior of the original system.
In mechanical control systems, the total mechanical energy, which is defined as the sum of potential and kinetic energy, is traditionally used as a Lyapunov candidate \cite{koditschek1989application}.
Here, the kinetic energy is composed of the velocity parts of the states, while the potential energy is composed of the position parts. It is quite straightforward to find out that the potential energy is closely related to the distance of two points, as the potential function and the distance metrics are both defined from the \emph{difference} of two points. 

Defining the difference between two points in Euclidean space is straightforward. Consider $x_1, x_2 \in \mathbb{R}^n$ in Euclidean space. Then, a \emph{vector} to describe the difference between two points can be considered as $v = x_1 - x_2$, and the Euclidean norm can be utilized to describe the distance, i.e., $d = \|v\|_2$.

On the other hand, the difference between the two points on the manifold (Lie group) is more subtle. Before we introduce the notion of difference in $\SE$, one must first note that the unit of the two elements $\SO$, $\mathbb{R}^3$ are not the same. 
As the unit of $\SO$ and $\mathbb{R}^3$, does not match, we have to use the notion of the \emph{error function}, $\Psi: \SE \times \SE \to \mathbb{R}$, instead of using the terminology of distance. For this reason, we will now use the word \emph{error function} to describe the scalar variable between two elements on $\SE$ that represents the ``distance'' between two points.
There are two major perspectives to represent such error function on $\SE$ \cite{seo2024comparison}: 


\begin{enumerate}
    \item Using the fact that $\SE$ is a General Linear (matrix) group
    \item Using the property of Lie algebra 
\end{enumerate}
Two different perspectives of error function lead to different perspectives of potential functions, resulting in different control laws. The first aspect leads to the definition of distance and potential function on the Lie group \cite{seo2023geometric, bullo1999tracking, lee2010geometric}. On the other hand, the Lie algebra-based potential function can be constructed \cite{teng2022lie, bullo1995proportional}. 
We note that the error function in this paper means the squared distance-like metric.

\subsubsection{Configuration Error}
The transformation matrix, a difference, between the desired $g_d = (p_d,R_d)$ and current configuration, $g = (p, R)$ is defined as
\[
    g_{de} = g_d^{-1}g.
\]
We will call $g_{de}$ as the \emph{configuration error}. This notion of configuration error has been utilized in many of the geometric control literature, e.g., \cite{seo2023geometric, bullo1995proportional, lee2010geometric,bullo1999tracking,teng2022lie,lee2012exponential}.

\subsubsection{Error Function from Matrix Group Perspective}
One can just focus on the fact that a homogeneous matrix representation is a matrix Lie group. This concept was first proposed in \cite{koditschek1989application}, further developed in \cite{bullo1999tracking}, revisited in \cite{lee2010geometric}, and widely utilized thereafter, especially on $\SO$, i.e., UAV field. In $\SO$, a error function on this perspective $\Psi(R,R_d)$ is given by
\[
    \Psi(R,R_d) = \text{tr}(I - R_d^T R)
\]
However, what has been overlooked from the error function $\Psi(R,R_d)$ is that it is identical to the Frobenius norm. In the $\SO$ case, the following equality holds \cite{huynh2009metrics}:
\begin{equation*}
    \Psi(R,R_d) = \text{tr}(I - R_d^T R) = \dfrac{1}{2}\text{tr} \left((I - R_d^T R)^T(I - R_d^T R)\right) 
\end{equation*}
The definition of the Frobenius norm, i.e., $\|A\|_F^2 = \text{tr}(A^T A)$, can be used to derive
\begin{equation} \label{eq:Frob_norm_SO3}
    \Psi(R,R_d) = \dfrac{1}{2}\|I - R_d^T R\|_F^2.
\end{equation}

Let us chew on \eqref{eq:Frob_norm_SO3}. When $R = R_d$, then $R_d^TR = I$; $\Psi(R,R_d) = \|0\|_F = 0$. Thus, what it means is that we first see the difference between two points in $\SO$, compare the difference rotation matrix with the identity matrix, and quantify its discrepancy from the identity as the Frobenius norm. Recalling that the transpose operator in $\SO$ is the same as the inverse operator, one could extend \eqref{eq:Frob_norm_SO3} defined on $\SO$ to the $\SE$ as the following formula, which is also based on the Frobenius norm. 
\begin{align} \label{eq:Frob_norm_SE3}
        &\Psi_1(g,g_d) = \dfrac{1}{2}\|I - g_d^T g\|_F^2 \\
        &= \tfrac{1}{2} \tr\left((I - g_d^{-1}g)^T(I - g_d^{-1}g)^T \right) = \tfrac{1}{2} \tr (X^T X) \nonumber,
\end{align}
with $X$ defined as
\begin{equation*}
    X = X(g,g_d) = \begin{bmatrix}
    (I - R_d^TR) & -R_d^T(p - p_d) \\
    0 & 0 
    \end{bmatrix}. \nonumber
\end{equation*}
By further unraveling the equation, one can show that
\begin{equation} \label{eq:error_function_SE3}
    \Psi_1(g,g_d) = \tr(I - R_d^TR) + \tfrac{1}{2}(p - p_d)^T(p - p_d)
\end{equation}
Therefore, we can see that the error function in $\SE$ is composed of $\SO$ and $\mathbb{R}^3$ similar to the first perspective. 

\subsubsection{Error Function from Lie Algebra Perspective}
The central intuition here is that the Lie algebra can be considered as the local chart of the $\SE$ manifold. Note that a Lie algebra is defined as the elements on the tangent space of the identity. Thus, a Lie algebra is a vector space but not (necessarily) attached to the tangent space of the current configuration.  As we can see from the Euclidean space example, the first intuitive way to define distance is by using a vector.
In \cite{teng2022lie}, the Lyapunov candidate was proposed using this concept. The implication is that the proposed Lyapunov candidate function is a potential function. If we remove the \emph{weights} in the potential, it can be considered the squared distance metric. The vector representation $\xi_{de} \in \mathbb{R}^6$ of a configuration error is defined as
\begin{equation}
    \xi_{de} = \log{(g_{de})}^\vee = 
    \begin{bmatrix}
        \hat{\psi}_{de} & b_{de} \\
        0 & 0
    \end{bmatrix}^\vee = \begin{bmatrix}
        b_{de} \\ \psi_{de}
    \end{bmatrix}
\end{equation}
where (see pp. 413-414 of \cite{murray1994mathematical})
\begin{align*}
    &\hat{\psi}_{de} = \log(R_d^TR), \quad b_{de} = A^{-1}(\psi_{de}) R_d^T(p - p_d)\\
    &A^{\text{-}1}(\psi) \!=\! I \!-\! \dfrac{1}{2}\hat{\psi} + \dfrac{2\sin\|\psi\| - \|\psi\|(1 + \cos\|\psi\|)}{2 \|\psi\|^2 \sin \|\psi\|} \hat{\psi}^2,
\end{align*}
when $\psi \neq 0$.
Note that the logarithm of the matrix is not unique \cite{Matrix_log_wiki}, and there are many alternative definitions, e.g., \cite{seo2024comparison, bullo1995proportional}
\begin{align*}
    A^{-1}(\psi) &= I - \dfrac{1}{2}\hat{\psi} + \left(1 - \alpha(\|\psi\|)\right) \dfrac{\hat{\psi}^2}{\|\psi\|^2},\\ &\text{with} \; \alpha(y) = (y/2)\cot{(y/2)}.
\end{align*}

As mentioned in \cite{bullo1995proportional}, the inner product on $\so$ is well defined. Moreover, it also has good properties of positive definiteness and bi-invariance. The inner product of two elements $\hat{\psi}_1, \hat{\psi}_2 \in \so$ can be defined using the Killing form, a bilinear operator defined by $\langle \cdot, \cdot \rangle _K: \so \times \so \to \mathbb{R}$, as \cite{park1995distance}
\begin{equation*}
    \begin{split}
        \langle\hat{\psi}_1, \hat{\psi}_2\rangle_{\so} &\triangleq -\tfrac{1}{4} \langle\hat{\psi}_1, \hat{\psi}_2\rangle_K, \;
        \langle\hat{\psi}_1, \hat{\psi}_2\rangle_K \triangleq \tr( \ad_{\psi_1} \ad_{\psi_2} ).
    \end{split}    
\end{equation*}
where $\ad$ is an adjoint (small adjoint) operator, where if $\psi \in \so$, $\ad_{\psi}= \hat{\psi}$.

On the other hand, there is no symmetric bilinear form on $\se$ that is both positive definite and bi-invariant \cite{bullo1995proportional, park1995distance}.
However, one can still define a left-invariant (or right-invariant) inner product on $\se$ by exploiting the fact that $\se$ is isomorphic to $\mathbb{R}^6$. We will denote the inner product by $\langle \hat{\xi}_1, \hat{\xi}_2\rangle_{(P,I)}$, as follows \cite{park1995distance, teng2022lie, vzefran1996choice, belta2002euclidean}:
\begin{equation}
\label{eq:inner_product_$\SE$}
    \langle \hat{\xi}_1, \hat{\xi}_2 \rangle_{(P,I)} = \xi_1^T P \xi_2
\end{equation}
where $\hat{\xi}_1, \hat{\xi}_2 \in \se$, and $P \in \mathbb{R}^{6\times 6}$ is symmetric positive definite matrix. 
The subscript $P$ in \eqref{eq:inner_product_$\SE$} denotes that the inner product is equipped with the Riemannian metric $P$, while the subscript $I$ means that the Lie algebra $\se$ is defined on the tangent space at the identity element of $\SE$. The introduction of these subscriptions is to fill the discrepancies between the definition of the inner product of Lie algebra and Riemannian metric; the inner product of the Lie algebra is defined literally on the Lie algebra, but the Riemannian metric is defined on the tangent space attached to each point on the manifold. 

An important factor to note is that the tangent vector on $g \in \SE$ is not the same as the Lie algebra $\se$. This can be revealed by observing $\dot{g}$, with $g = (p,R)$ and $V = (v,\hat{w})$:
\begin{equation*}
    \dot{g} = g \hat{V}^b = \begin{bmatrix}
        R & p \\
        0 & 1 
    \end{bmatrix} \begin{bmatrix}
        \hat{w} & v \\ 
        0 & 1
    \end{bmatrix} = \begin{bmatrix}
        R\hat{w} & Rv \\ 
        0 & 1
    \end{bmatrix},
\end{equation*}
but as $R\hat{w} \notin \so$, i.e., not skew-symmetric, we can conclude that $\dot{g} \notin \se$, but $\dot{g} \in T_g\SE$.

With this definition, we can define the Riemannian metric on $\SE$ on the point $g \in G$ 
using the inner product on $\se$, $\langle \cdot, \cdot \rangle_{(P,g)} : T_gG\times T_gG \to \mathbb{R}$, as follows. Consider two tangent \emph{vectors} on the tangent space at point g, $\delta g_i \in T_gG$, such that $\delta g_i = g \hat{\xi}_i$ with $\hat{\xi}_i = (b_i, \hat{\psi}_i)$ for $i = \{1,2\}$.
The Riemannian metric is then defined as \cite{park1995distance, vzefran1996choice, belta2002euclidean}
\begin{equation} \label{eq:Riemannian_metric}
    \langle \delta g_1, \delta g_2 \rangle_{(P,g)} \!=\! \langle g^{-1} \delta g_1, g^{-1} \delta g_2 \rangle_{(P,I)} = \langle \hat{\xi}_1, \hat{\xi}_2 \rangle_{(P,I)}.
\end{equation}
The Riemannian metric \eqref{eq:Riemannian_metric} herein is often referred to as a left-invariant Riemannian metric because $g^{-1} \delta g_i = \hat{\xi}_i$ is left invariant, i.e., for any arbitrary left transformation matrix $g_l$, $(g_l g)^{-1} (g_l g) \hat{\xi}_i = \hat{\xi}_i$. 

Finally, we can define an error function on $\SE$ from Lie-algebra perspective by letting $P = \tfrac{1}{2}I_{6\times6}$ as follows:
\begin{equation} \label{eq:error_function_SE3_algebra}
    \Psi_2(g,g_d) = \langle g_{de}\hat{\xi}_{de}, g_{de}\hat{\xi}_{de} \rangle_{\left(0.5I, g_{de}\right)} = \tfrac{1}{2} \|\psi_{de}\|^2 + \tfrac{1}{2}\|b_{de}\|^2,
\end{equation}
where $\hat{\xi}_{de} = \log(g_{de}) = (b_{de}, \hat{\psi}_{de})$. 
\eqref{eq:error_function_SE3_algebra} can also be obtained as presented in \cite{belta2002euclidean} and \cite{liu2013finite}, by considering $\SE$ as a general affine group. 

\subsection{Error Vectors on $\SE$}
\subsubsection{Positional Error Vectors on $\SE$}
\paragraph*{Geometrically Consistent Error Vector (GCEV)} GCEV was proposed in \cite{seo2023geometric} and serves as a positional error vector in $\SE$. Note that this error vector formulation is derived from the Lie group-based error function. The error vector derivation on $\SO$ was first proposed in \cite{lee2010geometric}. We first perturb a configuration $g \in \SE$ with a right translation in the direction of an arbitrary vector $\eta$, as was done in \cite{lee2010geometric} for $\SO$.
\begin{equation*}\begin{aligned}
    g_{\hat{\eta}}= g e^{\hat{\eta}\epsilon}\!, \quad \eta=\begin{bmatrix}\eta_1\\ \eta_2\end{bmatrix}\!\in\!\mathbb{R}^6 \implies \hat{\eta}\!\in\! \SE, \;\epsilon\!\in\!\mathbb{R},
\end{aligned}\end{equation*}
where $\eta_1, \eta_2\!\in\!\mathbb{R}^3$ refer to translational and rotational components, respectively. The perturbed configuration matrix $\delta g$ can be found by
\begin{equation*}
    \begin{split}
        \delta g &= \left.\dfrac{d}{d\epsilon}(g e^{\hat{\eta}\epsilon})\right|_{\epsilon=0} = \left. g \hat{\eta}  e^{\hat{\eta}\epsilon}\right|_{\epsilon = 0} = g \hat{\eta} \in T_g \SE, \\
        &\implies \begin{bmatrix}
        \delta R & \delta p \\
        0 & 0 
        \end{bmatrix}  = \begin{bmatrix}
        R\hat{\eta}_2 & R \eta_1 \\ 0 & 0 
        \end{bmatrix}.
    \end{split}    
\end{equation*}
The perturbed error function $\delta \Psi(g,g_d)$ can be obtained and evaluated by
\begin{equation*} \label{eq:perturbation_of_error_metric}
    \begin{split}
    \delta \Psi(g,g_d)&=\dfrac{1}{2}\tr\begin{bmatrix}\!
    -\delta R^T R_d-R_d^T \delta R & \ast \nonumber
    \\
    \ast &  2(p-p_d)^T \delta p
    \end{bmatrix} \nonumber \\
    &=\!\dfrac{1}{2}\tr\begin{bmatrix}
    -\hat{\eta}_2^T R^TR_d-R_d^T R \hat{\eta}_2 & \ast
    \\
    \ast &  2(p-p_d)^T R \eta_1 
    \end{bmatrix}\nonumber \\
    &= -\tr(R_d^T R \hat{\eta}_2) + (p-p_d)^T R \eta_1 \nonumber \\
    &= (R_d^T R - R^T R_d)^\vee \cdot \eta_2 + (p - p_d)^T R \eta_1,
    \end{split}
\end{equation*}
where $\cdot$ is an inner product between two vectors. We can describe the perturbation of the error function in $\SE$ $\delta \Psi$ using the Lie derivative from \eqref{eq:lie_deriv_group}:
\begin{equation*}
    \delta \Psi = \Lie_{\hat{\eta}} \Psi(g,g_d) = \dfrac{d}{d\varepsilon} \left. \Psi(g e^{\hat{\eta}\varepsilon}, g_d)\right|_{\varepsilon = 0}.
\end{equation*}
Note also that we only evaluate the Lie derivative on the point $g$, not the point $g_d$.

We now define the \emph{position} error vector or Geometrically Consistent Error Vector (GCEV) $\eg$
as
\begin{equation}\begin{aligned}\label{eq:eg}
    \eg = \begin{bmatrix}
    e_p \\ e_R
    \end{bmatrix} = \begin{bmatrix}
    R^T(p - p_d) \\
    (R_d^T R - R^T R_d)^\vee 
    \end{bmatrix} \in \mathbb{R}^6,
\end{aligned}\end{equation}
so that $\delta \Psi(g,g_d) = \eg ^T \eta$ and
\begin{equation*}\begin{aligned}
    \hat{e}_g = \begin{bmatrix}
    R_d^T R - R^T R_d & R^T (p - p_d) \\
    0 & 0 
    \end{bmatrix} \in \se.
\end{aligned}\end{equation*}

\paragraph*{Lie algebra-based Error Vector}
Similar to the way that the GCEV is derived from its Lie group-based error function, we can also derive the positional error vector from the Lie algebra-based error function. To put a conclusion, such a positional error vector is a log map of the configuration error $g_{de}$, i.e., $\xi_{de} = \log(g_{de})^\vee$ \cite{teng2022lie}.

\subsubsection{Velocity Error Vector}
Velocity error vector should be defined properly on the $\SE$ manifold, as the two tangent vectors $\dot{g} = g \hat{V}^b$ and $\dot{g}_d = g_d \hat{V}^b_d$ do not lie on the same tangent spaces. Specifically, the desired velocity vector $V_d^b$ is translated using the following formula (see Ch 2.4 of \cite{murray1994mathematical} for the details. See also \cite{seo2023geometric, lee2010geometric}): 
\begin{equation*}\begin{aligned}
    \hat{V}_d^* = g_{ed} \hat{V}_d^b g_{ed}^{-1}, \text{ where } g_{ed} = g^{-1}g_d.
\end{aligned}\end{equation*}
Then, it is straightforward to have in the vector form:
\begin{equation}\begin{aligned}\label{eq:Ad_velo}
    V_d^* = \Ad_{g_{ed}}V_d^b, \text{ with } \Ad_{g_{ed}} = \begin{bmatrix}
   R_{ed} & \hat{p}_{ed} R_{ed} \\ 0 & R_{ed}
    \end{bmatrix},
\end{aligned}\end{equation}
where $\Ad_{g_{ed}}: \mathbb{R}^6 \to \mathbb{R}^6$ is an Adjoint map, $R_{ed} = R^T R_d$, and $p_{ed} =-R^T(p - p_d)$. 

Based on this, we define the \emph{velocity} error vector $\ev$ as 
\begin{equation}\begin{aligned} \label{eq:ev}
    \ev = \underbrace{\begin{bmatrix}
    v^b \\ w^b \end{bmatrix}}_{V^b} - \underbrace{\begin{bmatrix}
     R^TR_d v_d + R^T R_d \hat{\omega}_d R_d^T (p - p_d)\\
     R^T R_d \omega_d \end{bmatrix}}_{V_d^*} = \begin{bmatrix}
    e_v \\ e_\Omega \end{bmatrix}.
\end{aligned}\end{equation}

This procedure to calculate velocity error vector originates from the fact that the two tangent vectors $\dot{g} \in T_g\SE$ and $\dot{g}_d \in T_{g_d}\SE$ cannot be directly compared since they are defined on the different points on the manifold. In \cite{lee2010geometric}, $\dot{g}_d$ is first translated via right multiplication $g_d^{-1}g$. This right multiplication can be interpreted in the following procedures, with $\dot{g}_d = g_d \hat{V}_d^b$:
\begin{enumerate}
    \item Apply left transformation on tangent space $dL_{g_d^{-1}}(\dot{g}_d)$by $g_d^{-1}\dot{g}_d = \hat{V}_d$.
    \item Apply coordinate transformation on $\se$ via $\hat{V}_d^{*} = g_{ed} \hat{V}_d^b g_{ed}^{-1}$, with $g_{ed} = g^{-1}g_d$
    \item Translate to the point $g$ by $dL_g(\hat{V}_d^*) = g \hat{V}_d^{*}$
    \item Compare two tangent vectors $g \hat{V}^b - g \hat{V}_d^{*} = g \hat{e}_V$
\end{enumerate}

The other perspective is to interpret the configuration error $g_{de}$ as an element on the manifold $\SE$, and the time-derivative of $g_{de} = g_d^{-1}g$ is handled as follows \cite{teng2022lie}:
\begin{align*}
    \dfrac{d}{dt}(g_d^{-1}g) &=  g_d^{-1}\dot{g} + \dot{g}_d^{-1} g = g_d^{-1}\dot{g} -  g_d^{-1}\dot{g}_d g_d^{-1} g \nonumber \\
    &= g_d^{-1}g \hat{V}^b - g_d^{-1}g_d\hat{V}_d^b g_d^{-1}g \\
    &= g_d^{-1}g \left(\hat{V}^b - (g_d^{-1}g)^{-1}\hat{V}_d^b (g_d^{-1}g)\right) \triangleq g_d^{-1}g (\hat{e}_V) \nonumber\\
    \implies \dot{g}_{de} &= g_{de} \hat{e}_V \in T_{g_{de}}\SE\nonumber
\end{align*}
Notice that $g_{de} = g_d^{-1}g$, $g_{de}^{-1} = g_{ed}$, and $(g \hat{\xi} g^{-1})^\vee = \Ad_g\xi$, for $g \in \SE$ and $\xi \in \se$. Also, $\tfrac{d}{dt}M^{-1} = M^{-1} \tfrac{dM}{dt} M^{-1}$. To conclude, the tangent vector is directly derived on the tangent space $\dot{g}_{de} \in T_{g_{de}}\SE$, and it turns out that both approaches lead to the same velocity error vector.

Therefore, we conclude that although both approaches have different rationales, they share the same structure of velocity error vector, which leads to the same Kinetic energy function and the same impedance control law. We refer to the description from \cite{seo2024comparison} for the details of the velocity error vector.

\subsection{Energy Functions on $\SE$}
We consider two energy functions for our control design: the potential energy function and the kinetic energy function.

\subsubsection{Potential Energy Function} 
We interpret the potential energy function as a weighted error function. This concept is motivated by the potential energy of the spring in the Euclidean space. Consider a linear spring with one end attached to the origin $x_0 = 0 \in \mathbb{R}^3$ and the other end attached to $x \in \mathbb{R}^3$. Then, the potential energy stored in the spring can be written by 
\begin{equation*}
    P = \frac{1}{2} x^T K x = \frac{1}{2}\|x\|_K^2,
\end{equation*}
where $K \in \mathbb{R}^{3\times 3}$ is a symmetric positive definite stiffness matrix, and the potential energy can be interpreted as the weighted 2-norm or weighted squared Euclidean distance. Recalling that we have derived the error metric on $\SE$ in two different ways, we will define the potential functions on $\SE$ accordingly. 

\paragraph*{Potential Energy Function from Matrix Group Perspective} We weight the distance metric in \eqref{eq:Frob_norm_SE3} by first introducing a weighted matrix $X_K$ given by:
\begin{equation}
    X_K = \begin{bmatrix}
        \sqrt{K_R}(I - R_d^TR) & -\sqrt{K_p}R_d^T(p - p_d)\\
        0 & 0
    \end{bmatrix},
\end{equation}
where $K_p, K_R \in \mathbb{R}^{3\times3}$ are symmetric positive definite stiffness matrices for translational and rotational parts, respectively, and the square root of the matrix can be defined from the standard singular value decomposition. Note that the square root of the matrix is utilized to recover the original matrix in the potential energy form.
Then, the potential function $P_1(g,g_d)$ from Lie group $\SE$ perspective is defined in the following way:
\begin{equation} \label{eq:potential_lie_group}
    \begin{split}
        & P_1(g,g_d) = \tfrac{1}{2}\tr(X_K^T X_K) \\
        &\;\;= \tr(K_R(I - R_d^T R)) + \tfrac{1}{2}(p - p_d)^T R_d K_p R_d^T (p - p_d).
    \end{split}
\end{equation}

\paragraph*{Potential Energy Function from Lie algebra perspective}
Given the inner product on $\se$ and the error function defined previously, the potential function from the Lie algebra perspective can be given as \cite{teng2022lie, seo2024comparison}:
\begin{equation} \label{eq:potential_lie_algebra}
    P_2(g,g_d) = \langle \hat{\xi}_{de}, \hat{\xi}_{de}\rangle_{(0.5K_\xi, I)} = \tfrac{1}{2}\xi_{de}^T K_{\xi} \xi_{de},
\end{equation}
where $P=\tfrac{1}{2}K_{\xi}$ and $K_\xi \in \mathbb{R}^{6\times6}$ is symmetric positive definite. Henceforth, we interpret $K_\xi$ as a stiffness matrix. We emphasize that $P_1(g,g_d)$ and $P_2(g,g_d)$ are left-invariant, positive definite, and quadratic. Left-invariance of these potential functions will be utilized later in this Section to prove the $\SE$-Equivariance. 

\subsubsection{Kinetic Energy on $\SE$}
The kinetic energy function on $\SE$ for the impedance control design can be formulated by using a velocity error vector $\ev$ and inertia matrix in the following form:
\begin{equation} \label{eq:kinetic_energy}
    K(t,q,\dot{q}) = \tfrac{1}{2}e_V^T \tilde{M} e_V,
\end{equation}

\subsection{Geometric Impedance Control}
We now introduce a geometric impedance control law (GIC) based on the potential and kinetic energy functions. The main idea behind the GIC is that the total mechanical energy function (potential + kinetic energy) should be dissipated over time. 

\subsubsection{Impedance Control as a Dissipative Control Law}
The total mechanical energy function, our Lyapunov function is defined as
\begin{equation} \label{eq:lyapunov}
    V_i(t,q,\dot{q}) = K(t,q,\dot{q}) + P_i(t,q), \quad i \in \{1,2\},
\end{equation}
where $K$ and $P_i$ are the kinetic and potential energy components of the Lyapunov function. Note that $P_1$ is given in \eqref{eq:potential_lie_group} and $P_2$ is in \eqref{eq:potential_lie_algebra}. The desired property of the impedance control to be dissipative control law \cite{kronander2016stability} is defined as
\begin{equation} \label{eq:dissipative}
    \dot{V}_i = -e_V^T K_d e_V,
\end{equation}
where $K_d \in \mathbb{R}^6$ is symmetric positive definite damping matrix. 
To satisfy \eqref{eq:dissipative}, the impedance control laws for each potential function are
\begin{equation} \label{eq:control_law}
        \tilde{T}_i \!=\! \tilde{M}\dot{V}^*_{d} \!+\! \tilde{C}{V}^*_{d} + \tilde{G} \!-\! f_{_{G,i}} \!-\! K_d e_V, \;\text{for}\; i \in \{1,2\},
\end{equation}
where $f_{_{G,i}} \in \mathbb{R}^6$, $\hat{f}_{_{G,i}}\in \se^*$, are given respectively by
\begin{equation}
    \begin{split}
        f_{_{G,1}} &= \begin{bmatrix}
        f_p \\ f_R
        \end{bmatrix} = \begin{bmatrix}
            R^T R_d K_p R_d^T (p - p_d)\\
            (K_R R_d^T R - R^T R_d K_R)^\vee
        \end{bmatrix} \\ 
        f_{_{G,2}} &= K_\xi \xi_{de}.
    \end{split}    
\end{equation}
The derivation process of $f_{_{G,1}}$ is shown in \cite{seo2023geometric} and $f_{_{G,2}}$ in \cite{seo2024comparison}.
We can show that both control laws satisfy almost global asymptotic stability. We refer to \cite{seo2023geometric, seo2024comparison} for the details of such properties.

\subsubsection{Geometric Impedance Control is $\SE$-Equivariant Control}
One of the crucial advantages of the GICs is that the control law is $\SE$ equivariant if it is described on the spatial frame. In \cite{seo2023contact}, the GIC law with learning gain scheduling policy is proved to be equivariant if the gain scheduling policy is left-invariant. Without any gain scheduling law, i.e., fixed impedance gains, similar analysis is also possible. Take $f_{_{G,1}} : \SE \times \SE \to se^*(3)$ for example, the left-invariance of $f_{_{G,1}}(g,g_d)$ can be shown as follows (see also Fig.~\ref{fig:pih_transformed}):
\begin{align}
    \label{eq:geometric_elastic}
    &f_{_{G,1}}(g_lg,g_lg_d) \nonumber\\&= \begin{bmatrix}
        (R_l R)^T R_l R_d K_p (R_l R_d)^T (R_l p + p_l - R_l p_d - p_l)\\
        (K_R(R_l R_d)^TR_lR - (R_l R)^T R_l R_d K_R)^\vee
    \end{bmatrix} \nonumber \\
    &= \begin{bmatrix}
        R^T R_d K_p R_d(p - p_d)\\
        (K_R R_d R - R^T R_d K_R)^\vee
    \end{bmatrix} = f_{_{G,1}}(g,g_d).
\end{align}
Note that the elastic force $f_{_{G,1}}$ is described on the body-frame coordinate. 
\begin{figure}[t]
    \centering
    \begin{subfigure}{0.45\columnwidth}
        \centering
        \includegraphics[width=0.85\linewidth]{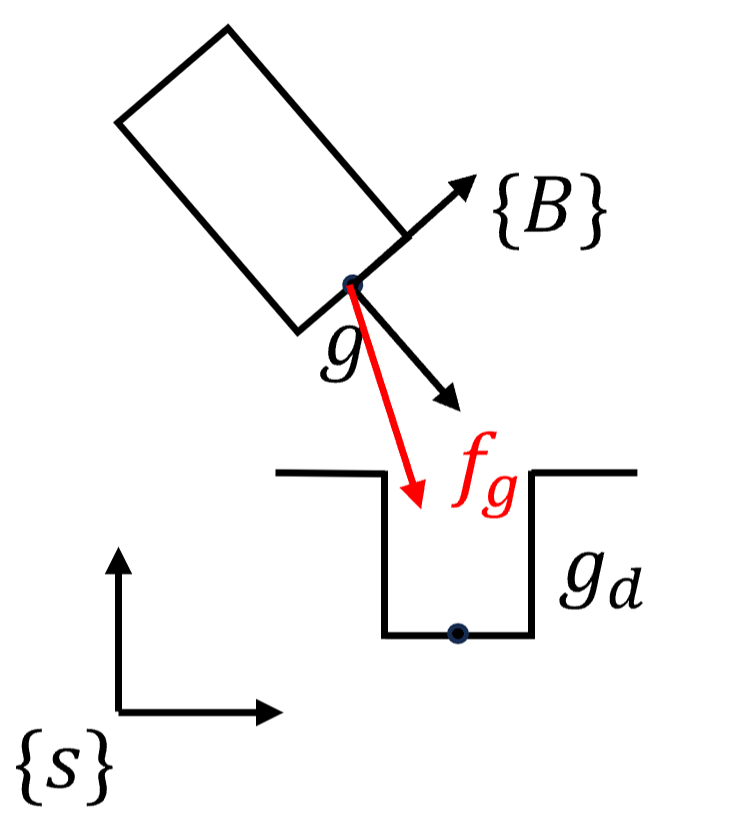}
    \end{subfigure}
    \begin{subfigure}{0.45\columnwidth}
        \centering
        \includegraphics[width=0.85\linewidth]{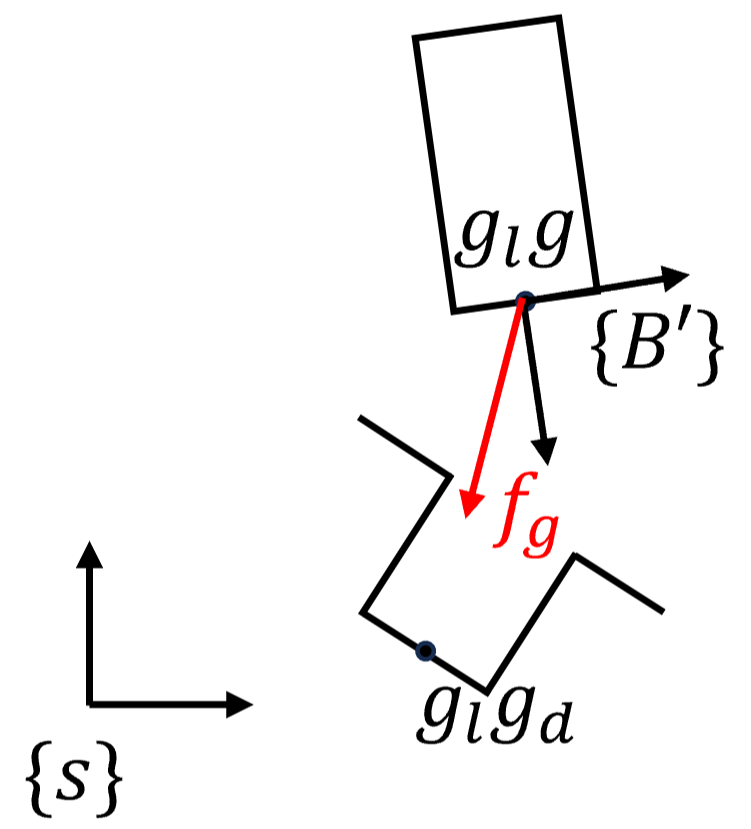}
    \end{subfigure}
    \vspace{-3pt}
    \caption{(Left) Peg and hole configurations are represented by $g$ and $g_d$, respectively. (Right) Peg and hole undergo left transformation via an action $g_l \in \SE$. The GCEV $\eg$ \eqref{eq:eg} and elastic force $f_{_{G,1}}$ \eqref{eq:geometric_elastic} are invariant to the left transformation of $\SE$. $\{s\}$ represents spatial coordinate frame, while $\{B\}$ and $\{B'\}$ represent body coordinate frames.}
    \label{fig:pih_transformed}
\end{figure}
Similarly for $f_{_{G,2}}$,
\begin{equation}
    \begin{split}
        f_{_{G,2}}(g_lg, g_lg_d) &= K_\xi \log{((g_l g_d)^{-1} (g_l g_e))} \\
    &= K_{\xi} \log{(g_d^{-1} g_e)} = K_{\xi}\xi_{de}.
    \end{split}
\end{equation}

Then, the feedback terms of the GIC law are $\SE$ equivariant if they are described in the spatial frame, as can be shown from Proposition 2 of \cite{seo2023contact}. We will skip the proof in this note, as it has already been proved in the paper. 
However, it is worth mentioning that a generalized recipe for the velocity and force-based policy to be $\SE$-equivariant can be obtained. According to \cite{seo2023contact}, such a general recipe is given by:
\begin{enumerate}
    \item The policy is left-invariant.
    \item The policy is described on the body-frame coordinate.
\end{enumerate}
Although in \cite{seo2023contact} only mentioned velocity and force-based policy, it can be trivially extended to displacement $\Delta g$ within a sampling period. 

We have added a few comments regarding the right invariant control design in Appendix~\ref{appendix:right_invariance}.

\section{Future Works} \label{sec:Future_work}
\subsection{Towards Vision-to-Force $\SE$-Equivariance}
Current vision-based approaches are formulated on the philosophical foundation that every manipulation can be considered as the sequence of pick-and-places \cite{shridhar2022cliport}. Although this argument well describes the manipulation tasks for simple tabletop manipulation, many real-life tasks require force interaction and compliance. For example, if a robot wants to twist off a bottle, the simple vision-based model cannot obtain the necessary accuracy and handle delicate force interaction. Many examples of assembly tasks also require force interaction, such as bolt-nut assembly or brute-force tight-fit assembly. Forces, or wrenches, can be interpreted as two type-1 vectors steered by the large adjoint map on the dual space of $\SE$ manifold. The handling of such wrenches is implied in the diffusion process in Diff-EDF \cite{ryu2024diffusion} but for the twists rather than wrenches. Therefore, we hypothesize that incorporating equivariance can significantly enhance the sample efficiency of the learning problem for the manipulation tasks associated with force interaction. The GIC proposed in \cite{seo2023geometric, seo2024comparison} and its application to assembly tasks \cite{seo2023contact} can be considered as a good baseline for such force-related equivariant assembly method.

\subsection{Symmetry Breaking in Robotics and  Systems}
The robot's end-effector has an $\SE$ manifold structure with symmetry properties that can be exploited. However, the symmetry in the end-effector space of the manipulator only holds when there is enough manipulability. For instance, if the manipulator approaches the singular configuration, the dynamics cannot be well-described by \eqref{eq:robot_dynamics_eef}. Therefore, the kinematic constraint is the main reason for symmetry breaking.
From the control perspective, such symmetry-breaking scenarios can be easily observed when there are constraints in the control system, e.g., constraints on the input. 

Symmetry breaking can arise from imperfections in the observation space, such as images captured from a 3D scene using a camera. Equivariant networks, which model rotational symmetry from 2D inputs, typically rely on a perfect top-down view to fully exploit this symmetry. However, practical deviations, such as tilted camera angles, disrupt the assumed symmetry, resulting in suboptimal performance compared to ideal settings. While equivariant networks outperform unstructured networks under these conditions \cite{wang2022surprising}, they remain sensitive to such imperfections. Additionally, occlusions in the observation space can further disrupt symmetry, diminishing the ability to leverage it effectively.

Symmetry breaking also occurs in equivariant reinforcement learning based on group-invariant MDP (Section \ref{rl}). Group-invariant MDP assumes group-invariant reward functions and transition probabilities; however, these can be corrupted by factors such as the kinematic constraints of a robotic manipulator.

\section{Conclusions} \label{sec:Conclusion}
In this tutorial survey paper, recent advances in geometric deep learning and control for robotics applications are reviewed. 
The relevant mathematical backgrounds of the Special Euclidean group, i.e., $\SE$ manifold structure, are first presented in the manipulator's end-effector. 
Next, we review the formulation and relevant theories of geometric deep learning, focusing on $\SE$-equivariant deep learning, together with the introduction of backbone neural networks. 
We have provided recent works in $\SE$-equivariant models utilized for two major applications in robotics, namely, imitation learning and reinforcement learning. 
Furthermore, the $\SE$-equivariant control method is presented, which may serve as a low-level control layer for $\SE$-equivariant learning modules in robotics applications.
Finally, the limitations and future works of equivariant methods in robotics are presented.  

\section{Conflict of Interest}
Joneun Choi
is a Senior Editor of International Journal of Control, Automation,
and Systems. Senior Editor status has no bearing
on editorial consideration. The authors declared no potential conflicts of interest with respect to the research, authorship, and/or publication of this paper.

\appendix
\subsection{Smooth Manifolds} \label{sec:appendix_smooth}
As the concept of smooth manifolds provide fundamental theoretical background, we include them for the completeness of the paper.
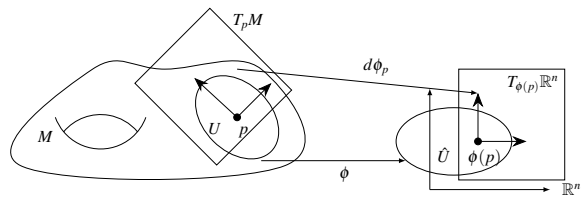
\begin{figure}[!t]
\centering
\resizebox{\columnwidth}{!}{
\begin{tikzpicture}[>=latex]
\draw[smooth cycle,tension=.7] plot coordinates{(-1,0) (0.5,2) (2,2) (4,2) (4.5,0)};
\coordinate (A) at (-.1,.7);
\draw  (A) arc(140:40:1) (A) arc(-140:-20:1) (A) arc(-140:-160:1)
node[midway, below left]{$M$};

\coordinate (EllipseCenter) at (3.5,1);
\draw[rotate around={45:(EllipseCenter)}] (EllipseCenter) ellipse (0.7cm and 1cm);
\draw[rotate around={45:(EllipseCenter)}] ([shift={(-1.0cm,-0.4cm)}]EllipseCenter) rectangle ([shift={(1.2cm,2cm)}]EllipseCenter) 
node[below right]{$\quad T_pM$};

\coordinate (Rn) at (8,0.5);
\draw (Rn) ellipse (1.2cm and 0.7cm);
\draw ([shift={(0.1cm,-0.8cm)}]Rn) rectangle ([shift={(2.3cm,1.5cm)}]Rn) 
node[below left]{$T_{\phi(p)} \Real^n$};

\begin{scope}[shift={(EllipseCenter)}, rotate=45]
    \draw[-{Stealth[length=8pt]},thin] (EllipseCenter) -- (-0.05,1.2);
    \draw[-{Stealth[length=8pt]},thin] (EllipseCenter) -- (1, 0);
\end{scope}
\filldraw[black] (EllipseCenter) circle (2pt) node[below]{$U \quad p \quad$};

\draw[-{Stealth[length=8pt]},thin] (8.5,0.5) -- (9.5,0.5);
\draw[-{Stealth[length=8pt]},thin] (8.5,0.5) -- (8.5,1.5);
\filldraw[black] (8.5,0.5) circle (2pt) node[below]{$\hat U \quad \phi(p) \quad$};

\draw[thin, ->] (7.5, -0.5) -- (10, -0.5) node [label=right:$\mathbb{R}^n$] {};
\draw[thin, ->] (7.5, -0.5) -- (7.5, 1.6) ;

 \draw[thin, ->] (3.5, 2) -- (8.5, 1.5) node [midway,above right] {$d \phi_p$};

 \draw[thin, ->] (4, 0.1) -- (7, 0.1) node [midway,below right] {$\phi$};
 
\end{tikzpicture}
}
\caption{Tangent vectors in a coordinate chart}
\label{fig:Chart}
\end{figure}

\subsubsection{Tangent and Cotangent Spaces} The tangent space $T_pM$ at a point $p$ on a smooth manifold $M$ is defined as the set of all tangent vectors to $M$ at $p$.  $T_pM = \{v \in M : v\text{ is a tangent vector to }M\text{ at }p\}.$
The elements of the tangent space are tangent vectors. Assume that there is a coordinate chart on $M$ with local
coordinates $(x^1, \cdots, x^n)$. The set of derivations  $\{ \frac{\partial }{\partial x^1}, \cdots, \frac{\partial }{\partial x^n} \}$ then form a basis
for $T_pM$, i.e.,
\[
X_p=\sum_{i=1}^nX^i\frac{\partial }{\partial x^i} = X^i\frac{\partial }{\partial x^i},
\]
where $X^i$ is the component of the tangent vector $X_p$ at $p$ along the derivation $\frac{\partial }{\partial x_i}$.
In the second term of the equation, we have eliminated the summation sign by following the so-called Einstein convention:  if the same index name (such as $i$ in the expression above) appears exactly twice in any monomial term, once as an upper index and once as a lower index, that term is understood to be summed over all possible values of that index, generally from 1 to the dimension of the space in question. We also interpret that an upper index in a denominator becomes a lower index, i.e. the index $i$ in $\frac{\partial }{\partial x^i}$ is considered a lower index.


The cotangent space $T^*_pM$ at a point $p$ on a smooth manifold $M$ is defined as the dual space of the tangent space at $p$, i.e.,  
${\displaystyle T_{p}^{*}\!{{M}}=(T_{p}{{M}})^{*}}$.
It consists of all linear functionals on the tangent space at $p$. i.e., 
$T^*_pM = \{\alpha : T_pM \rightarrow \mathbb{R}, \; \alpha\text{ is linear}\}.$
The element  of $T_p^{*}M$ are called
a dual vector, cotangent vector or, in the context of differential
forms, a one-form at $p$.
%
The action of a cotangent vector $\alpha_p \in T_p^*M$ on a tangent
vector $X_p\in T_pM$ is denoted by $  \alpha_p, (X_p)  .$ The dual basis   for $T_p^*M$ are given by
$\{dx^1, \cdots, dx^n\}$, and their actions on the  the basis for $T_p M$ are
\[
 dx^i \left ( \frac{\partial }{\partial x_j} \right )   =\delta^i_j = 
 \left \{ \begin{array}{cc} 1 \:\:\mbox{if} \:\: i=j\\ 0 \:\:\mbox{if}\:\: i\ne j\end{array} \right ., \:\:\quad i, j = 1, \cdots, n.
\]
An arbitrary one-form $\alpha_p$ at point $p$ is written as $\alpha_p= \alpha_i dx^i,$ were again we are following the Einstein summation convention. Therefore,
\[
\alpha_p(X_p) = \alpha_i dx^i \left ( X^j \frac{\partial}{\partial x^j} \right ) = \alpha_i X^i\,.
\]

Given a smooth function $f: M \rightarrow \Real$, the action of the tangent vector $X_p$ on $f$ is 
\[
X_p(f) = X^i \left . \frac{\partial f(x)}{\partial x_i} \right |_{p}\,.
\]
The differential of $f$, $df_p  = \left . \frac{\partial f(x)}{\partial x_i} \right |_{p} dx^i\in T^*_p M$  satisfies
\[
 df_p(X_p) = X_p(f), \quad X_p \in T_p M.
\]

The tangent
bundle of $M$, denoted by $TM$, is defined as the disjoint union of the tangent spaces at all
points of $M$  such as 
\[TM=\bigsqcup_{p \in M} T_p M,\]
where $\sqcup$ is a disjoint union of sets. The tangent bundle comes equipped with a natural projection map $\pi : T_p M \to M$, which sends each vector in $T _p M$ to the point $p$ at which it is tangent: $\pi(X_p) = p$.
Similarly, the cotangent bundle of $M$ is denoted by   $T^*M$.




\subsubsection{The Differential of a Smooth Map} Suppose $M$ and $N$ are smooth manifolds. Let $\F: M \rightarrow N$ be smooth and be a diffeomorphism. Also, let $f: N \to \Real$ be also smooth.
Given  ${\displaystyle p \in M,}$ the differential of 
$\F$  at 
$p$
is a linear map
 \[ {\displaystyle d\F _{p}\colon \ T_{p}M\to T_{\F (p)}N\,}\]
from the tangent space of 
$M$ at 
$p$ (i.e., $T_{p}M)$
 to the tangent space of 
$N$ at 
${\displaystyle \F (p)}$ (i.e., $T_{\F (p)}$) which satisfies
\[
dF_p(X_p)(f) = X_p(f \circ F),
\]
where $\circ$ is a function composition operator, e.g., $(h\circ g)(x) = h(g(x))$. As mentioned in Section~\ref{sec:lie_group}, $\circ$ will be mainly used for the composition of functions in the Appendix \ref{sec:appendix_smooth}.
The differential $d\F_p$, is  the best linear approximation of $\F$ near $p$. 
Let $(x^i, \cdots , x^n)$ and $(y^i, \cdots, y^n)$ be respectively  local coordinates of  manifolds  $M$ and $N$. Then the action of $dF_p(X)$ can be expressed in local coordinates as
\begin{equation}
\label{eq:pushforward_coordinates}
dF_p(X_p) =  \left . \left ( \frac{\partial F^j(x)}{\partial x^i} X^i  \right ) \right |_p  \left . \frac{\partial}{\partial y^j} \right |_{F(p)} \:\: \in T_{F(p)}N\,.
\end{equation}
and, for $f : N \to \Real$,
\[
dF_p(X_p)(f) =  \left . \left ( \frac{\partial F^j(x)}{\partial x^i}  X^i \right ) \right |_p  
 \left . \frac{\partial f(y)}{\partial y^j} \right |_{F(p)}  \in \Real\,.
\]

 


\subsubsection{Vector Fields} \label{sec:appendix_SH} A smooth vector field $X$ on a manifold $M$ is defined as a smooth map $X: M \rightarrow TM$
 with $\pi \circoo X= \Id$, where $\pi: TM \rightarrow M$ is the previously defined projection map. Let $\mathfrak{X}(M)$ be the set of all smooth vector fields on $M$.
A vector field can be expressed in terms of the local coordinate basis  of a chart $(\phi, U)$ as
\[
X_p= X^i(x) \left . \frac{\partial }{\partial x_i} \right |_p = X^i(p) \left . \frac{\partial }{\partial x_i} \right |_p 
\]
where each $X^i$ is a smooth function on an open ball of $x=\phi(p)$ 
(see Fig.~\ref{fig:Chart}).


\begin{figure}[!t]
\begin{center}
\includegraphics[width=0.8\columnwidth]{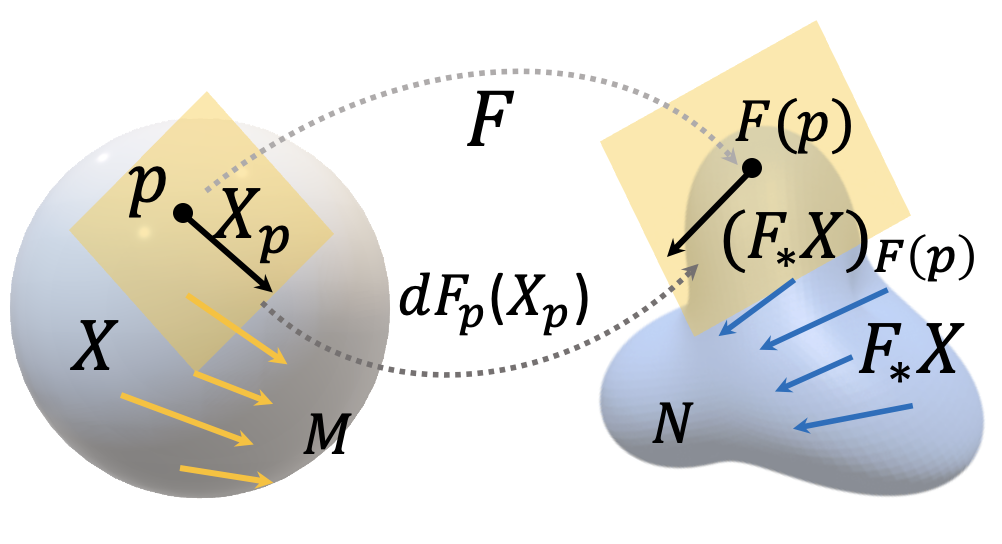} 
\vskip -0.5pc
\caption{The pushforward of the vector field X. 
}\label{fig:Pushforward}
\end{center}
\vskip -1.5pc
\end{figure}

\begin{figure}[!t]
\begin{center}
\includegraphics[width=0.8\columnwidth]{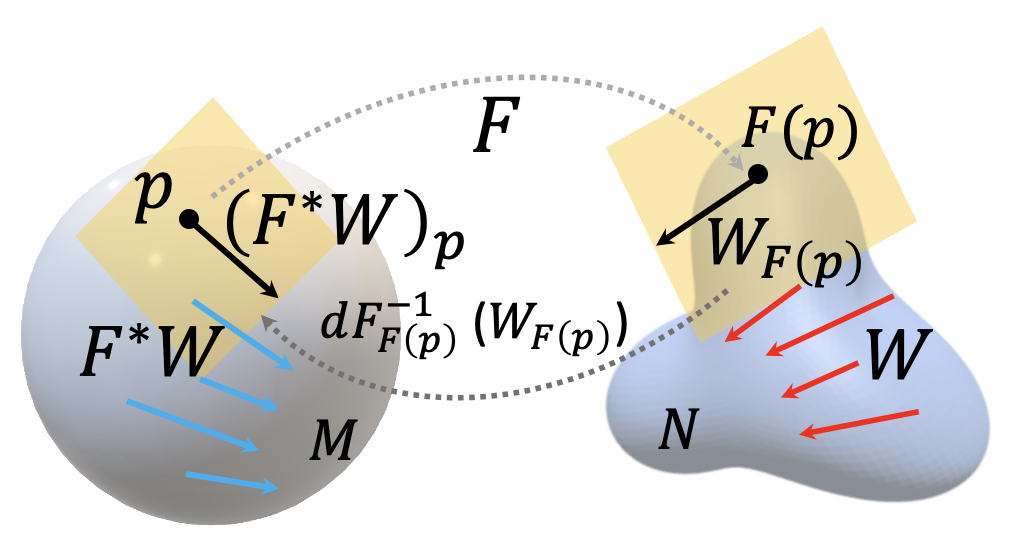} 
\vskip -0.5pc
\caption{The pullback of the vector field W. 
}\label{fig:Pullback}
\end{center}
\vskip -1.5pc
\end{figure}

\subsubsection{Pushforwards and Pullbacks}
\paragraph*{Pushforward}  Let $F : M \rightarrow N$ be a smooth mapping between manifolds and $X\in \mathfrak{X}(M)$ and $Y \in \mathfrak{X}(N)$ smooth vector fields. $X$ and $Y$ are $\F$-related if 
\[
Y_{F(p)}=dF_p (X_p) 
\]
where $dF_p$ is the differential of $F$ at $p$.
If $F$ is a diffeomorphism (i.e. $F$ has a differentiable inverse in addition to being smooth), for every $X\in \mathfrak{X}(M)$, there is a unique smooth vector field on $N$ that is F -related to $X$ \cite{leesmooth}. This unique  F-related field $Y \in \mathfrak{X}(N)$ is denoted as $Y = F_* X$ and is called the pushforward of $X$ by $F$.

The push-forward takes tangent vectors at $p \in M$  and "pushes'' them forward to tangent vectors at the corresponding point $q = F(p)$ in   $N$ as shown in Fig.~\ref{fig:Pushforward}.
 The  formula for the pushforward of $X$ by $\F$ at $q$ is given by 
\[
(\F_\ast X)_q  = dF_{\F^{-1} (q)} ( X_{\F^{-1}(q)} ) = d\F_p(X_p)
\]
and is given in explicit coordinates by \eqref{eq:pushforward_coordinates}.

\paragraph*{Pullback} 
Let $\F: M \rightarrow N$ be a smooth map between two smooth manifolds $M$ and $N$. The differential 
$dF_p : T_p M \rightarrow T_{F(p)} N$
induces a dual linear map
$dF^*_{F(p)} : T^*_{F(p)} N \rightarrow T^*_p M  $ called the (pointwise) pullback by $F$ at $F(p)$ or the cotangent map of $F$. Given a one form $\omega \in T^*_{F(p)} N$, and a vector $v \in T_p M$, $dF^*_{F(p)}$ is characterized as
\[
dF^*_{F(p)}(\omega)(v) = \omega (dF_p(v))\,.
\]
In contrast to vector fields, which can only be pushforwarded by diffeomorphisms, covector fields can always be pullback. Given a covector field $\omega  = \omega_i  dy^i \in T^*  N$ we define the covector field $F^* \omega \in T^* M$, called the pullback  of $\omega$ by $F$
\[
(F^* \omega)_p = dF^*_{F(p)}(\omega_{F(p)})\,.
\]
In coordinates,
\[
F^*\omega = F^*(\omega_i dy^i) = (\omega_i \circ F) F^* dy^i = (\omega_i \circ F) d ( y^i \circ F)
\]
which yields 
\[
(F^*\omega)_p =  \left . (\omega_i \circ F) dF^i \right |_p  =  (\omega_i(F(p))) \left . \frac{\partial F^i(x)}{\partial x^j} dx^j \right |_p .
\]
The pullback of the vector field $W$ is illustrated in Fig.~\ref{fig:Pullback}.
%

\begin{figure}[!t]
\begin{center}
\includegraphics[width=0.8\columnwidth]{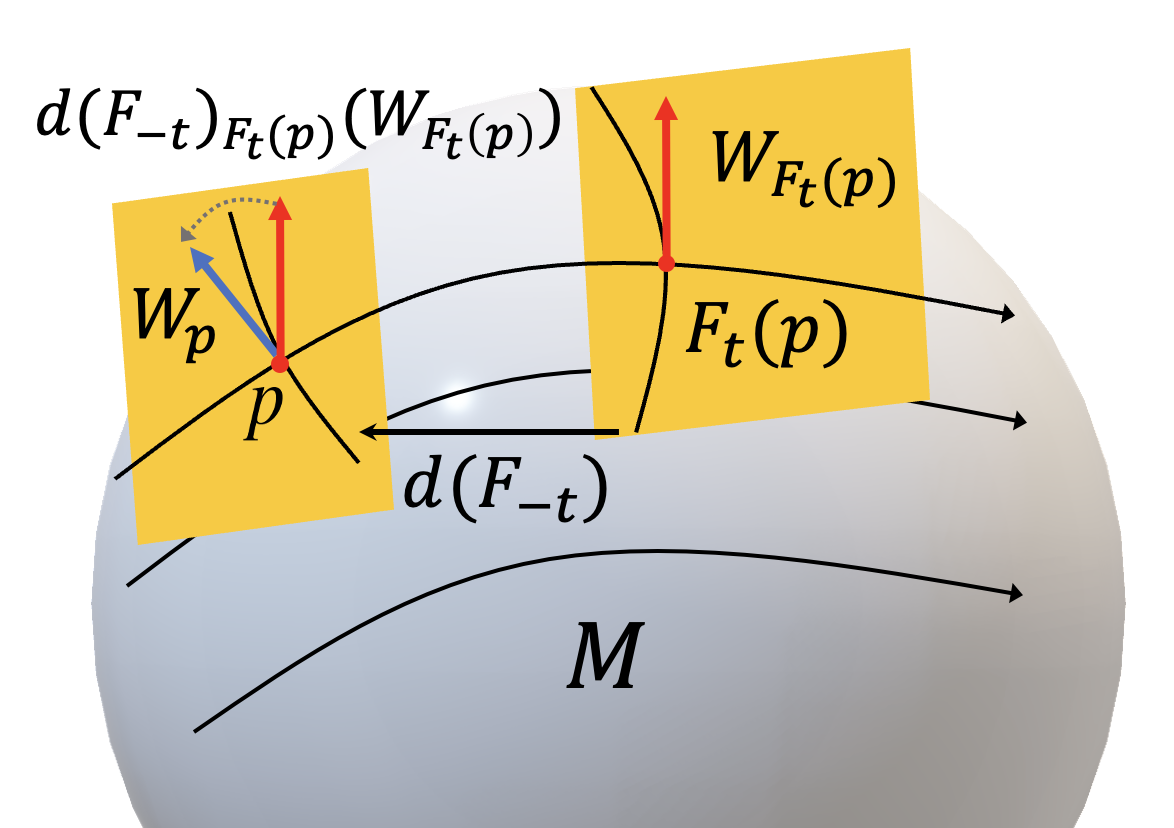} 
\vskip -0.5pc
\caption{The Lie derivative of the vector field $W$. 
}\label{LD}
\end{center}
\vskip -1.5pc
\end{figure}

\subsubsection{Lie Derivatives on Vector Fields} 
 Suppose $M$ is a smooth manifold, $V \in \mathfrak{X}(M)$ is a smooth vector field on $M$. Let $J$ be a time interval. A smooth curve $\gamma : J \to M$ determines a tangent vector $\gamma' (t)\in T_{\gamma(t)} M$. $\gamma$ is an integral curve of the vector field $V
$ if the velocity vector $\gamma' (t)
 $ is a tangent vector in $T_{\gamma(t)}M$ and
\[
\gamma' (t) = \dot \gamma^i(t) \left . \frac{\partial}{\partial x^i} \right | _{\gamma (t)} = V_{\gamma(t)} = V^i(\gamma(t)) \left . \frac{\partial}{\partial x^i} \right | _{\gamma (t)},\:\: \forall t \in J\,.
\]
Therefore, $\dot \gamma^i (t) = V^i(\gamma(t))$, $\forall t \in J$.  Suppose now that $V$ has a unique integral curve starting at $p$ and defined for all $t \in \Real$. We call this map $F_t : \Real \times M \to M$  the flow of $V$ on $M$, which sends  $p \in M$ to the point obtained by following this integral curve starting at $p$,  such  that $F_t(p) = F(t,p)$ and $F_0(p) = p$. Therefore, $F_t(p)$ is a one parameter group action
\[
F_t \circ F_s(p) = F_{t+s}(p) = F(t+s,p)
\]
and
\begin{equation}
\label{eq:vp_ddt_flow_Ft}
V_p = F_0'(p) = \left. \frac{d}{dt} \right |_{t=0} F_t(p) =  \left. \frac{\partial}{\partial t} F(t,p)\right |_{t = 0} \,.
\end{equation}

Suppose $M$ is a smooth manifold, $V$ is a smooth vector field on $M$ and $\F$ is the flow of $V$. For any
smooth vector field $W$ on $M$, the Lie derivative of $W$ with respect to $V$ at $p$, denoted by $(\Lie_V W)_p$, is given by
\begin{equation}
\label{eq:lie_derivative_vector_field}
\begin{split}
(\Lie_V W)_p:&= \frac{d}{dt}  d (\F_{-t})_{\F_t(p)}  (W_{\F_t(p)}) \bigg |_{t=0}\\
&=\lim_{t\rightarrow 0} \frac{d(\F_{-t})_{\F_t(p)} (W_{\F_t(p)})-W_p}{t}\\
&=\lim_{t\rightarrow 0} \frac{ (d (\F_{t})_{p})^{-1} (W_{\F_t(p)}) -W_p}{t}\\
 \end{split}
\end{equation}
 Notice that the vector fields $W_{\F_t(p)} \in T_{\F_t(p)} M$ and $W_p \in T_p M$ are in different tangent spaces and cannot be directly substrated. Therefore, to calculate the Lie derivative, $W_{\F_{t}(p)}$ must be pushed forward by the differential $d (\F_{-t})_{\F_t(p)}$ (the tangent map) from $T_{\F_t(p)} M$ to $T_pM$  (see Fig.~\ref{LD}), before $W_p$ can be substracted.


Let $V_p = V^i(x)  \frac{\partial}{\partial x_i}  \bigg |_{p}$ and $W_p = W^i(x)  \frac{\partial}{\partial x_i} \bigg |_{p}$.  From \eqref{eq:pushforward_coordinates} and  \eqref{eq:lie_derivative_vector_field}, we obtain
\[
\begin{split}
	(\Lie_V W)_p &= \frac{d}{dt} \left . \left ( d (\F_{-t})_{\F_t(p)}  (W_{\F_t(p)}) \right ) \right |_{t=0}\\
	&=  \frac{d}{dt} \left . \left (   \left . \left . \frac{\partial F_{-t}^j(x)}{\partial x^i} W^i(x) \right |_{F_t(p)} \frac{\partial}{\partial x^j }  \right |_{p} \right ) \right |_{t=0} \\
	&= \left (   \frac{d}{dt} \left ( \left . \frac{\partial F_{-t}^j(x)}{\partial x^i} \right |_{\F_t(p)}  \right )  W^i \circ F_t(p)  \right . \\
	& \left . \left .\hspace{2em} +   \left . \frac{\partial F_{-t}^j(x)}{\partial x^i} \right |_{\F_t(p)} \frac{d}{dt} \left ( W^i \circ F_t(p) \right ) \right ) \right |_{t=0} \left . \frac{\partial} {\partial x^j }  \right |_{p} \\
&= \left (   - \frac{\partial}{\partial x} \left ( \frac{\partial F_t^j(x)}{\partial t} \right )  W^i(x) \right . \\
	& \hspace{2em} +  \left . \left . \left .  \frac{\partial F_{-t}^j(x)}{\partial x^i}  \left ( \frac{ \partial W^i (x) }{ \partial x^l} V^l(x) \right ) \right |_{\F_t(p)}  \right ) \right |_{t=0} \left . \frac{\partial} {\partial x^j } \right |_{p} \\
&= \left (   -\frac{\partial V^j(x)}{\partial x^i}  W^i(x)    
	  +    \left .    \delta^j_i  \left ( \frac{ \partial W^i (x) }{ \partial x^l} V^l(x) \right )   \right ) \frac{\partial} {\partial x^j } \right |_{p} \\
&= \left .  \left (  \frac{ \partial W^j (x) }{ \partial x_l} V^l(x)   -\frac{\partial V^j(x)}{\partial x^i}  W^i(x)    
	         \right )  \frac{\partial} {\partial x^j } \right |_{p} \\
&=[V, W]_p.
 \end{split}
\]
Notice that we have utilized  \eqref{eq:vp_ddt_flow_Ft}, and 
\[
F_0(p)  = p \: \Rightarrow \left . \left ( \left . \frac{\partial F_{-t}^j(x)}{\partial x^i} \right |_{F_0(p)} \right ) \right|_{t=0} = \left .  \frac{\partial F_{0}^j(x)}{\partial x^i} \right |_{p} = \delta^j_i \,.
\]

\subsection{Spherical Harmonics}\label{Appendix:SH}
Spherical harmonics (SH) are motivated by studying the Laplace equation in  spherical coordinates. We will first build  SH in the complex domain, but general applications of SH are built in the real domain, and the interchange between the two domains is possible. Note also that the derivation presented here is from \cite{SH_wiki}.

Consider a function $f:\mathbb{R}^3 \to \mathbb{C}$. For the function without any sink or source, the following Laplace's equation holds:
\begin{equation} \label{eq:Laplaces}
    \nabla \cdot \nabla f(x,y,z) = \nabla ^2 f(x,y,z) = 0,
\end{equation}
where $(\cdot)$ here stands for the standard inner product.
SH naturally arises as we try to solve  \eqref{eq:Laplaces} in spherical coordinate systems, i.e., using the following change of coordinates (see Fig.~\ref{fig:change_of_coordinates}):
\begin{equation}
    x = r \sin{\theta}\cos{\phi}, \;\; y = r \sin{\theta}\sin{\phi}, \;\; z = r \cos{\theta}
\end{equation}

\begin{figure}
    \centering
    \includegraphics[width = 0.6\linewidth]{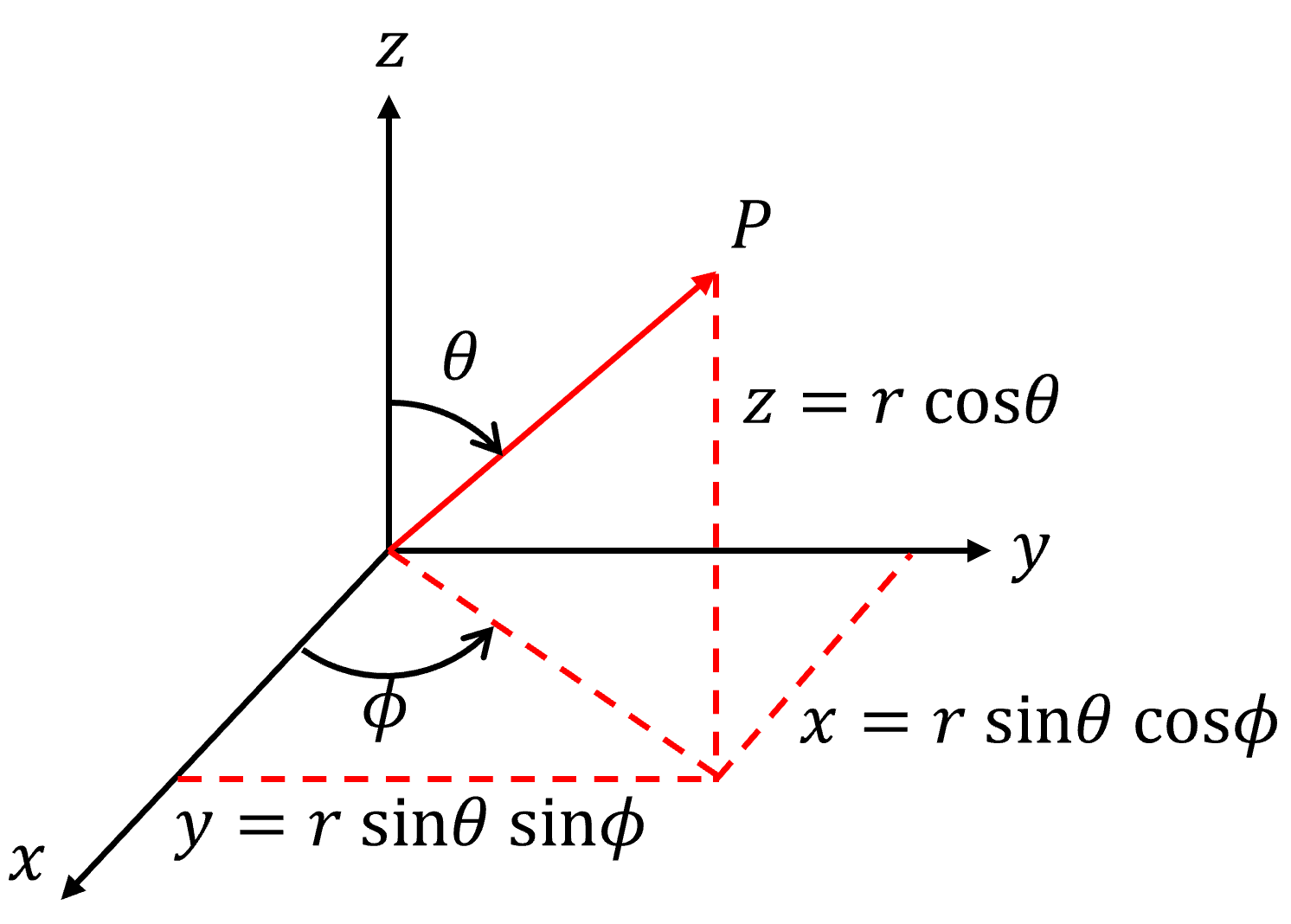}
    \caption{Change of coordinates; from Cartesian $(x,y,z)$ to the spherical coordinates $(r,\theta,\phi)$.}
    \label{fig:change_of_coordinates}
\end{figure}

Then, after some algebra, we have
\begin{equation} \label{eq:Laplace_spherical}
    \begin{split}
        \nabla^2f &= \underbrace{\dfrac{1}{r^2}\dfrac{\partial}{\partial r} \left(r^2 \dfrac{\partial f}{\partial r} \right)}_{\triangleq\lambda} \\
        &+ \underbrace{\dfrac{1}{r^2 \sin\theta} \dfrac{\partial}{\partial \theta} \left(\sin{\theta} \dfrac{\partial f}{\partial \theta}\right) + \dfrac{1}{r^2 \sin^2\theta} \dfrac{\partial^2 f}{\partial \phi ^2}}_{\triangleq-\lambda} = 0.
    \end{split}    
\end{equation}
To solve the partial differential equation (PDE) \eqref{eq:Laplace_spherical}, we first use  separation of variables as follows:
\begin{equation*}
    f(r,\theta,\phi) = R(r) Y(\theta,\phi)
\end{equation*}
By dividing \eqref{eq:Laplace_spherical} into two terms and with more algebra, we have
\begin{align}
     & \dfrac{1}{R}\dfrac{d}{dr}\left(r^2 \dfrac{dR}{dr}\right) = \lambda 
    \label{eq:radius_part}
    \\
    & \dfrac{1}{Y\sin\theta} \dfrac{\partial}{\partial \theta}\left(\sin{\theta}\dfrac{\partial Y}{\partial \theta}\right) + \dfrac{1}{Y\sin^2\theta} \dfrac{\partial^2 Y}{\partial \phi^2} = - \lambda. \label{eq:angle_part}
\end{align}
The angular terms in \eqref{eq:angle_part} can be evaluated by using separation of variable again, i.e., $Y(\theta,\phi) = \Theta(\theta)\Phi(\phi)$, which leads to
\begin{align}
    &\dfrac{1}{\Phi} \dfrac{d^2 \Phi}{d\phi^2} = - m^2 \label{eq:phi_part} \\
    &\lambda \sin^2\theta + \dfrac{\sin{\theta}}{\Theta} \dfrac{d}{d\theta} \left(\sin\theta \dfrac{d \Theta}{d \theta}\right) = m^2.  \label{eq:theta_part}
\end{align}
for some number $m$. \eqref{eq:phi_part} is a classic harmonic oscillator equation with respect to $\phi$. Therefore, $\Phi$ must be a periodic function whose period evenly divides $2\pi$, forming a boundary condition for \eqref{eq:phi_part}. Then, $m$ is necessarily an integer, and $\Phi$ is a linear combination of  complex exponentials $e^{\pm im\phi}$. 

Another fact that imposes boundary conditions on the solution to \eqref{eq:theta_part} is that, at the sphere's poles, i.e., $\theta = 0, \pi$, the function must be regular: analytic and single-valued. Therefore, by imposing this regularity into the solution of $\Theta$ at the boundary points of the domains, we have the Strum-Liouville problem. As a result, the parameter $\lambda$ in \eqref{eq:theta_part} must have the form of $ \lambda = l(l+1)$ for some non-negative integer with $l \geq |m|$.

Applying a change of variables $t = \cos{\theta}$, transforms  equation \eqref{eq:theta_part} into a Legendre equation, and the solution is given by a multiple of an associated Legendre polynomial $P^l_m$ \cite{associated_Legendre_polynomial}. Additionally, the solution of \eqref{eq:radius_part} has a solution of the form 
\begin{equation*}
    R(r) = Ar^l + Br^{-l-1},
\end{equation*}
but by requiring the solution to be regular over $\mathbb{R}^3$ makes $B = 0$, i.e., when $r \to 0$.

Combining these results altogether, for a given value of $l$, there are $2l+1$ independent solutions of the form, $Y(\theta,\phi) = \Theta(\theta)\Phi(\phi)$, one for each integer $m$ with $-l \leq m \leq l$. We denote these independent solutions associated with the integer $m$ and $l$ as $Y_m^l$, $Y_m^l:S^2 \to \mathbb{C}$, and they are the product of complex exponentials and associated Legendre polynomials as follows:
\begin{equation} \label{eq:sol_Y_detail}
    Y_m^l(\theta,\phi) = N e^{im\phi} P_m^l(\cos{\theta}).
\end{equation}
Here, $Y_m^l:S^2 \to \mathbb{C}$ is referred to as a spherical harmonic function of degree $l$ and order $m$.

Finally, the general solution $f:\mathbb{R}^3 \to \mathbb{C}$ to Laplace's equation in \eqref{eq:Laplace_spherical} in a ball centered at the origin is a linear combination of the spherical harmonic functions multiplied by the appropriate scale factor $r^l$,
\begin{equation} \label{eq:spherical_harmonics_expansion}
    f(r,\theta,\phi) = \sum_{l=0}^{\infty} \sum_{m=-l}^{l} f_m^l r^l Y_m^l(\theta,\phi)
\end{equation}
where $f_m^l \in \mathbb{C}$ are constants. 

\subsection{Equivariant Spherical Channel Network}  \label{sec:appendix_eSCN}
Specifically, eSCN exploits the sparsity of Clebsch-Gordon coefficients when the reference frame is aligned to the message-passing direction such that $x/\|x\|$ in \eqref{eqn:tfn_kernel_const} becomes $\hat{e}_y=(0,1,0)$. In this reference frame, the spherical harmonics component becomes sparse:
\begin{equation}
    \label{eqn:sparse_sh}
    Y^{l}_m(\hat{e}_y)=\delta_{m0}=\begin{cases}
        1 & \text{if } \ m=0
        \\
        0 & \text{else}
    \end{cases}
\end{equation}
To exploit this sparsity, \cite{passaro2023reducing} proposes to 1) rotate the reference frame where the message passing direction is aligned to $y$-axis, 2) apply tensor product, 3) and revert the transformation:
\begin{equation}
    \begin{split}
        f_{(out)}(x)&=\sum_{i}D(R_y^{-1}) \W(\|x-x_j\|\hat{e}_y)D(R_y)f_j
        \\
        R_y(x-x_j)&=\|x-x_j\|\hat{e}_y
    \end{split}
\end{equation}
Note that $\W(x)=D(R^{-1}) \W(Rx)D(R)$ for all $x,R$ due to the kernel equivariance constraint.

By substituting \eqref{eqn:sparse_sh} into \eqref{eqn:tfn_kernel_const}, the convolution kernel becomes
\begin{equation}
    \label{eqn:escn-kernel}
    \begin{split}
        {\W}^{(n',n)}( \hat{e}_y)\ &=\sum_{J=|l_{n'}-l_n|}^{l_{n'}+l_n}
        \phi^{(n',n)}_{J}(\| {x}\|)
        \bar{C}^{(l_n'\leftarrow l_n)}_{J}
        \\
        \left[\bar{C}^{(l_n'\leftarrow l_n)}_{J}\right]^{m'}_{\phantom{m'}m}&=C^{(l_{n'},m')}_{(l_n,m)(J,0)}
    \end{split}
\end{equation}
As pointed out by \cite{passaro2023reducing}, this tensor product can be understood as an $SO(2)$ convolution in the spectral domain. 
To show this, let us first drop the superscript $(l_n'\leftarrow l_n)$ for brevity. 
From Proposition~3.1 in \cite{passaro2023reducing}, one can show that $\bar{C}_{J}=\bar{S}_{J}+\bar{A}_{J}$ where  $\bar{S}_J$ is a symmetric diagonal matrix and $\bar{A}_J$ is an antisymmetric antidiagonal matrix. Additionally, $\bar{S}_J$ is symmetric to the row inversion.
\begin{equation}
    \label{eqn:escn-symmetry}
    \begin{split}
        \left[\bar{S}_J\right]^{m}_{\phantom{m}m}=\left[\bar{S}_J\right]^{-m}_{\phantom{-m}-m} 
        \ \ \forall\ m,\quad 
        \left[\bar{S}_J\right]^{m'}_{\phantom{m'}m}&=0\ \  \text{if}\ \ m'\neq m
        \\
        \left[\bar{A}_J\right]^{m}_{\phantom{m}-m}=-\left[\bar{A}_J\right]^{-m}_{\phantom{-m}m} 
        \ \ \forall\ m,\quad 
        \left[\bar{A}_J\right]^{m'}_{\phantom{m'}m}&=0\ \  \text{if}\ \ m'\neq -m
    \end{split}
\end{equation}
Hence, the convolution kernel ${\W}^{(n',n)}( \hat{e}_y)$ in \eqref{eqn:escn-kernel} can also be decomposed into symmetric and antisymmetric parts that follows the same condition in \eqref{eqn:escn-symmetry}. 
This is because ${\W}^{(n',n)}(\hat{e}_y)$ in \eqref{eqn:escn-kernel} is a linear combination of $\mathbf{C}_J^{(l_n'\leftarrow l_n)}$ whose coefficients are $\phi^{(n',n)}_{J}$.
\begin{equation}
    \begin{split}
        {\W}^{(n',n)}(\hat{e}_y)&=S^{(n',n)}+A^{(n',n)}
        \\
        S^{(n',n)}&=\sum_{J=|l_{n'}-l_n|}^{l_{n'}+l_n}\phi^{(n',n)}_{J}(\| {x}\|)\bar{S}_J^{(l_n'\leftarrow l_n)}
        \\
        A^{(n',n)}&=\sum_{J=|l_{n'}-l_n|}^{l_{n'}+l_n}\phi^{(n',n)}_{J}(\| {x}\|)\bar{A}_J^{(l_n'\leftarrow l_n)}
    \end{split}
\end{equation}
Therefore, the convolution kernel ${\W}^{(n',n)}(\hat{e}_y)$ is fully characterized by $2l_n+1$-dimensional vector $\W^{(n',n)}_m$ defined as
\begin{equation*}
    \W^{(n',n)}_m =
    \begin{cases}
        \left[S^{(n',n)}\right]^{m}_{\phantom{m}m} & m\geq 0
        \\
        \left[A^{(n',n)}\right]^{m}_{\phantom{m}-m} & m < 0
    \end{cases}
\end{equation*}

To show the connection with Fourier analysis, let us denote the complexification map from  $\mathbb{R}^{2l+1}$ to $\mathbb{C}^{l+1}$ as
\begin{equation}
    \tilde{f}_m
    =\begin{cases}
            \left[f\right]_{m}+i\left[f\right]_{-m} & m=1,2\cdots,l
            \\
            \left[f\right]_{0} & m =0
    \end{cases}
\end{equation}
The output tensor could be computed as the complex multiplication of these two vectors, which is similar to the $SO(2)$ convolution, which is simply a pointwise multiplication in Fourier space.
\begin{equation}
    \label{eqn:escn-fourier}
    \widetilde{(\W f)}^{(n')}_m
    =\begin{cases}
        \tilde{\W}_m^{(n',n)}\tilde{f}_m^{(n)} & 0\leq m \leq l_n
        \\
        0 & m>l_n
    \end{cases}
\end{equation}
Indeed, $\tilde{h}_m^{(n',n)}$ is related to the Fourier coefficient\footnote{Note the Fourier coefficients of real functions are fully determined by non-negative frequencies because $\tilde{f}_{-m}=\tilde{f}_m^*$.} of the convolution kernel with frequency $m$ \cite{passaro2023reducing}. 
Furthermore, there exist a linear bijection from $\phi^{(n',n)}_{J}$ to $h^{(n',n)}$ \cite{passaro2023reducing}, and hence a bijection to $\tilde{h}^{(n',n)}$. Consequently, for any TFN-like $SO(3)$ tensor product (parameterized by $SO(3)$-invariant coefficients $\phi^{(n',n)}_{J}$), there exist a unique eSCN-like $SO(2)$ convolution (parameterized by the kernel's Fourier coefficients $\tilde{h}^{(n',n)}$) that is mathematically identical but much cheaper in terms of computation.

\subsection{Comments on Right-invariant Metric} \label{appendix:right_invariance}
Bi-invariance has been proven not achievable on SE(3); either left or right invariance can be satisfied, but not both \cite{park1995distance}. While the analysis of left-invariance has provided meaningful conclusions in Section~\ref{sec:GIC}, we now aim to explore the insights that may arise from examining the system's right-invariance. Physically, right invariance means that the distance metric remains unaffected by the specific location of the body-fixed frame on the robotic manipulator arm.  

The right-invariance of the error function $\Psi^R:\SE \times \SE \to \mathbb{R}_{\geq0}$ can be expressed as follows:
\begin{equation}
\begin{split}
    &\Psi^R(g g_r, g_d g_r) = \frac{1}{2} \| I - g g_r g_r^{-1} g_d^{-1} \|_F^2 = \Psi^R(g, g_d) \\
   &\Psi^R(g, g_d) = \frac{1}{2} \| I - g g_d^T \|_F^2 = \frac{1}{2} \, \text{tr} \left( (I - g g_d^{-1})^T (I - g g_d^{-1}) \right) \\
    &= \frac{1}{2} \, \text{tr} \left( I - g_d^{-T} g^T - g g_d^{-1} + g_d^{-T} g^T g g_d^{-1} \right) \\
    &= \text{tr} (I - R R_d^T) + \frac{1}{2} (p - R R_d^T p_d)^T (p - R R_d^T p_d) \nonumber
\end{split}
\end{equation}

Breaking this down, the error function yields a translation component that is influenced by rotation under right invariance, while for left invariance, the translation is independent of the rotation \eqref{eq:eg}.

To proceed, since left invariance was previously established by perturbing $g \in SE(3)$ with a right transformation, we now investigate right invariance by perturbing $g$ with a left transformation. For a small perturbation, we define:
\begin{equation}
\begin{split}
    g_\eta &= e^{\hat{\eta} \epsilon} \, g, \quad \eta = \begin{bmatrix} \eta_1 \\ \eta_2 \end{bmatrix} \in \mathbb{R}^6 \implies \hat{\eta} \in se(3), \; \epsilon \in \mathbb{R}, \\
    \delta g &= \left. \frac{d}{d\epsilon} \left( e^{\hat{\eta} \epsilon} \right) g \right|_{\epsilon=0} = \hat{\eta}\left. e^{\hat{\eta} \epsilon} g \right|_{\epsilon=0} = \hat{\eta} g \\
    &= \begin{bmatrix} \delta R & \delta p \\ 0 & 0 \end{bmatrix} 
    = \begin{bmatrix} \hat{\eta}_2 & \eta_1 \\ 0 & 0 \end{bmatrix} 
    \begin{bmatrix} R & p \\ 0 & 1 \end{bmatrix} \\
    &= \begin{bmatrix} \hat{\eta}_2 R & \hat{\eta}_2 p + \eta_1 \\ 0 & 0 \end{bmatrix} \nonumber
\end{split}
\end{equation}

The perturbed error function becomes:
\begin{equation}
\begin{split}
    \delta \Psi^R &= -\text{tr}(\delta R R_d^T) + (p - R R_d^T p_d)^T (\delta p - \delta R R_d^T p_d) \\
    &= -\text{tr}(\hat{\eta_2} R R_d^T) + (p - R R_d^T p_d)^T \eta_1, \nonumber \\
\end{split}
\end{equation}

so that $\delta \Psi^R = e_g^R \cdot \eta$ and
\begin{equation}
\begin{split}
     e_G^R = \begin{bmatrix}
         e_p^R \\ e_R^R
     \end{bmatrix} = \begin{bmatrix} (p - R R_d^T p_d) \\ (R R_d^T - R_d R^T)^\vee \end{bmatrix}. \nonumber
\end{split}
\end{equation}

If we further investigate $e_p^R$:
\begin{equation*}
    e_p^R = p - R R_d^T p_d  = R ( R^T p - R^T_d p_d)
\end{equation*}

The left-invariant error, obtained earlier, is $e_p^L = R^T \left (p - p_d \right )$. The key difference is that the right-invariant error vector depends on the spatial frame, while the left-invariant error vector remains independent. 

Since right invariance ties the error vector to the reference frame, it makes the interpretation of the error more complex. 
While left invariance has shown practical benefits, the implications of right invariance remain unclear. Therefore, as bi-invariance has been proven unachievable on $\SE$, it is sufficient to focus on addressing the specific invariance limitations of the system \cite{park1995distance}.

\bibliographystyle{unsrt}
\bibliography{reference}

\begin{thebibliography}{10}

\bibitem{ryu2022equivariant}
Hyunwoo Ryu, Jeong-Hoon Lee, Hong-in Lee, and Jongeun Choi.
\newblock Equivariant descriptor fields: Se (3)-equivariant energy-based models for end-to-end visual robotic manipulation learning.
\newblock {\em in the Proceedings of 2023 International Conference on Learning Representations (ICLR) 2023arXiv preprint arXiv:2206.08321}, 2022.

\bibitem{ryu2024diffusion}
Hyunwoo Ryu, Jiwoo Kim, Hyunseok An, Junwoo Chang, Joohwan Seo, Taehan Kim, Yubin Kim, Chaewon Hwang, Jongeun Choi, and Roberto Horowitz.
\newblock Diffusion-edfs: Bi-equivariant denoising generative modeling on {SE(3)} for visual robotic manipulation.
\newblock In {\em Proceedings of the IEEE/CVF Conference on Computer Vision and Pattern Recognition}, pages 18007--18018, 2024.

\bibitem{kim2023robotic}
Jiwoo Kim, Hyunwoo Ryu, Jongeun Choi, Joohwan Seo, Nikhil Potu~Surya Prakash, Ruolin Li, and Roberto Horowitz.
\newblock Robotic manipulation learning with equivariant descriptor fields: Generative modeling, bi-equivariance, steerability, and locality.
\newblock In {\em RSS 2023 Workshop on Symmetries in Robot Learning}, 2023.

\bibitem{seo2023geometric}
Joohwan Seo, Nikhil Potu~Surya Prakash, Alexander Rose, Jongeun Choi, and Roberto Horowitz.
\newblock Geometric impedance control on {SE(3)} for robotic manipulators.
\newblock {\em IFAC World Congress 2023, Yokohama, Japan}, 2023.

\bibitem{seo2024comparison}
Joohwan Seo, Nikhil Potu~Surya Prakash, Jongeun Choi, and Roberto Horowitz.
\newblock A comparison between lie group-and lie algebra-based potential functions for geometric impedance control.
\newblock {\em arXiv preprint arXiv:2401.13190}, 2024.

\bibitem{seo2023contact}
Joohwan Seo, Nikhil~PS Prakash, Xiang Zhang, Changhao Wang, Jongeun Choi, Masayoshi Tomizuka, and Roberto Horowitz.
\newblock Contact-rich {SE(3)}-equivariant robot manipulation task learning via geometric impedance control.
\newblock {\em IEEE Robotics and Automation Letters}, 2023.

\bibitem{ping2002group}
Jialun Ping, Fan Wang, and Jin-Quan Chen.
\newblock {\em Group representation theory for physicists}, chapter~5.
\newblock World Scientific Publishing Company, 2002.

\bibitem{huang2020lie}
Jiaqi Huang.
\newblock Lie groups and their applications to particle physics: A tutorial for undergraduate physics majors.
\newblock {\em arXiv preprint arXiv:2012.00834}, 2020.

\bibitem{chirikjian2011stochastic}
Gregory~S Chirikjian.
\newblock {\em Stochastic models, information theory, and Lie groups, volume 2: Analytic methods and modern applications}, volume~2.
\newblock Springer Science \& Business Media, 2011.

\bibitem{lynch2017modern}
Kevin~M Lynch and Frank~C Park.
\newblock {\em Modern robotics}.
\newblock Cambridge University Press, 2017.

\bibitem{murray1994mathematical}
Richard~M Murray, Zexiang Li, S~Shankar Sastry, and S~Shankara Sastry.
\newblock {\em A mathematical introduction to robotic manipulation}.
\newblock CRC press, 1994.

\bibitem{veefkind2024probabilistic}
Lars Veefkind and Gabriele Cesa.
\newblock A probabilistic approach to learning the degree of equivariance in steerable cnns.
\newblock {\em arXiv preprint arXiv:2406.03946}, 2024.

\bibitem{UvA_website}
Erik Bekkers.
\newblock {University of Amsterdam, An Introduction to Group Equivariant Deep Learning, Course Website}.
\newblock \url{https://uvagedl.github.io/}.
\newblock Accessed: 2024-12-02.

\bibitem{krizhevsky2012imagenet}
Alex Krizhevsky, Ilya Sutskever, and Geoffrey~E Hinton.
\newblock Imagenet classification with deep convolutional neural networks.
\newblock {\em Advances in neural information processing systems}, 25, 2012.

\bibitem{deng2009imagenet}
Jia Deng, Wei Dong, Richard Socher, Li-Jia Li, Kai Li, and Li~Fei-Fei.
\newblock Imagenet: A large-scale hierarchical image database.
\newblock In {\em 2009 IEEE conference on computer vision and pattern recognition}, pages 248--255. Ieee, 2009.

\bibitem{PytorchAlexNet}
Pre-trained {AlexNet from PyTorch}.
\newblock \url{https://pytorch.org/hub/pytorch_vision_alexnet/}.
\newblock Accessed: 2024-07-14.

\bibitem{cohen2016group}
Taco Cohen and Max Welling.
\newblock Group equivariant convolutional networks.
\newblock In {\em International conference on machine learning}, pages 2990--2999. PMLR, 2016.

\bibitem{leesmooth}
John~M Lee.
\newblock {\em Introduction to Smooth manifolds}.
\newblock Springer, 2012.

\bibitem{lee2018real}
Taeyoung Lee.
\newblock Real harmonic analysis on the special orthogonal group.
\newblock {\em arXiv preprint arXiv:1809.10533}, 2018.

\bibitem{aubert2013alternative}
G~Aubert.
\newblock An alternative to wigner d-matrices for rotating real spherical harmonics.
\newblock {\em AIP Advances}, 3(6):062121, 2013.

\bibitem{se3t}
Fabian Fuchs, Daniel Worrall, Volker Fischer, and Max Welling.
\newblock {SE(3)}-transformers: 3d roto-translation equivariant attention networks.
\newblock {\em Advances in Neural Information Processing Systems}, 33:1970--1981, 2020.

\bibitem{thomas2018tensor}
Nathaniel Thomas, Tess Smidt, Steven Kearnes, Lusann Yang, Li~Li, Kai Kohlhoff, and Patrick Riley.
\newblock Tensor field networks: Rotation-and translation-equivariant neural networks for 3d point clouds.
\newblock {\em arXiv preprint arXiv:1802.08219}, 2018.

\bibitem{freeman1991design}
William~T Freeman, Edward~H Adelson, et~al.
\newblock The design and use of steerable filters.
\newblock {\em IEEE Transactions on Pattern analysis and machine intelligence}, 13(9):891--906, 1991.

\bibitem{brandstetter2021geometric}
Johannes Brandstetter, Rob Hesselink, Elise van~der Pol, Erik~J Bekkers, and Max Welling.
\newblock Geometric and physical quantities improve {E(3)} equivariant message passing.
\newblock {\em arXiv preprint arXiv:2110.02905}, 2021.

\bibitem{zee}
Anthony Zee.
\newblock {\em Group theory in a nutshell for physicists}, volume~17.
\newblock Princeton University Press, 2016.

\bibitem{griffiths2018introduction}
David~J Griffiths and Darrell~F Schroeter.
\newblock {\em Introduction to quantum mechanics}.
\newblock Cambridge university press, 2018.

\bibitem{wang2019dynamic}
Yue Wang, Yongbin Sun, Ziwei Liu, Sanjay~E Sarma, Michael~M Bronstein, and Justin~M Solomon.
\newblock Dynamic graph cnn for learning on point clouds.
\newblock {\em Acm Transactions On Graphics (tog)}, 38(5):1--12, 2019.

\bibitem{te2018rgcnn}
Gusi Te, Wei Hu, Amin Zheng, and Zongming Guo.
\newblock Rgcnn: Regularized graph cnn for point cloud segmentation.
\newblock In {\em Proceedings of the 26th ACM international conference on Multimedia}, pages 746--754, 2018.

\bibitem{shi2020point}
Weijing Shi and Raj Rajkumar.
\newblock Point-gnn: Graph neural network for 3d object detection in a point cloud.
\newblock In {\em Proceedings of the IEEE/CVF conference on computer vision and pattern recognition}, pages 1711--1719, 2020.

\bibitem{liao2022equiformer}
Yi-Lun Liao and Tess Smidt.
\newblock Equiformer: Equivariant graph attention transformer for 3d atomistic graphs.
\newblock {\em arXiv preprint arXiv:2206.11990}, 2022.

\bibitem{weiler20183d}
Maurice Weiler, Mario Geiger, Max Welling, Wouter Boomsma, and Taco~S Cohen.
\newblock 3d steerable cnns: Learning rotationally equivariant features in volumetric data.
\newblock {\em Advances in Neural Information Processing Systems}, 31, 2018.

\bibitem{passaro2023reducing}
Saro Passaro and C~Lawrence Zitnick.
\newblock Reducing {SO(3)} convolutions to {SO(2)} for efficient equivariant gnns.
\newblock In {\em International Conference on Machine Learning}, pages 27420--27438. PMLR, 2023.

\bibitem{gao2024riemann}
Chongkai Gao, Zhengrong Xue, Shuying Deng, Tianhai Liang, Siqi Yang, Lin Shao, and Huazhe Xu.
\newblock Riemann: Near real-time {SE(3)}-equivariant robot manipulation without point cloud segmentation.
\newblock {\em arXiv preprint arXiv:2403.19460}, 2024.

\bibitem{hu2024orbitgrasp}
Boce Hu, Xupeng Zhu, Dian Wang, Zihao Dong, Haojie Huang, Chenghao Wang, Robin Walters, and Robert Platt.
\newblock Orbitgrasp: $ {SE(3)} $-equivariant grasp learning.
\newblock {\em arXiv preprint arXiv:2407.03531}, 2024.

\bibitem{luo2024enabling}
Shengjie Luo, Tianlang Chen, and Aditi~S Krishnapriyan.
\newblock Enabling efficient equivariant operations in the fourier basis via gaunt tensor products.
\newblock {\em arXiv preprint arXiv:2401.10216}, 2024.

\bibitem{qi2017pointnet}
Charles~R Qi, Hao Su, Kaichun Mo, and Leonidas~J Guibas.
\newblock Pointnet: Deep learning on point sets for 3d classification and segmentation.
\newblock In {\em Proceedings of the IEEE conference on computer vision and pattern recognition}, pages 652--660, 2017.

\bibitem{deng2021vector}
Congyue Deng, Or~Litany, Yueqi Duan, Adrien Poulenard, Andrea Tagliasacchi, and Leonidas~J Guibas.
\newblock Vector neurons: A general framework for {SO(3)}-equivariant networks.
\newblock In {\em Proceedings of the IEEE/CVF International Conference on Computer Vision}, pages 12200--12209, 2021.

\bibitem{zeng2020transporter}
Andy Zeng, Pete Florence, Jonathan Tompson, Stefan Welker, Jonathan Chien, Maria Attarian, Travis Armstrong, Ivan Krasin, Dan Duong, Vikas Sindhwani, and Johnny Lee.
\newblock Transporter networks: Rearranging the visual world for robotic manipulation.
\newblock {\em Conference on Robot Learning (CoRL)}, 2020.

\bibitem{huang2022equivariant}
Haojie Huang, Dian Wang, Robin Walter, and Robert Platt.
\newblock Equivariant transporter network.
\newblock {\em arXiv preprint arXiv:2202.09400}, 2022.

\bibitem{huang2024fourier}
Haojie Huang, Owen Howell, Xupeng Zhu, Dian Wang, Robin Walters, and Robert Platt.
\newblock Fourier transporter: Bi-equivariant robotic manipulation in 3d.
\newblock {\em arXiv preprint arXiv:2401.12046}, 2024.

\bibitem{huang2023edge}
Haojie Huang, Dian Wang, Xupeng Zhu, Robin Walters, and Robert Platt.
\newblock Edge grasp network: A graph-based {SE(3)}-invariant approach to grasp detection.
\newblock In {\em 2023 IEEE International Conference on Robotics and Automation (ICRA)}, pages 3882--3888. IEEE, 2023.

\bibitem{chi2023diffusion}
Cheng Chi, Zhenjia Xu, Siyuan Feng, Eric Cousineau, Yilun Du, Benjamin Burchfiel, Russ Tedrake, and Shuran Song.
\newblock Diffusion policy: Visuomotor policy learning via action diffusion.
\newblock {\em The International Journal of Robotics Research}, page 02783649241273668, 2023.

\bibitem{yang2024equibot}
Jingyun Yang, Zi-ang Cao, Congyue Deng, Rika Antonova, Shuran Song, and Jeannette Bohg.
\newblock Equibot: {Sim(3)}-equivariant diffusion policy for generalizable and data efficient learning.
\newblock {\em arXiv preprint arXiv:2407.01479}, 2024.

\bibitem{yang2024equivact}
Jingyun Yang, Congyue Deng, Jimmy Wu, Rika Antonova, Leonidas Guibas, and Jeannette Bohg.
\newblock Equivact: {Sim(3)}-equivariant visuomotor policies beyond rigid object manipulation.
\newblock In {\em 2024 IEEE International Conference on Robotics and Automation (ICRA)}, pages 9249--9255. IEEE, 2024.

\bibitem{wang2024equivariant}
Dian Wang, Stephen Hart, David Surovik, Tarik Kelestemur, Haojie Huang, Haibo Zhao, Mark Yeatman, Jiuguang Wang, Robin Walters, and Robert Platt.
\newblock Equivariant diffusion policy.
\newblock {\em arXiv preprint arXiv:2407.01812}, 2024.

\bibitem{tie2024seed}
Chenrui Tie, Yue Chen, Ruihai Wu, Boxuan Dong, Zeyi Li, Chongkai Gao, and Hao Dong.
\newblock Et-seed: Efficient trajectory-level {SE(3)} equivariant diffusion policy.
\newblock {\em arXiv preprint arXiv:2411.03990}, 2024.

\bibitem{funk2024actionflow}
Niklas Funk, Julen Urain, Joao Carvalho, Vignesh Prasad, Georgia Chalvatzaki, and Jan Peters.
\newblock Actionflow: Equivariant, accurate, and efficient policies with spatially symmetric flow matching.
\newblock {\em arXiv preprint arXiv:2409.04576}, 2024.

\bibitem{lim2024equigraspflow}
Byeongdo Lim, Jongmin Kim, Jihwan Kim, Yonghyeon Lee, and Frank~C Park.
\newblock Equigraspflow: {SE(3)}-equivariant 6-dof grasp pose generative flows.
\newblock In {\em 8th Annual Conference on Robot Learning}, 2024.

\bibitem{zeng}
Andy Zeng, Pete Florence, Jonathan Tompson, Stefan Welker, Jonathan Chien, Maria Attarian, Travis Armstrong, Ivan Krasin, Dan Duong, Vikas Sindhwani, et~al.
\newblock Transporter networks: Rearranging the visual world for robotic manipulation.
\newblock {\em arXiv preprint arXiv:2010.14406}, 2020.

\bibitem{fourtran}
Haojie Huang, Owen Howell, Xupeng Zhu, Dian Wang, Robin Walters, and Robert Platt.
\newblock Fourier transporter: Bi-equivariant robotic manipulation in 3d.
\newblock {\em arXiv preprint arXiv:2401.12046}, 2024.

\bibitem{ndf}
Anthony Simeonov, Yilun Du, Andrea Tagliasacchi, Joshua~B Tenenbaum, Alberto Rodriguez, Pulkit Agrawal, and Vincent Sitzmann.
\newblock Neural descriptor fields: {SE(3)}-equivariant object representations for manipulation.
\newblock {\em arXiv preprint arXiv:2112.05124}, 2021.

\bibitem{rndf}
Anthony Simeonov, Yilun Du, Lin Yen-Chen, Alberto Rodriguez, Leslie~Pack Kaelbling, Tomas Lozano-Perez, and Pulkit Agrawal.
\newblock {SE(3)}-equivariant relational rearrangement with neural descriptor fields.
\newblock {\em arXiv preprint arXiv:2211.09786}, 2022.

\bibitem{rl3}
Elise van~der Pol, Daniel Worrall, Herke van Hoof, Frans Oliehoek, and Max Welling.
\newblock Mdp homomorphic networks: Group symmetries in reinforcement learning.
\newblock {\em Advances in Neural Information Processing Systems}, 33:4199--4210, 2020.

\bibitem{finzi2021residual}
Marc Finzi, Gregory Benton, and Andrew~G Wilson.
\newblock Residual pathway priors for soft equivariance constraints.
\newblock {\em Advances in Neural Information Processing Systems}, 34:30037--30049, 2021.

\bibitem{theile2024equivariant}
Mirco Theile, Hongpeng Cao, Marco Caccamo, and Alberto~L Sangiovanni-Vincentelli.
\newblock Equivariant ensembles and regularization for reinforcement learning in map-based path planning.
\newblock {\em arXiv preprint arXiv:2403.12856}, 2024.

\bibitem{wang2022so}
Dian Wang and Robin Walters.
\newblock {SO(2)} equivariant reinforcement learning.
\newblock In {\em International Conference on Learning Representations}, 2022.

\bibitem{kohler2023symmetric}
Colin Kohler, Anuj~Shrivatsav Srikanth, Eshan Arora, and Robert Platt.
\newblock Symmetric models for visual force policy learning.
\newblock {\em arXiv preprint arXiv:2308.14670}, 2023.

\bibitem{wang2022robot}
Dian Wang, Mingxi Jia, Xupeng Zhu, Robin Walters, and Robert Platt.
\newblock On-robot learning with equivariant models.
\newblock {\em arXiv preprint arXiv:2203.04923}, 2022.

\bibitem{wang2022equivariant}
Dian Wang, Robin Walters, Xupeng Zhu, and Robert Platt.
\newblock Equivariant $ q $ learning in spatial action spaces.
\newblock In {\em Conference on Robot Learning}, pages 1713--1723. PMLR, 2022.

\bibitem{nguyen2023equivariant}
Hai~Huu Nguyen, Andrea Baisero, David Klee, Dian Wang, Robert Platt, and Christopher Amato.
\newblock Equivariant reinforcement learning under partial observability.
\newblock In {\em Conference on Robot Learning}, pages 3309--3320. PMLR, 2023.

\bibitem{zhao2024equivariant}
Linfeng Zhao, Owen Howell, Xupeng Zhu, Jung~Yeon Park, Zhewen Zhang, Robin Walters, and Lawson~LS Wong.
\newblock Equivariant action sampling for reinforcement learning and planning.
\newblock {\em arXiv preprint arXiv:2412.12237}, 2024.

\bibitem{zhao2024E2}
Linfeng Zhao, Hongyu Li, Ta{\c{s}}k{\i}n Pad{\i}r, Huaizu Jiang, and Lawson~LS Wong.
\newblock {E(2)}-equivariant graph planning for navigation.
\newblock {\em IEEE Robotics and Automation Letters}, 9(4):3371--3378, 2024.

\bibitem{zhao2022integrating}
Linfeng Zhao, Xupeng Zhu, Lingzhi Kong, Robin Walters, and Lawson~LS Wong.
\newblock Integrating symmetry into differentiable planning with steerable convolutions.
\newblock {\em arXiv preprint arXiv:2206.03674}, 2022.

\bibitem{hoffmann1989iterative}
Walter Hoffmann.
\newblock Iterative algorithmen f{\"u}r die gram-schmidt-orthogonalisierung.
\newblock {\em Computing}, 41:335--348, 1989.

\bibitem{ravindran2001symmetries}
Balaraman Ravindran and Andrew~G Barto.
\newblock Symmetries and model minimization in markov decision processes, 2001.

\bibitem{ravindran2004approximate}
Balaraman Ravindran and Andrew~G Barto.
\newblock Approximate homomorphisms: A framework for non-exact minimization in markov decision processes.
\newblock 2004.

\bibitem{weiler2019general}
Maurice Weiler and Gabriele Cesa.
\newblock General {E(2)}-equivariant steerable cnns.
\newblock {\em Advances in neural information processing systems}, 32, 2019.

\bibitem{laskin2020reinforcement}
Misha Laskin, Kimin Lee, Adam Stooke, Lerrel Pinto, Pieter Abbeel, and Aravind Srinivas.
\newblock Reinforcement learning with augmented data.
\newblock {\em Advances in neural information processing systems}, 33:19884--19895, 2020.

\bibitem{kostrikov2020image}
Ilya Kostrikov, Denis Yarats, and Rob Fergus.
\newblock Image augmentation is all you need: Regularizing deep reinforcement learning from pixels.
\newblock {\em arXiv preprint arXiv:2004.13649}, 2020.

\bibitem{laskin2020curl}
Michael Laskin, Aravind Srinivas, and Pieter Abbeel.
\newblock Curl: Contrastive unsupervised representations for reinforcement learning.
\newblock In {\em International conference on machine learning}, pages 5639--5650. PMLR, 2020.

\bibitem{tangri2024equivariant}
Arsh Tangri, Ondrej Biza, Dian Wang, David Klee, Owen Howell, and Robert Platt.
\newblock Equivariant offline reinforcement learning.
\newblock {\em arXiv preprint arXiv:2406.13961}, 2024.

\bibitem{kumar2020conservative}
Aviral Kumar, Aurick Zhou, George Tucker, and Sergey Levine.
\newblock Conservative q-learning for offline reinforcement learning.
\newblock {\em Advances in Neural Information Processing Systems}, 33:1179--1191, 2020.

\bibitem{kostrikov2021offline}
Ilya Kostrikov, Ashvin Nair, and Sergey Levine.
\newblock Offline reinforcement learning with implicit q-learning.
\newblock {\em arXiv preprint arXiv:2110.06169}, 2021.

\bibitem{khatib1987unified}
Oussama Khatib.
\newblock A unified approach for motion and force control of robot manipulators: The operational space formulation.
\newblock {\em IEEE Journal on Robotics and Automation}, 3(1):43--53, 1987.

\bibitem{hogan1985impedance}
Neville Hogan.
\newblock Impedance control: An approach to manipulation: Part ii—implementation.
\newblock 1985.

\bibitem{caccavale1999six}
Fabrizio Caccavale, Ciro Natale, Bruno Siciliano, and Luigi Villani.
\newblock Six-dof impedance control based on angle/axis representations.
\newblock {\em IEEE Transactions on Robotics and Automation}, 15(2):289--300, 1999.

\bibitem{koditschek1989application}
Daniel~E Koditschek.
\newblock The application of total energy as a lyapunov function for mechanical control systems.
\newblock {\em Contemporary mathematics}, 97:131, 1989.

\bibitem{bullo1999tracking}
Francesco Bullo and Richard~M Murray.
\newblock Tracking for fully actuated mechanical systems: a geometric framework.
\newblock {\em Automatica}, 35(1):17--34, 1999.

\bibitem{lee2010geometric}
Taeyoung Lee, Melvin Leok, and N~Harris McClamroch.
\newblock Geometric tracking control of a quadrotor uav on {SE (3)}.
\newblock In {\em 49th IEEE conference on decision and control (CDC)}, pages 5420--5425. IEEE, 2010.

\bibitem{teng2022lie}
Sangli Teng, William Clark, Anthony Bloch, Ram Vasudevan, and Maani Ghaffari.
\newblock Lie algebraic cost function design for control on lie groups.
\newblock In {\em 2022 IEEE 61st Conference on Decision and Control (CDC)}, pages 1867--1874. IEEE, 2022.

\bibitem{bullo1995proportional}
Francesco Bullo and Richard~M Murray.
\newblock Proportional derivative {(PD)} control on the euclidean group.
\newblock 1995.

\bibitem{lee2012exponential}
Taeyoung Lee.
\newblock Exponential stability of an attitude tracking control system on {SO(3)} for large-angle rotational maneuvers.
\newblock {\em Systems \& Control Letters}, 61(1):231--237, 2012.

\bibitem{huynh2009metrics}
Du~Q Huynh.
\newblock Metrics for 3d rotations: Comparison and analysis.
\newblock {\em Journal of Mathematical Imaging and Vision}, 35(2):155--164, 2009.

\bibitem{Matrix_log_wiki}
{Logarithm of a matrix Wikipedia}.
\newblock \url{https://en.wikipedia.org/wiki/Logarithm_of_a_matrix}.
\newblock Accessed: 2024-06-07.

\bibitem{park1995distance}
Frank~C Park.
\newblock Distance metrics on the rigid-body motions with applications to mechanism design.
\newblock 1995.

\bibitem{vzefran1996choice}
Milo{\v{s}} {\v{Z}}efran, Vijay Kumar, and Christopher Croke.
\newblock Choice of riemannian metrics for rigid body kinematics.
\newblock In {\em International Design Engineering Technical Conferences and Computers and Information in Engineering Conference}, volume 97584, page V02BT02A030. American Society of Mechanical Engineers, 1996.

\bibitem{belta2002euclidean}
Calin Belta and Vijay Kumar.
\newblock Euclidean metrics for motion generation on {SE(3)}.
\newblock {\em Proceedings of the Institution of Mechanical Engineers, Part C: Journal of Mechanical Engineering Science}, 216(1):47--60, 2002.

\bibitem{liu2013finite}
Yongfang Liu and Zhiyong Geng.
\newblock Finite-time optimal formation control of multi-agent systems on the lie group {SE(3)}.
\newblock {\em International Journal of Control}, 86(10):1675--1686, 2013.

\bibitem{kronander2016stability}
Klas Kronander and Aude Billard.
\newblock Stability considerations for variable impedance control.
\newblock {\em IEEE Transactions on Robotics}, 32(5):1298--1305, 2016.

\bibitem{shridhar2022cliport}
Mohit Shridhar, Lucas Manuelli, and Dieter Fox.
\newblock Cliport: What and where pathways for robotic manipulation.
\newblock In {\em Conference on Robot Learning}, pages 894--906. PMLR, 2022.

\bibitem{wang2022surprising}
Dian Wang, Jung~Yeon Park, Neel Sortur, Lawson~LS Wong, Robin Walters, and Robert Platt.
\newblock The surprising effectiveness of equivariant models in domains with latent symmetry.
\newblock {\em arXiv preprint arXiv:2211.09231}, 2022.

\bibitem{SH_wiki}
{Spherical Harmonics Wikipedia}.
\newblock \url{https://en.wikipedia.org/wiki/Spherical_harmonics}.
\newblock Accessed: 2024-06-21.

\bibitem{associated_Legendre_polynomial}
{Associated Legendre Polynomial Wikipedia}.
\newblock \url{https://en.wikipedia.org/wiki/Associated_Legendre_polynomials}.
\newblock Accessed: 2024-06-21.

\end{thebibliography}

\end{document}